\documentclass{article}

\usepackage{graphicx}
\usepackage{xcolor} % For defining custom colors
\usepackage{amsmath}
\usepackage{wrapfig}
\usepackage{booktabs}
\usepackage{inconsolata}
\usepackage{pifont}
\usepackage{comment}
\usepackage{colortbl} % For colored cells and rows
\usepackage{enumitem}
\usepackage{booktabs} % For better tables
\usepackage{booktabs}
\usepackage{enumitem}
\usepackage{graphicx}
\usepackage{geometry}
\usepackage{caption}
\usepackage{pgf}           % Needed for calculations
\usepackage{amssymb} % For more symbols
\usepackage{multirow}
\usepackage{multicol}
\usepackage{array}
\usepackage{float}
\usepackage{subcaption}

\usepackage{caption} 
\newtheorem{definition}{Definition}[section]

\newcommand{\scorecolor}[1]{%
  \pgfmathsetmacro{\intensity}{#1 * 100} % Scale the value to range 0-100
  \edef\x{\noexpand\cellcolor{green!\intensity}} % Create the cell color command
  \x\textcolor{black}{#1} % Set the text color and print the value
}

\usepackage[final]{neurips_data_2024}

\usepackage[utf8]{inputenc} % allow utf-8 input
\usepackage[T1]{fontenc}    % use 8-bit T1 fonts
\usepackage{hyperref}       % hyperlinks
\usepackage{url}            % simple URL typesetting
\usepackage{booktabs}       % professional-quality tables
\usepackage{amsfonts}       % blackboard math symbols
\usepackage{nicefrac}       % compact symbols for 1/2, etc.
\usepackage{microtype}      % microtypography
\usepackage{xcolor}         % colors
\usepackage{fontawesome}

\newcommand{\libertineWord}[1]{{\fontfamily{LinuxLibertineT-TLF}\selectfont #1}}

\newcommand{\vinod}[1]{\textcolor{purple}{#1 -VP}}
\newcommand{\nithish}[1]{\textcolor{blue}{#1 -NK}}
\newcommand{\marco}[1]{\textcolor{violet}{#1 -MA}}

\title{  Beyond Aesthetics: Cultural Competence in Text-to-Image Models}

\title{Beyond Aesthetics: Cultural Competence in Text-to-Image Models}

\author{%
  Nithish Kannen$^\mathsection$, \quad
  Arif Ahmad$^\ddagger$\thanks{Work done while Arif Ahmad was a student researcher at Google Research.}, \quad
  Marco Andreetto$^\mathsection$, \quad
  Vinodkumar Prabhakaran$^\spadesuit$, \quad \\
  \bfseries Utsav Prabhu$^\mathsection$, \quad
  Adji Bousso Dieng$^{\mathparagraph\mathsection}$, \quad
  Pushpak Bhattacharyya$^\ddagger$, \quad
  Shachi Dave$^\mathsection$ \\
  \\
  $^\mathsection$Google DeepMind, \quad
  $^\spadesuit$Google Research, \quad
  $^\ddagger$IIT Bombay, \quad
  $^\mathparagraph$Princeton \\
  \vspace{2mm}
  \footnotesize \textbf{Correspondence}: \href{mailto:nitkan@google.com}{\texttt{\{nitkan}}, \href{mailto:shachi@google.com}{\texttt{ shachi\}}@google.com}
}

\definecolor{ContinentColor}{RGB}{230, 230, 255}
\definecolor{CountryColor}{RGB}{255, 230, 230}
\definecolor{ArtifactColor}{RGB}{230, 255, 230}
\definecolor{HierarchicalColor}{RGB}{255, 255, 230}
\definecolor{UniformColor}{RGB}{230, 255, 255}

\begin{document}

\maketitle

\vspace{-4mm}

\begin{figure*}[!ht]
    \centering
    \includegraphics[width=1.0\textwidth]{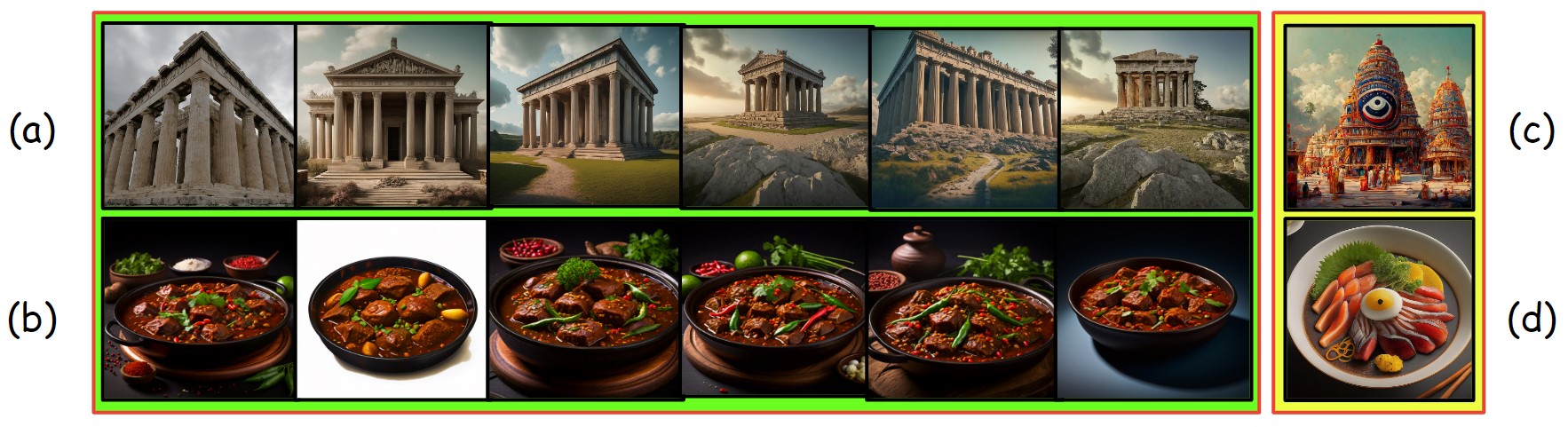} % Replace 'example-image' with 
    
    \vspace{-0.2\baselineskip}
    \caption{Images from a SOTA T2I model demonstrating its lack of \textcolor{green!50!black}{cultural diversity}: (a) and (b) and \textcolor{orange!60!yellow}{cultural awareness}: (c) and (d). 
    %To show lack of cultural diversity we sample multiple images with the same prompt. 
    (a) Images for  "\emph{High definition photo of a monument}" lack architectural and global diversity. (b) Images for "\emph{Image of Nigerian dish}" lack the rich diversity in Nigerian cuisine. (c) "\emph{Image of Jagannath Temple from India}" produces an incorrect depiction of the temple. (d) "\emph{Image of Japanese dish Kabayaki}" produces an incorrect and cartoonized photo. 
    }
    \label{fig:exampleimage}
\end{figure*}

    %In the set of images to the left for the prompt, ``Traditional Indian Clothing.'', we see that there is a lack of regional (and gender) diversity. In the set of images to the right for the prompt, ``Famous Tourist Location in Nigeria.", we see that the images produced are unfaithful (tourist spots not necessarily from Nigeria) and unrealistic (car in front of water falls)} 
\vspace{-1.5mm}
\begin{abstract}
\vspace{-2mm}
Text-to-Image (T2I) models are being
increasingly adopted in diverse global communities where they create visual representations of their unique cultures. Current T2I benchmarks primarily focus on faithfulness, aesthetics, and realism of generated images,
% for complex prompts,
% with generic objects, 
overlooking the critical dimension of cultural competence. 
In this work, we introduce a framework to evaluate cultural competence of T2I models along two crucial dimensions: \textit{cultural awareness} and \textit{cultural diversity}, and present a scalable approach using a combination of structured knowledge bases and large language models to build a large dataset of cultural artifacts to enable this evaluation. In particular, we apply this approach to build CUBE (CUltural BEnchmark for Text-to-Image models), a first-of-its-kind benchmark to evaluate cultural competence of T2I models. CUBE covers cultural artifacts associated with 8 countries across different geo-cultural regions and along 3 concepts: cuisine, landmarks, and art. CUBE consists of 1) CUBE-1K, a set of high-quality prompts that enable the evaluation of cultural awareness, and 2) CUBE-CSpace, a larger dataset of cultural artifacts that serves as grounding to evaluate cultural diversity. We also introduce cultural diversity as a novel T2I evaluation component, leveraging quality-weighted Vendi score. Our evaluations reveal significant gaps in the cultural awareness of existing models across countries and provide valuable insights into the cultural diversity of T2I outputs for under-specified prompts. Our methodology is extendable to other cultural regions and concepts, and can facilitate the development of T2I models that better cater to the global population.\footnote{
  \faGithub \ \href{{https://github.com/google-deepmind/cube}}{\texttt{https://github.com/google-deepmind/cube}}}

%\footnote{\url{https://anonymous.4open.science/r/CUBE_T2I_Benchmark-30C1/README.md}}

\end{abstract}

% \vspace{-2mm}
\section{Introduction} \label{sec:intro}
\vspace{-2mm}

% \begin{quote}  
% \textit{"Technology is not merely a system of machines with certain functions; rather it is an expression of the social world"} 
% \end{quote}
% \begin{flushright}
% -
% \citet{Nye2006TechnologyMQ}
% \end{flushright}

\begin{comment}
%Vinod's Intro tightening plan

Para1 [Context]: T2I models have rapid advancements in capabilities. Use across the globe and in various contexts, and are increasingly becoming a form of cultural production. 

Para 2 [Problem]: While evals traditionally focus on realism etc., recent work has pointed out gaps in cultural focused evals. This is important why? When we say culture here -- -say we mean geo cultural differences

Para 3 [Core Aspects of the Problem]: Argue that when it comes to cultural competence of these models, two main aspects are cultural awareness and cultural diversity. Give examples to both, pointing to the figures, and mention the gaps in measure diversity and lack there of. 

Para 4 [Major Challenges/Gaps]: Describe the main gap in knowledge (coverage of a large set of cultural awareness artifacts. Describe the main gap in diversity --- metric itself.

Para 5 [Solution Approach]: High-level description of what we do -- how we address the problem of cultural awareness evals (i.e., using Knowledge Base to expand), and how we address the problem of diversity. 
Para 6 [Solution details]: Go one step further and explain the process, and which aspects of culture we focused on, why, etc. 

Para 7 [Contributions] Summarize the specific contributions, maybe the bulleted list.

\end{comment}

\newif\ifvpintro
\vpintrotrue
\ifvpintro
% Para1 [Context]: T2I models have rapid advancements in capabilities. Use across the globe and in various contexts, and are increasingly becoming a form of cultural production. 
Text-to-image (T2I) generative capabilities have advanced rapidly in recent years, exemplified by models such as Imagen 2 \citep{saharia2022photorealistic}, and DALLE-3 \citep{betker2023improving}. As powerful tools for creative expression and communication, they have the potential to revolutionize numerous industries such as digital arts, advertising, and education. However, their widespread adoption across the globe raises important ethical and social considerations \citep{bird2023typology,weidinger2023sociotechnical}, in particular, in ensuring that these models work well for all people across the world \citep{qadri2023ai,mim2024between}.
% 
% Para 2 [Problem]: While evals traditionally focus on realism etc., recent work has pointed out gaps in cultural focused evals. This is important why? When we say culture here -- -say we mean geo cultural differences
While early T2I model evaluations focused on photo-realism \citep{saharia2022photorealistic} and faithfulness\citep{hu2023tifa, cho2024davidsonian, huang2023t2icompbench},
recent work has demonstrated various societal biases that they reflect \citep{cho2023dall,bianchi2023easily,luccioni2024stable}. However, the predominantly mono-cultural development ecosystems of these models risks unequal representation of cultural awareness in them, potentially exacerbating existing technological inequalities \citep{prabhakaran2022cultural}.
While the term ``culture'' has a myriad  definitions across disciplines \citep{rapport2002social}, in this paper we focus on cultures formed within societies demarcated geographically through national boundaries (similar to \cite{li2024culturegen}), rather than cultures defined through organizational or other socio-demographic categories. This focus stems from our aim to assess global disparities in the capabilities of T2I models.
Such disparities are shown to perpetuate harmful stereotypes about cultures \citep{jha2024visage,basu2023inspecting}, as well as cause the erasure and suppression of sub- and co-cultures \citep{qadri2023ai}, and limit their utility across geo-cultural contexts \citep{mim2024between}. While recent work has focussed on biases and stereotypes these models propagate \citep{jha2024visage,basu2023inspecting}, not much work has looked into how competent these models are in capturing the richness and diversity of various cultures. 

% Para 3 [Core Aspects of the Problem]: Argue that when it comes to cultural competence of these models, two main aspects are cultural awareness and cultural diversity. Give examples to both, pointing to the figures, and mention the gaps in measure diversity and lack there of. 

Gaps in cultural competence may manifest primarily along two aspects of model generations: (i) \textit{cultural awareness}: failure to recognize or generate the breadth of concepts/artifacts associated with a culture (Figure 1(c) and 1(d)), and (ii) \textit{cultural diversity}: the tendency to adopt an oversimplified and homogenized view of a culture that associates (and generates) a narrow set of concepts/artifacts within that culture (Figure 1(b)) or across global cultures (Figure 1(a)).
% 
% Para 4 [Major Challenges/Gaps]: Describe the main gap in knowledge (coverage of a large set of cultural awareness artifacts. Describe the main gap in diversity --- metric itself.
While the lack of cultural awareness in text to image models has been documented before \citep{hutchinson2022underspecification,ventura2023navigating}, a major challenge in effectively assessing it at scale is the lack of resources that have a broad representation of cultural artifacts. Similarly, while dataset diversity has also been identified as an important part of the data-centric AI agenda~\citep{oala2023dmlr} and has been investigated for text \citep{Chung2023IncreasingDW} and image modalities \citep{srinivasan2024generalized, dunlap2023diversify}, there has been limited focus on diversity of model generations \citep{lahoti-etal-2023-improving}, especially for T2I models. While works studying diversity of image generations focus on visual similarity \citep{hall2024dig, zameshina2023diverse}, we study the diversity of generated cultural artifacts (aka \textit{cultural diversity}).
% \vinod{Nithish, could you find some citations to wokr on diversity in T2I models that Caroline mentioned about, and add here (and update the previous sentence accordingly?)} \nithish{I have added 2 papers, Lahoti et a - LLM diversity, Srinivasan et al. - Image dataset diversity, couldn't find any work on T2I diversity}. 
% In the literature there exists several choices to demarcate culture \citep{alkhamissi2024investigating, naous2023having}, we chose to go with the idea of the geo-cultural demarcation (by country) similar to \citep{li2024culturegen}. \vinod{Add a sentence to say culture is hard to define and we go with the idea of geo-cultural differences, at the country level} \nithish{I took a shot, pls modify}. 

% Para 5 [Solution Approach]: High-level description of what we do -- how we address the problem of cultural awareness evals (i.e., using Knowledge Base to expand), and how we address the problem of diversity. 
% Para 6 [Solution details]: Go one step further and explain the process, and which aspects of culture we focused on, why, etc. 
In this paper, we present CUBE: \textbf{CU}ltural \textbf{BE}nchmark, a first-of-its-kind benchmark designed to facilitate the evaluation of cultural competence of T2I models along two axes: {cultural awareness} and {cultural diversity}. 
%We build this benchmark at the country-level (in line with other recent work \citep{jha2024visage,li2024culturegen}),  encompassing eight countries --- Brazil, France, India, Italy, Japan, Nigeria, Turkey, United States
% \vinod{list the countries} \nithish{done} 
%--- chosen from different geo-cultural regions across continents and the Global South-North divide, and representing three different concepts of cultural artifacts:  cuisine, tourist landmarks, and art --- chosen as concepts of clear visual elements and hence of importance to T2I models.
% We build this benchmark at the country-level (in line with other recent work \citep{jha2024visage,li2024culturegen}), encompassing eight countries, and representing three different concepts of cultural artifacts that have clear visual elements and hence of importance to T2I models.
We build this benchmark at the country level (in line with recent works \citep{jha2024visage,li2024culturegen}), encompassing eight countries and representing three different concepts of cultural artifacts chosen as concepts of clear visual elements, and hence of importance to T2I models. We employed a large-scale extraction strategy that leverages a Knowledge Graph (KG) augmented with a Large Language Model (LLM) to build a broad-coverage compilation of country-specific artifacts to ground our evaluation. CUBE consists of a) \textbf{CUBE-1K} - a carefully curated subset of 1000 artifacts, made into prompts that enable evaluation of cultural awareness \citep{nguyen2023extracting}, and b) \textbf{CUBE-CSpace} - a collection of $\sim$300K cultural artifacts spanning the 8 countries and 3 concepts we consider, with a potential to be scaled to other concepts and countries. Furthermore, we introduce cultural diversity (CD) as a new evaluation component for T2I models, adapting the quality-weighted Vendi score \citep{nguyen2024quality}.We detail the CUBE curation process in Section \ref{cube_3}, cultural awareness evaluation in Section \ref{awareness_4} and cultural diversity evaluation in Section \ref{diversity_5}. To summarize, our main contributions are:

% Para 7 [Contributions] Summarize the specific contributions, maybe the bulleted list.

\begin{itemize}[leftmargin=*,noitemsep,nosep]
\item A new T2I CUltural BEnchmark (\textbf{CUBE}), that assesses the cultural competence of T2I models along two key dimensions: (1) Cultural Awareness and (2) Cultural Diversity. We curate a dataset of 300K cultural artifacts spanning three concepts with a potential to be scaled to other concepts.

% . We curate a dataset of 300K cultural artifacts spanning three concepts that can be scaled to other concepts.
\item An extensive human evaluation measuring the faithfulness and realism of T2I-generated cultural artifacts across eight countries and three concepts, revealing substantial gaps in cultural awareness.
\item A novel T2I evaluation component leveraging the quality-weighted Vendi score that satisfies the desirable properties to assess cultural diversity in T2I models.
% This allows us to draw useful insights into models' intrinsic cultural diversity across different seed generations. 
\end{itemize}

\else

Text-to-image (T2I) models have demonstrated rapid advancements in the quality and photorealism of generated images, exemplified by models such as Stable Diffusion-XL \citep{podell2023sdxl}, Imagen \citep{saharia2022photorealistic} and DALLE-3 \citep{betker2023improving} . While substantial progress has been made in image realism \citep{saharia2022photorealistic} and the accuracy with which generated images align with input prompts \citep{hu2023tifa, cho2024davidsonian, huang2023t2icompbench}, limited emphasis has been placed on evaluating the diversity of model outputs. Studies addressing output diversity have been conducted for text-based modalities; \citep{lahoti-etal-2023-improving} formalize diversity measures for generative LLMs focusing on demographic representations.  However, this remains a vastly under-explored area for multimodal models. As images convey more nuanced information than textual responses, investigating diversity in text-to-image systems holds significant promise.

As generative AI models grow towards becoming pivotal and pervasive technology, with broad global reach, it becomes increasingly important that they work well for all people across the world. Disproportionate abilities or performance for some communities could exacerbate prevalent technological inequalities, with untold long-term socio-economic consequences. The predominantly westernized development of generative AI systems underlines the risks \citep{eiras2024near} that not all cultures may enjoy similar levels of cultural awareness, nuanced depiction, etc. A notable symptom of intra-cultural under-performance is in how well these systems capture the underlying natural richness and diversity within a culture, when prompted with an under-specified concept - yet without misrepresentation. Instances of under- or over-representation of cultural entities - especially the homogenization of majority or dominant viewpoints - could perpetuate harmful biases and stereotypes. For example, if a T2I model consistently produces images of hamburgers in response to the prompt "American cuisine" (as depicted in Fig. 2), it may reinforce the stereotype that "Americans only eat burgers". This could also lead to erasure & suppression of sub- & co-cultures, or alienation of certain segments of the population. Diverse output sets that remain faithful to the input prompts, but enriching towards multiple possible equivalent intents, are perhaps more inclusive and equitable. 

While dataset diversity has been identified as an important part of the data-centric AI agenda~\citep{oala2023dmlr} and has been investigated for text \citep{Chung2023IncreasingDW} and image modalities \citep{dunlap2023diversify}, there has been limited focus on diversity of model generations, especially for T2I models. We propose focusing on culture as a crucial area for research on T2I diversity. By providing cultural prompts that are under-specified, we can analyze the cultural diversity of generated outputs. For instance, in Fig \ref{fig:exampleimage}, the prompt "Traditional Indian clothing" yields results lacking regional diversity (note also lack of gender diversity, which is not the focus in this work). Conversely, artificially injecting diversity into T2I outputs can result in inaccurate generations. As seen in Fig \ref{fig:exampleimage}, the prompt "Famous tourist locations in Nigeria" produces images that are unfaithful etc.,. While diversity in these instances is not strictly cultural, our primary focus is on cultural diversity. More concretely,  we are interested in the diversity of cultural artifacts in the generated outputs and do not look into other forms of visual or aesthetic diversity . For this work, we limit our scope to non-human artifacts. Although our ultimate goal is to ensure culturally diverse outputs that reflect the richness and diversity of a culture, a crucial first step is to reliably measure the cultural diversity of T2I models. This serves as a proxy for analyzing a model's cultural awareness. We further examine the ability of T2I models to generate images of cultural artifacts from specific concepts within a culture.

We present CUBE, a first-of-its-kind benchmark designed to facilitate the cultural evaluation of text-to-image (T2I) models. As a first step, we define culture in terms of cultural concepts, demarcated by geography. For this study, we select 8 diverse countries. This framing is chosen due to the relative ease of demarcating culture by country \citep{li2024culturegen}, compared to alternatives such as language used in literature, particularly given our chosen concepts (e.g., tourist attractions). Following established convention \citep{li2024culturellm}, we use the term "culture" interchangeably with these countries throughout this work. We begin by analyzing three cultural concepts for each culture: cuisine, tourist landmarks, and art. These concepts constitute a subset of those otulined in \citep{Halpern1955TheDE}.  

For each country and concept, we define specific cultural concepts. For instance, within the concept of cuisine, cultural concepts might include "American cuisine". Similarly, within the concept of landmarks, examples could be "Indian tourist landmarks". Each cultural concept thus pertains to a specific country (representing a culture) and a particular concept. Next, we compile a collection of cultural artifacts for each cultural concept to construct its corresponding concept space. This process, termed "grounding the concept space," entails associating each abstract concept with tangible real-world entities. For example, the concept space for "American food" might include various dishes like hamburgers, pizza etc,. To achieve this grounding, we employ a large-scale extraction strategy leveraging a Knowledge Graph (KG) augmented with a large language model (LLM). This approach results in a set of artifacts for each cultural concept, establishing a comprehensive framework for evaluating the cultural awareness of T2I models.

\begin{table*}[t]
\centering
\caption{Overview of text-to-image Benchmarks} 
\begin{tabular}{@{}llccc@{}} % Adjust column widths as needed
\toprule
\textbf{Benchmark} & \textbf{Skill} & \multicolumn{3}{c}{\textbf{Evaluation Aspect}} \\ 
\cmidrule(lr){3-5}
 &  & \textbf{faithfulness} & \textbf{Realism} & \textbf{Diversity} \\ 
\midrule
DrawBench & Spatial \& Object & \checkmark & \checkmark & \ding{55} \\ 
% ABC-6K & Composition (color) & \checkmark & \checkmark & \ding{55} \\ 
CC500 & Composition (color) & \checkmark & \checkmark & \ding{55} \\ 
T2I-CompBench & Composition & \checkmark & \ding{55} & \ding{55} \\ 
Tifa160 & Spatial & \checkmark & \ding{55} & \ding{55}\\ 
DSG1k & Spatial & \checkmark & \ding{55} & \ding{55} \\ 
GenEval & Object & \checkmark & \ding{55} & \ding{55} \\ 
GenAIBench & Spatial & \checkmark & \ding{55} & \ding{55} \\ 
\midrule
\libertineWord{CUBE} & Cultural & \checkmark & \checkmark & \checkmark  \\
\bottomrule
\end{tabular}
\label{tab:overview} % Add a label if you want to reference the table
\end{table*}

The proposed benchmark, with its prompts, cultural framework, and diversity metrics, offers a new dimension for T2I evaluation. Table \ref{tab:overview} contrasts existing T2I evaluation benchmarks with ours - which we believe is a timely contribution to track and foster culturally inclusive T2I technology. To summarize, our main contributions are:

\begin{itemize}
\item A new T2I benchmark, \textbf{CUBE}, that assesses the cultural competence of T2I models along two key dimensions: (1) Cultural Awareness and (2) Cultural Diversity.
\item A scalable framework for extracting and refining concept spaces from Wikidata, leveraging large language models (LLMs) to address potential gaps. We curate a dataset of 300K cultural artifacts spanning three concepts that can be scaled to other concepts.
\item Extensive human evaluation measuring the faithfulness and realism of T2I-generated cultural artifacts across eight countries and three concepts, revealing substantial gaps in existing models' cultural awareness.
\item Useful insights into models' intrinsic cultural diversity across different seed generations using the quality-weighted Vendi score.
\end{itemize}

\fi

% \begin{itemize}
% \item A new T2I benchmark, \textbf{CUBE}, that evaluates T2I models on its cultural competence along two areas: 1) Cultural Awareness and 2) Cultural Diversity. 

% \item A scalable strategy to extract concept spaces from Wikidata with LLMs for refining and filling the gaps. We collect a total of 300K cultural artifacts for 3 concepts that can be scaled to other concepts

% \item Extensive human evaluation measuring the faithfulness and realism of T2I generations of cultural artifacts across 8 countries and 3 concepts revealing significant gaps in about model's cultural awareness.

% \item A novel metric to evaluate cultural diversity of T2I generations for under-specified prompts adapting Vendi score and useful insights about model's instrinsic diversity across different seed generations 
% \end{itemize}

\newif\ifvprelated

\vprelatedtrue

\ifvprelated

\vspace{-3mm}
\section{Related Work}
\vspace{-2mm}

\begin{comment}
\begin{figure*}[t]
    \centering
    % \includegraphics[width=\textwidth]{images/intro image.png} % Replace 'example-image' with your image file's name
    \includegraphics[width=0.98\textwidth]{images/main-figure-3-v3.pdf}
    
    \vspace{-0.2\baselineskip}
    \caption{\textbf{Framework for evaluating cultural competence in T2I models}. The top subfigure shows the definition of \textit{cultural concepts} and the extraction of \textit{concept space} from KB + LLM. The bottom shows example task prompts to probe the model for cultural awareness and cultural diversity.}
    \label{tab:task_prompts}
\end{figure*}

\end{comment}

In our discussion of related work, we focus on T2I evaluation and culture in large models.

% T2I evaluations in general
% T2I model evaluation has become a significant area of research in recent years. 
\paragraph{T2I Evaluation.} Inception Score \citep{salimans2016improved} and  Frechet Inception Distance \citep{heusel2018gans} focus on similarity of generated images to real ones, also called realism. Metrics like DSG \citep{cho2024davidsonian} and VQAScore \citep{lin2024evaluating} measure the prompt-image alignment, also called faithfulness.  Other metrics such as ImageReward \citep{xu2023imagereward}, PickScore \citep{kirstain2023pickapic}, and HPSv2 \citep{wu2023human} fine-tune vision-language models on human ratings to better align with human preferences. There have been recent works on bias and fairness evaluation \citep{feng2022towards,naik2023social,zhang2023iti, jha2024visage} of T2I models. There have also been efforts to build comprehensive evaluation benchmarks aimed at tracking the progress of model capabilities over time, focusing on tasks such as
realism, text faithfulness, and compositional abilities. These benchmarks, such as DrawBench \citep{saharia2022photorealistic}, CC500 \citep{feng2023trainingfree},
T2I-CompBench \citep{huang2023t2icompbench}, 
TIFA v1.0 \citep{hu2023tifa}, 
DSG-1k \citep{cho2024davidsonian}, 
GenEval \citep{ghosh2023geneval}, and 
GenAIBench \citep{lin2024evaluating} employ diverse prompts and metrics to assess factors such as image-text coherence, perceptual quality, attribute binding, faithfulness, semantic competence, and compositionality, to list a few.

% Evals on culture in NLP
% Recent studies in the 
\paragraph{Culture in Language Technologies.} NLP researchers have long argued for the need for cross-cultural awareness in language technologies \citep{hovy2016social,hershcovich-etal-2022-challenges}, and built
datasets to assess cultural biases in language technologies \citep{jha2023seegull,naous2023having,seth2024dosa}.
% SeeGULL \citep{jha2023seegull}, CAMeL  \citep{naous2023having} and DOSA \citep{seth2024dosa} datasets reveal biases in LLMs dominant cultures in different contexts.
There have also been efforts to identify cultural keywords across languages \citep{lim2024computational}, extract cultural commonsense knowledge \citep{nguyen2023extracting}, as well as to generate culture-conditioned content  \citep{li2024culture}. Along those lines, CultureLLM \citep{li2024culturellm} proposes generating training data using the World Value Survey for semantic data augmentation to integrate cultural differences into large language models.

% \begin{wraptable}{r}{0.65\textwidth}
% % \begin{table*}[t]
% \centering
% \small
% \begin{tabular}{@{}lccc@{}} % Adjust column widths as needed
% \toprule
% \textbf{Benchmark}  (\textbf{Skill}) & \multicolumn{3}{c}{\textbf{Evaluation Aspect}} \\ 
% \cmidrule(l){2-4}
%  &   \textbf{faithfulness} & \textbf{Realsim} & \textbf{Diversity} \\ 
% \midrule
% DrawBench (Spatial \& Object ) & \checkmark & \checkmark & \ding{55} \\ 
% % ABC-6K (Composition (color)) & \checkmark & \checkmark & \ding{55} \\ 
% CC500 (Composition (color)) & \checkmark & \checkmark & \ding{55} \\ 
% T2I-CompBench (Composition) & \checkmark & \ding{55} & \ding{55} \\ 
% Tifa160 (Spatial) & \checkmark & \ding{55} & \ding{55}\\ 
% DSG1k (Spatial) & \checkmark & \ding{55} & \ding{55} \\ 
% GenEval (Object) & \checkmark & \ding{55} & \ding{55} \\ 
% GenAIBench (Spatial) & \checkmark & \ding{55} & \ding{55} \\ 
% \midrule
% \libertineWord{CUBE} (Cultural) & \checkmark & \checkmark & \checkmark  \\
% \bottomrule
% \end{tabular}
% \caption{Overview of text-to-image Benchmarks \label{tab:overview}} 
%  % Add a label if you want to reference the table
% % \end{table*}
% \end{wraptable}
% 
% Evals on culture in Vision/T2I
\paragraph{Culture in Vision.} Efforts to understand cultural competence in computer vision technologies are relatively more recent and limited. \cite{basu2023inspecting}
explored the geographical representation of under-specified prompts and found that most of them default to United States or India. Dig In \citep{hall2024dig} evaluates disparity in geographical diversities of household objects.
SCoFT \citep{liu2024scoft} enhances cultural fairness using the cross-cultural awareness Benchmark (CCUB). Recent work also shows that cultural and linguistic diversity in datasets enriches semantic understanding and helps address cultural dimensions in text-to-image models \citep{ye2023cultural, ventura2023navigating}. Proposals for more inclusive model design and dataset development have been made to address cultural stereotypes and Western-centric biases, to better represent global cultural diversity \citep{bianchi2023easily, liu2021visually}.
Our work contributes to this line of work, where we introduce a large benchmark dataset and associated metrics to assess cultural competence along cultural awareness and cultural diversity in T2I models.

\begin{figure*}[t]
    \centering
    \includegraphics[width=0.98\textwidth]{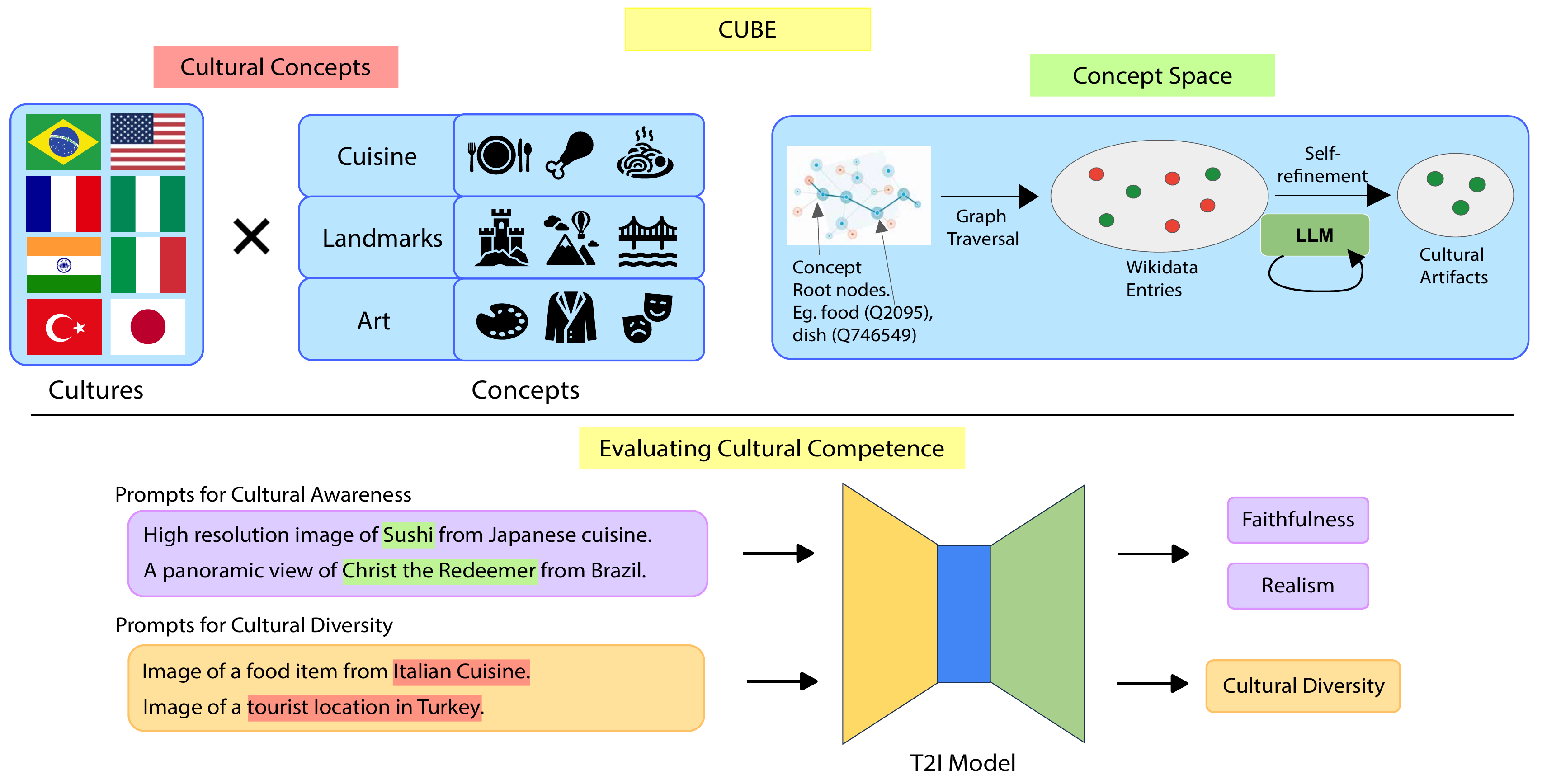}
    
    \vspace{-0.2\baselineskip}
    \caption{\textbf{Framework for evaluating cultural competence in T2I models}. The top subfigure shows the definition of \textit{cultural concepts} and the extraction of \textit{concept space} from KB + LLM. The bottom shows example task prompts to probe the model for cultural awareness and cultural diversity.}
    \label{tab:task_prompts}
\end{figure*}

\begin{table*}[t]
\centering
\caption{ \textbf{Overview of text-to-image benchmarks}. Existing benchmarks focus only on faithfulness and realism as evaluation aspects and overlook the cultural skill. CUBE is the first T2I benchmark that evaluates cultural competence while introducing diversity as an evaluation aspect.  } 
\begin{tabular}{@{}llccc@{}} % Adjust column widths as needed
\toprule
\textbf{Benchmark} & \textbf{Skill} & \multicolumn{3}{c}{\textbf{Evaluation Aspect}} \\ 
\cmidrule(lr){3-5}
 &  & \textbf{Faithfulness} & \textbf{Realism} & \textbf{Diversity} \\ 
\midrule
DrawBench & Spatial \& Object & \checkmark & \checkmark & \ding{55} \\ 
% ABC-6K & Composition (color) & \checkmark & \checkmark & \ding{55} \\ 
CC500 & Composition (color) & \checkmark & \checkmark & \ding{55} \\ 
T2I-CompBench & Composition & \checkmark & \ding{55} & \ding{55} \\ 
Tifa160 & Spatial & \checkmark & \ding{55} & \ding{55}\\ 
DSG1k & Spatial & \checkmark & \ding{55} & \ding{55} \\ 
GenEval & Object & \checkmark & \ding{55} & \ding{55} \\ 
GenAIBench & Spatial & \checkmark & \ding{55} & \ding{55} \\ 
\midrule
\libertineWord{CUBE} & Cultural & \checkmark & \checkmark & \checkmark  \\
\bottomrule
\end{tabular}
\label{tab:overview} % Add a label if you want to reference the table
\end{table*}

\else

\section{Related Work}

\subsection{Evaluation of text-to-image models}

\paragraph{Automated Metrics.} Traditional T2I evaluation metrics like Inception Score (IS) \citep{salimans2016improved} and Frechet Inception Distance (FID) \citep{heusel2018gans} were proposed to capture the similarity of generated images to real ones, also called as the realism of the synthesised images. CMMD \citep{jayasumana2024rethinking} is an improvement over the FID metric for realism. More recently, several works have looked into the faithfulness of T2I models. The assessment of faithfulness in T2I models has also gained attention, with several metrics emerging across different categories. Embedding-based metrics such as CLIPScore \citep{hessel2022clipscore} and ALIGNScore \cite{zha2023alignscore} examine the semantic faithfulness between text and images. VQA-based metrics, including TIFA \citep{hu2023tifa}, DSG \citep{cho2024davidsonian}, and VQAScore \citep{lin2024evaluating}, leverage visual question answering frameworks to assess  faithfulness. Captioning-based metrics like LLMScore \citep{lu2023llmscore} and VIEScore \citep{ku2023viescore} focus on how well captions generated from images correspond to the input text. Additionally, innovative approaches like VPEval \citep{cho2023visual} and ViperGPT \citep{suris2023vipergpt} use visual programming for evaluating complex prompts that require reasoning. Other metrics such as ImageReward \citep{xu2023imagereward}, PickScore \citep{kirstain2023pickapic}, and HPSv2 \citep{wu2023human} fine-tune vision-language models on human ratings to better align with human preferences in T2I generation.

\paragraph{Benchmarks.} Many T2I evaluation benchmarks have been proposed recentl applicable across a variety of image generation tasks. ImagenHub dataset \citep{ku2024imagenhub} consists of 7 subtasks for a variety of text-guided image generation tasks. DrawBench \citep{saharia2022photorealistic} released alongside the Imagen consists of 200 prompts for evaluation of T2I models ABC-6K and CC500 \citep{feng2023trainingfree} benchmarks evaluate attribute binding for T2I models with a main focus on color attributes. T2I-CompBench \citep{huang2023t2icompbench} builds on the compositional T2I benchmarks introducing additional attributes such as shape and texture.  Pick-a-pick \citep{kirstain2023pickapic} is a large dataset consisting of text prompts and human preferences. TIFA v1.0 \citep{hu2023tifa} consists of 4k diverse text prompts along with questions to evaluate faithfulness of T2I models. DSG-1k \citep{cho2024davidsonian} consists of 1060 prompts with fine-grained semantic categories. The above benchmarks consist of diverse prompts and metrics designed to evaluate things like image-text coherence and perceptive quality. There has been no benchmark probing the cultural awareness of T2I models. GenEval \citep{ghosh2023geneval} and GenAIBench \citep{lin2024evaluating} are more recent T2I benchmarks. Holistic Evaluation of Text-To-Image Models (HEIM) \citep{lee2023holistic} discusses a series of 12 evaluation aspects including toxicity, bias and aesthetics.  Many of the evaluation aspects are not directly relevant to the cultural concept.

\subsection{Culture in Generative Models}

\paragraph{Large Language Models.} Recent studies in the NLP have emphasized the need for linguistic diversity and cross-cultural awareness in language models. The importance of multilingual capabilities \citep{joshi2020state} and sensitivity to cultural contexts \citep{hershcovich-etal-2022-challenges} to support diverse user bases has been shown to be crucial for developing inclusive and globally applicable language technologies. CAMeL  \citep{naous2023having} and DOSA \citep{seth2024dosa} benchmark datasets reveal biases in LLMs towards Western and dominant Indian cultures, respectively, to measure cultural bias. Often cultural keywords or markers are used for identifying the culture related to which an LLM generated response adheres to. These include, computational methods for identifying cultural keywords across languages \citep{lim2024computational}, extracting cultural commonsense knowledge \citep{nguyen2023extracting} to enhance a models' socio-cultural awareness, as well as having language models generate culture-conditioned content by recognizing linguistic markers that distinguish marginalized cultures \citep{li2024culture}. CultureLLM \citep{li2024culturellm} proposes generating training data using the World Value Survey for semantic data augmentation to integrate cultural differences into LLMs.

\paragraph{Vision Language Models.} The focus on cultural biases in LLMs extends to Vision Language Models, where recent efforts aim to quantify and mitigate these biases. SCoFT \citep{liu2024scoft} is a technique, which enhances cultural fairness in image generation using the Cross-cultural awareness Benchmark (CCUB) and a Self-Contrastive Fine-Tuning method. It has been shown that cultural and linguistic diversity in datasets enriches semantic understanding and helps address cultural dimensions in text-to-image models \citep{ye2023cultural, ventura2023navigating}. Proposals for more inclusive model design and dataset development have been made to address the challenges of demographic stereotypes and Western-centric biases, pushing for systems that better represent global cultural diversity \citep{bianchi2023easily, liu2021visually}.

\subsection{Bias and Diversity in Text-to-image models}
Biases in Text-to-Image models have been studied in various works focusing mostly on the axes of gender and skin tone (or race) \citep{feng2022towards}. Iti-gen \citep{zhang2023iti} introduces a novel method for inclusive text-to-image generation, utilizing reference images to guide prompt learning. ViSAGe \citep{jha2024visage} presents a global-scale analysis of stereotypes and \cite{naik2023social} analyse social biases in text-to-image models. There have also been several bias mitigation strategies for T2I models \citep{wan2024survey}. Studies have explored the geographical representation of under-specified prompts \citep{basu2023inspecting} and found most of them default to United States or India, but their evaluation aspect is limited to text-image faithfulness and photo-realism. Dig In \citep{hall2024dig} evaluates disparity in geographical diversities of household objects. Diverse Diffusion \citep{zameshina2023diverse} proposes diversifying T2I generations by finding vectors in latent space distant from each other. However, both these approaches focus on visual diversity based on aspects such as color, aesthetics and background using metrics such as LPIPS and coverage, as opposed to the concept or cultural diversity we describe in this paper.

\fi

\vspace{-2mm}
\section{Construction of \libertineWord{CUBE}}
\label{cube_3}

\newif\ifupCUBE

\upCUBEtrue

\ifupCUBE

Our benchmark aspires to enable reliable, trustworthy, and tangible measurement of text-to-image generative models for two distinct yet complementary behaviors: \textit{cultural awareness} (i.e., the model’s ability to reliably and accurately portray objects associated with a particular culture), and \textit{cultural diversity} (i.e., the model’s ability to suppress oversimplified stereotypical depiction for an underspecified input that references a specific culture). One of the core prerequisites to meaningfully evaluate these aspects of cultural competence is a broad-coverage repository of cultural artifacts to ground such an evaluation.
Inspired by previous work \citep{jha2024visage, li2024culturegen}, we focus on \textit{geo-cultures} (realized through the lens of national identity) to build such a repository, potentially extendable to other ways of categorizing culture, such as regions, religions, races, etc. We select eight countries from different geo-cultural regions across continents and the Global South-North divide: Brazil, France, India, Italy, Japan, Nigeria, Turkey, and USA. 
While we acknowledge that this list of countries is necessarily incomplete, and may result in a biased global sampling, future iterations of this work could include a wider range of countries for a more comprehensive evaluation.
% despite our best efforts to the contrary. 

Additionally, we focus on distinctive artifacts, i.e., cultural aspects that reference singular real objects with clear visual elements which are commonly held as belonging to a specific country – as opposed to cultural manifestations that are not visualizable (e.g. speech accents) or multifarious (e.g. complex scenes, or unique inter-object relationships). The three artifact categories ("concepts") included here are \textit{landmarks} (prominent and recognizable structures such as monuments and buildings, located in specific countries), \textit{art} (clothing and regional garments or traditional regalia, performance arts, and style of painting, associated at possibly a specific time in history), and \textit{cuisine} (specific dishes and culinary ingredients that are commonly associated with certain countries). In practice, for the art and cuisine categories, we additionally consider "country of origin" as a strong indicator of national association, acknowledging that there may be other factors.

% \begin{wrapfigure}{r}{0.6\textwidth}
%     \includegraphics[width=0.58\textwidth]{images/cult_img.png}
%     \caption{Culture Ontology (top) and examples (bottom) \utsav{Replace this image with one that shows both CUBE Collection and 1K}}
%     \label{fig:culture_ontology}
% \end{wrapfigure} 

Finally, for each country-concept combination, we aim to construct grounding "concept spaces", which leads to a collection of $\sim$300K cultural artifacts, which we call \textbf{CUBE-CSpace}. This is an extensive compilation of concept space instances, also intended to be used as grounding for diversity evaluation. From this, we create  \textbf{CUBE-1K}: a much smaller, curated set of the 1000 artifacts across the 8 countries and 3 concepts - selected for relevance and popularity, intended to be used for testing cultural awareness. The country and concept wise split of CUBE-1K is presented in Table \ref{tab:CUBE-1K-stats}. In order to build CUBE, we adopt a Knowledge-Base (KB)-augmented LLM approach wherein we use graph-traversal on a pre-existing KB to extract a broad-coverage set of candidate cultural artifacts, followed by a self-critiquing LLM step to iteratively refine the repository.

% \subsection{Compiling concept spaces}

\vspace{-2mm}
\subsection{CUBE-CSpace} 

\begin{wraptable}{r}{0.60\textwidth}
\label{alg:algo}
% \begin{table}[h]
\centering
\footnotesize
\begin{tabular}{l}
\hline
\textbf{Algorithm 1:} Cultural Artifact Extraction from Wikidata \\ \hline
\begin{tabular}[t]{l}
\textbf{Input:} Set of root nodes $R$, Maximum hops $H$ \\
\textbf{Output:} Set of cultural artifacts $A$ \\
\end{tabular} \\ \hline
\begin{tabular}[t]{l}
1:  $A \gets \emptyset$, $h \gets 0$ \\
2:  \textbf{while} $h < H$ \textbf{do} \\
3:  \hspace{0.5cm} $R_{new} \gets \emptyset$  \\
4:  \hspace{0.5cm} \textbf{for} $r \in R$ \textbf{do} \\
5:  \hspace{1cm}    $C \gets$ Children of $r$ along nodes (P31) or (P279) \\
6:  \hspace{1cm}    \textbf{for} $c \in C$ \textbf{do} \\
7:  \hspace{1.5cm}       \textbf{if} $c$ has property (P495) or (P17) \textbf{then} \\
8:  \hspace{2cm}          $A \gets A \cup \{c\}$ \\
9:  \hspace{1.5cm}       \textbf{else} \\
10: \hspace{2cm}          $R_{new} \gets R_{new} \cup \{c\}$ \\
11: \hspace{1.5cm}    \textbf{end for} \\
12: \hspace{1cm} \textbf{end for} \\
13: \hspace{0.5cm} $R \gets R_{new}$  \\
14: \hspace{0.5cm} $h \gets h + 1$ \\
15: \textbf{end while} \\
16: \textbf{return} $A$\\
\end{tabular} \\
\hline
\end{tabular}
% \end{table}
\end{wraptable}

We use WikiData \citep{10.1145/2629489} as the KB to extract cultural artifacts, as it is the world's largest publicly available knowledge base, with each entry intended to be supported by authoritative sources of information. We use the SLING framework\footnote{https://github.com/google/sling} to traverse the WikiData dump of April 2024, by first manually identifying \textit{root nodes} (see Table \ref{tab:wiki_nodes}), a small seed set of manually selected nodes that represent the concept in question. For example, the node 'dish' (WikiID: Q746549) 
is identified as a root node for the concept 'cuisine'. 
% A detailed list of the Wikidata root nodes for the different concepts is included in Table \ref{tab:wiki_nodes}. 
We then look for child nodes that lie along the 'instance of' (P31) 
and 'subclass of' (P279) 
edges; e.g. 'Biriyani',(Q271555), 
a popular dish from India, is a child node of 'dish' along the 'instance of' edge.
% P31. 
The child nodes that have the 'country-of-origin' (P495) 
 or the 'country' (P17) are extracted at the iteration. We recursively traverse the remaining nodes along the same edge classes in search of child nodes that satisfy these properties. For example 'bread' (Q7802) is a child of 'dish'; since it is a generic food item, it doesn't have the 'country-of-origin' (P495) property. However, 'Filone' (Q5449200) is a child of 'bread' and has 'country-of-origin' (P495) as Italy, which would be extracted at the step. We outline the extraction process in Algorithm 1. In practise, we iterated for H=4 hops and have detailed considerations in Appendix \ref{tech_details}.

\begin{comment}
\begin{table}[ht]
\centering
\footnotesize
\begin{tabular}{l}
\hline
\textbf{Algorithm 1:} Cultural Artifact Extraction from Wikidata \\ \hline
\begin{tabular}[t]{l}
\textbf{Input:} Set of root nodes $R$, Maximum hops $H$ \\
\textbf{Output:} Set of cultural artifacts $A$ \\
\end{tabular} \\ \hline
\begin{tabular}[t]{l}
1:  $A \gets \emptyset$, $h \gets 0$ \\
2:  \textbf{while} $h < H$ \textbf{do} \\
3:  \hspace{0.5cm} $R_{new} \gets \emptyset$  \\
4:  \hspace{0.5cm} \textbf{for} $r \in R$ \textbf{do} \\
5:  \hspace{1cm}    $C \gets$ Children of $r$ along nodes (P31) or (P279) \\
6:  \hspace{1cm}    \textbf{for} $c \in C$ \textbf{do} \\
7:  \hspace{1.5cm}       \textbf{if} $c$ has property (P495) or (P17) \textbf{then} \\
8:  \hspace{2cm}          $A \gets A \cup \{c\}$ \\
9:  \hspace{1.5cm}       \textbf{else} \\
10: \hspace{2cm}          $R_{new} \gets R_{new} \cup \{c\}$ \\
11: \hspace{1.5cm}    \textbf{end for} \\
12: \hspace{1cm} \textbf{end for} \\
13: \hspace{0.5cm} $R \gets R_{new}$  \\
14: \hspace{0.5cm} $h \gets h + 1$ \\
15: \textbf{end while} \\
16: \textbf{return} $A$\\
\end{tabular} \\
\hline
\end{tabular}
\end{table}

\end{comment}

\paragraph{Refinement.} The above KB extraction process results in $\sim$500K collection of WikiData nodes, which is expected to have missing and inconsistent entries, owing to the noisy nature of WikiData \citep{kannen-etal-2023-best}. We use GPT4-Turbo to filter out cultural artifacts that may not necessarily belong to a concept space, taking inspiration from existing self refinement \citep{madaan2023selfrefine} and self critiquing \citep{lahoti-etal-2023-improving} techniques. Once we filter out the erroneous artifacts, we prompt GPT-4 to fill out popular missing artifacts from the cultural concept, similar to the diversity expansion application in \citep{lahoti-etal-2023-improving}. This filtering and completion process brings down the count to $\sim$300K entries, which forms the \textbf{CUBE-CSpace}. Table \ref{example_cspace} presents some examples cultural artifacts extracted by this process.

\subsection{CUBE-1K }

As T2I models are primarily trained on English image-text pairs \citep{pouget2024filter}, we expect them to struggle with visualizing artifacts from non-English-speaking cultures. To this end, CUBE-1K consists of prompts focusing on widely recognized artifacts, reflecting a model's ability to capture mainstream cultural elements. The artifacts in CUBE-1K are a carefully curated subset of CUBE-CSpace. To ensure the inclusion of popular artifacts relevant to each country, we leverage the number of Google search results as a proxy for popularity.  Specifically, we employ the Google Search API, utilizing the geolocation feature ('gl' property) to tailor search results to a user located within the target country, thus capturing local popularity. We use this popularity estimate to sample artifacts for CUBE-1K. While search results serve as a useful proxy, we acknowledge they can be noisy, potentially inflated by the presence of popular keywords. Therefore, the final collection undergoes a manual verification process (detailed in Appendix \ref{tech_details}) to ensure relevance of selected artifacts. CUBE-1K consists of 1000 prompts, spanning 8 countries and 3 concepts described above. Table \ref{tab:CUBE-1K-stats} presents the distribution of artifacts across different countries in CUBE-1K.
We use prompt templates designed to probe the models for cultural awareness, along with a negative prompt \citep{hao2023optimizingpromptstexttoimagegeneration} to obtain images with desired qualities. Each prompt tests the model's ability to visualize a single artifact. The prompt templates and negative prompt are provided in the Table \ref{tab:prompt_templates} (in Appendix).

% \section{CUBE-1K: A T2I Benchmark for Evaluating Cultural Awareness}
\section{Evaluating Cultural Awareness}\label{sec:evaluating_cultural_awareness}

\label{awareness_4}
To assess the cultural awareness of text-to-image (T2I) models, we leverage prompts from the CUBE-1K dataset.   We use traditional T2I evaluation aspects like \textit{faithfulness} (adherence of the generated image to the input prompt) and \textit{realism} (similarity of the generated image to a real photograph) to measure cultural awareness. Conventionally, these are measured using automated metrics like DSG \citep{cho2024davidsonian} and FID \citep{heusel2018gans}. However, these prove insufficient for capturing the complexities of \textit{cultural} representation. Existing automated metrics are primarily trained on datasets lacking diverse cultural content and struggle to adequately assess the nuances of cultural elements. Therefore, we introduce a human annotation scheme specifically tailored to measure a model's cultural awareness along the two key dimensions: a) faithfulness and b) realism.

\begin{table}[t!]
\centering
\footnotesize % Reduce font size
\begin{tabular}{m{2cm} m{1.5cm} m{7cm}}
\toprule
\textbf{Geo-culture} & \textbf{Concept} & \textbf{Cultural Artifacts} \\ 
\midrule
Japan & Cuisine & Ramen, Soba, Sushi, Katsu sandwich  \\ 
France & Landmarks & Eiffel Tower, Mont Saint-Michel, Palace of Versailles  \\ 
India & Art/Clothing & Kurta, Lehanga Choli, Dhoti, Patola Saree \\ 
\bottomrule
\end{tabular}
\caption{Examples of cultural artifacts collected in CUBE-CSpace}
\label{example_cspace}
\end{table}

\begin{comment}

\begin{table*}[t]
\centering
\caption{Example task prompts to probe the model for cultural awareness and cultural diversity}
\label{tab:prompt_examples}
\begin{tabular}{p{0.57\textwidth} p{0.39\textwidth}}
\toprule
\textbf{Cultural Awareness} & \textbf{Cultural Diversity } \\
\midrule
\textbf{Cuisine} & \textbf{Cuisine} \\
- A high resolution image of \textcolor{blue}{Sushi} from Japanese cuisine & - Image of Nigerian clothing \\
- A high resolution image of \textcolor{blue}{Eba} from Nigerian cuisine & - Image of a dish from Japanese cuisine \\
\midrule
\textbf{Landmarks} & \textbf{Landmarks} \\
- A panoramic view of \textcolor{red}{Christ the Redeemer} in Brazil & - Image of tourist landmarks in Italy \\
- A panoramic view of \textcolor{red}{Qutub Minar} in India & - Image of a tourist spot in France\\
\midrule
%\textbf{Clothing \& Art} & \textbf{Clothing \& Art} \\
\textbf{Art} & \textbf{Art} \\
- An image of \textcolor{teal}{Bhangra} performance from India & - Image of traditional Japanese clothing \\
- An image of \textcolor{teal}{Kanzashi} from Japanese clothing & - Image of a costume from Turkey \\
\bottomrule
\end{tabular}
\end{table*}

\end{comment}

\else

We begin by presenting a cultural ontology that serves as the framework for our benchmark creation. Next, we define the specific cultural concepts under consideration and collect their corresponding concept spaces, referred to as grounding. After acquiring cultural artifacts within the concept space for each concept, we utilize GPT4-Turbo for filtering  artifacts based on Google Search results as a proxy for popularity. 

\subsection{Culture Ontology}

For our work, we consider 8 countries that are geographically diverse as shown in Table~\ref{tab:CUBE-1K-stats}. We would like to point out that we refer to "geo-culture" when we mention culture going forward and stick to calling it culture for brevity. Further, culture is composed of concepts. Concepts are essentially concepts that we consider to analyze culture. For example, "clothing" is a concept that represents culture. We refer to the intersection of culture and concept as a cultural concept, for example, "Nigerian clothing" is a cultural concept. 

While culture is a fairly broad term that could mean different things, we loosely define cultures to be demarcated by geographical boundaries similar to \citep{li2024culturegen}. More specifically, we deal with granularity of cultures at the level of countries. 
\begin{wrapfigure}{r}{0.6\textwidth}
    \includegraphics[width=0.58\textwidth]{images/cult_img.png}
    \caption{Culture Ontology (top) and examples (bottom)}
    \label{fig:culture_ontology}
\end{wrapfigure} 

Each  concept consists of real-world entities that form the concept space. For example, "Kimono", "Kebaya" and "Sari" belong to the concept space of the concept "clothing". Likewise, traditional clothing that is native to Nigeria such as "Yoruba", belongs to the concept space of the "Nigerian clothing". Figure 2 represents this framework that we define to study culture in T2I models. Note that we refer to diversity as within-cultural concept diversity unless mentioned otherwise.

\subsection{Collection of Cultural artifacts: Grounding}

In the culture ontology defined in Fig \ref{fig:culture_ontology}, a key step is to extract the concept space for a concept. Concept space refers to all the real-world entities that that  belong to cultural concept. We call this process of tying a cultural concept to real world entities that belong to it as \textbf{cultural grounding}. We first collect cultural artifacts for a cultural concept by traversing the WikiData knowledge graph and filter/clean the list using GPT-4. Subsequently, we bin the artifacts into a popularity scale (1-5) by using number of Google search results obtained as a weak estimator of the popularity of an artifact. The steps are detailed below:

We use the public WikiData knowledge base for extracting cultural artifacts. Cultural artifacts are nodes in the Wikidata that have certain desirable properties. In order to identify desirable cultural artifacts from a concept space that can potentially be far dispersed in the knowledge graph, we manually identify nodes that can represent the concept whose artifacts we are looking for, which we call root nodes. For example, the node 'dish' (WikiID: Q746549) can be a good representative node for the concept 'food'. We then look for child nodes that lie along the 'instance of' (P31) and 'subclass of' (P279) edges of the root node.  For examples 'Biriyani' (Q271555) a popular dish from India is a child node of 'dish' along the edge P31. We have presented a detail manual collection of Wikidata root nodes for the different concepts in Table \ref{tab:wiki_nodes}. It is important to note that we are looking to extract cultural artifacts that belong to a certain geo-culture or country. As a result, we look for child nodes that have the 'country-of-origin' (P495) property or the 'country' (P17) property in order for it to be extracted as a cultural artifact. These nodes also become the leaf nodes of the traversal path.  The child nodes that do not have the country property (non-leaf nodes) , serve as the root nodes for the subsequent iteration.  For example 'salad' (Q9266) is a child of 'dish', since it is a generic food item, it doesn't have the country (P17) property. However, 'Caesar salad' (Q275508) is a child of 'salad' and has 'country-of-origin' (P495) as Mexico, which is a cultural artifact at 2 hop distance from 'dish'. The detailed extraction algorithm is mentioned in Algorithm 1. We extract artifacts from upto H=4 hop distances from the initial roots nodes. We use the SLING framework \footnote{https://github.com/google/sling} for traversing the WikiData dump of April 2024. \textcolor{red} { another table with the statistics of this extraction. Add a cool example table or figure?}

% 1) nodes that are upto 4 hop children of selected root nodes we pick for a cultural concept,  2) nodes that have a 'country-of-origin' (P495) property or the 'country' (P17) property.  We manually pick the root nodes from Wikidata for each concept which is listed in Table \ref{tab:wiki_nodes}. Once we have the root nodes, we extract all the child nodes that lie along the 'instance of' (P31) and 'subclass of' (P279) edges of the root node. We do this process iteratively, where at every step, nodes that have the 'country-of-origin' (P495) property or the 'country' (P17) property are extracted as cultural artifacts (leaf nodes) corresponding the concept, and the nodes that do not have the country property (non-leaf nodes) , serve as the root nodes for the subsequent iteration. This way we iteratively traverse the KG upto 4 hop distances from the initial roots and extract nodes that serve as the artifacts in the concept space of the cultural concepts we consider. We use the SLING framework \footnote{https://github.com/google/sling} for traversing the WikiData dump of April 2024. \textcolor{red} {[to add[ ]The table below describes the root nodes from where we begin this search for the tourism concept. Add table for other concepts roots.. Add another table with the statistics of this extraction. Add a cool example table or figure?}

\begin{table}[h]
\centering
\begin{tabular}{l}
\hline
\textbf{Algorithm 1:} Cultural Artifact Extraction from Wikidata \\ \hline
\begin{tabular}[t]{l}
\textbf{Input:} Set of root nodes $R$, Maximum hops $H$ \\
\textbf{Output:} Set of cultural artifacts $A$ \\
\end{tabular} \\ \hline
\begin{tabular}[t]{l}
1:  $A \gets \emptyset$, $h \gets 0$ \\
2:  \textbf{while} $h < H$ \textbf{do} \\
3:  \hspace{0.5cm} $R_{new} \gets \emptyset$  \\
4:  \hspace{0.5cm} \textbf{for} $r \in R$ \textbf{do} \\
5:  \hspace{1cm}    $C \gets$ Children of $r$ along nodes (P31) or (P279) \\
6:  \hspace{1cm}    \textbf{for} $c \in C$ \textbf{do} \\
7:  \hspace{1.5cm}       \textbf{if} $c$ has property (P495) or (P17) \textbf{then} \\
8:  \hspace{2cm}          $A \gets A \cup \{c\}$ \\
9:  \hspace{1.5cm}       \textbf{else} \\
10: \hspace{2cm}          $R_{new} \gets R_{new} \cup \{c\}$ \\
11: \hspace{1.5cm}    \textbf{end for} \\
12: \hspace{1cm} \textbf{end for} \\
13: \hspace{0.5cm} $R \gets R_{new}$  \\
14: \hspace{0.5cm} $h \gets h + 1$ \\
15: \textbf{end while} \\
16: \textbf{return} $A$\\
\end{tabular} \\
\hline
\end{tabular}
\end{table}

\subsection{Cleaning of CUBE Collection}

In this step, we clean, extend and organize the noisy KB nodes using an LLM in loop. We also use Search results to help organize the collection better

\subsubsection{Filtering and expanding artifacts using LLM. }

\nithish{We do this step to make Cube Collection exhaustive}

While extraction from knowledge base is exhaustive, it is expected to have missing information or inconsistencies \citep{kannen-etal-2023-best}. In order to compliment the KB extraction, we use GPT4-Turbo to first filter out cultural artifacts that may not necessarily belong to a concept space. \textcolor{red}{Add examples}. We take inspiration from existing self refinement \citep{madaan2023selfrefine} and self critiquing \citep{lahoti-etal-2023-improving} techniques to perform this step. Once we filter out the erroneous artifacts, we prompt GPT-4 to fill out popular missing artifacts from the cultural concept, similar to the diversity expansion application in \citep{lahoti-etal-2023-improving}. This way we use an LLM to 1) correct and filter out incorrect extractions and 2) to fill out possible gaps in cultural artifacts extracted from the WikiData knowledge base.

\subsubsection{Sampling artifacts for Cube-1K}

\nithish{We do this step to pick artifacts for Cube-1K}

We further collect the Google Search results for each artifact that would servse as a noisy estimator of the popularity of the artifact. We change the Geo-location of Search API to get local results. This step not only improves the organization of cultural artifacts for picking cultural awareness prompts (Cube-1K). Using the search results, we sample artifacts based on the popularity search scores as prevalancs.

\subsection{Task Details}

In order to evaluate the cultural competence of models, we are interested in 2 tasks that are listed below:

% \begin{table*}[t]
% \centering
% \caption{Example Task Prompts}
% \label{tab:prompts}
% \begin{tabular}{p{0.45\textwidth} p{0.45\textwidth}}
% \toprule
% \textbf{Cultural Awareness Prompts} & \textbf{Cultural Diversity Prompts} \\
% \midrule
% A high resolution image of \textcolor{blue}{Sukiyaki} from Japanese cuisine & Image of Indian cuisine \\
% \multirow{2}{*}{A panoramic view of \textcolor{red}{Christ the Redeemer} in Brazil} & Image of Nigerian clothing \\
%  & Image of tourist landmarks in Italy \\
% An image of \textcolor{brown}{Bhangra} performance from India &  \\
% \bottomrule
% \end{tabular}
% \end{table*}

\begin{table*}[t]
\centering
\caption{Example task prompts to probe the model for Cultural Awareness and Cultural Diversity}
\label{tab:prompt_examples}
\begin{tabular}{p{0.57\textwidth} p{0.39\textwidth}}
\toprule
\textbf{Cultural Awareness} & \textbf{Cultural Diversity } \\
\midrule
\textbf{Cuisine} & \textbf{Cuisine} \\
- A high resolution image of \textcolor{blue}{Sushi} from Japanese cuisine & - Image of Nigerian clothing \\
- A high resolution image of \textcolor{blue}{Eba} from Nigerian cuisine & - Image of a dish from Japanese cuisine \\
\midrule
\textbf{Landmarks} & \textbf{Landmarks} \\
- A panoramic view of \textcolor{red}{Christ the Redeemer} in Brazil & - Image of tourist landmarks in Italy \\
- A panoramic view of \textcolor{red}{Qutub Minar} in India & - Image of a tourist spot in France\\
\midrule
\textbf{Art} & \textbf{Art} \\
- An image of \textcolor{teal}{Bhangra} performance from India & - Image of traditional Japanese clothing \\
- An image of \textcolor{teal}{Kanzashi} from Japanese clothing & - Image of a costume from Turkey \\
\bottomrule
\end{tabular}
\end{table*}

\fi

\subsection{Human Annotation}
% \subsection{Cultural Awareness Metrics} %%%%%%%%%%%%%%% Adji: I have added my edits/comments to this section %%%%%%%%%%%%%%%

% 
%In this section, we detail the metric used to evaluate the different aspects in the benchmark. We evaluate the \textbf{Cultural Awareness} and \textbf{Cultural Diversity} of the models given the corresponding prompts from the benchmark.

In order to evaluate cultural awareness of the T2I models, we asked human annotators questions that are analogous to standard metrics used in T2I evaluation: a) \textit{faithfulness} and b) \textit{realism} also called \emph{fidelity}. Each annotator was presented with the AI-generated image for an artifact and the corresponding description along with the country association, 
% and was asked three questions as described in Table~\ref{tab:annotation_guidelines}.
and was asked the following questions:
% 
% \begin{comment}
\begin{enumerate}[nosep,leftmargin=*]
%%%%% Adji's comment: we should keep the same names "faithfulness", "Realism" as above or replace those phrases above with "Image-Description faithfulness" and "Realism". 
    \item \textbf{Cultural Relevance:} Based solely on the image, does the item depicted belong to the annotator's country? (Yes/No/Maybe)
    \item \textbf{Faithfulness:} If the image is from the annotator's country, how well does it match the item in the text description? (1-5 Likert scale)
    \item \textbf{Realism:} How realistic does the image look, regardless of faithfulness? (1-5 Likert scale, with optional comment for scores $\leq$ 3)
\end{enumerate}
% \end{comment}
% 
% Annotators were instructed to consider only the visual aspects of the image when assessing cultural relevance. For the faithfulness, both the image and the text were taken into account by the annotators. Realism was evaluated independently of cultural relevance or the textual description. The detailed guidelines is presented in Appendix A.
% 
% Each one of the 8 locations considered had a specific pool of rater. The rater in each location were of diverse gender and are familiar with the specific culture of the location.
% Before collecting the annotation a pilot task was created to train the raters in each pool. The pilot tasks consisted of about 50 prompts and the corresponding generated images. The prompts were from multiple concepts in order to expose the raters to all of them. After compliting the pilot task the initial annotation and any additional feedback about the instruction task was collected and examined. We refined the instructions according to the the feedback to remove possible ambiguity. We also consider systemtic errors by raters and created a golden set for the raters to study to complete their training. Once this final stage of training was completed we proceeded with the annotation for the 1K prompts across the 3 concept and 8 location presented in Table~\ref{tab:prompt_availability_rev}.
% 
% To assess the cultural relevance, faithfulness, and realism of generated images, we established a rigorous annotation process.
% 
We recruited diverse groups of raters from each of the countries we consider. Each rater pool underwent comprehensive training and was also given a "golden set" of examples, as reference.
%, starting with a pilot task involving approximately 50 prompts and generated images across various cultural concepts.  Feedback from this pilot task was used to refine the annotation instructions and address any ambiguities.To further enhance rater accuracy, we identified and corrected systematic annotation errors by the taggers and developed a "golden set" of examples for they could use as reference.
%
Once training was complete, the raters proceeded to annotate the 1K prompts spanning the three concepts and the eight countries outlined in Table 3.
Raters were instructed to focus on both the image and text when evaluating cultural relevance, and solely the image for realism. Detailed guidelines for each criterion (Appendix \ref{human_annot_details}), the inter-annotator agreement (Appendix \ref{irr_section}),  and the interface used for human annotation (Figure \ref{fig:human_rater_ui}) can be found in Appendix.

% We report the a a majority vote as consensus for our analysis, but will retain the individual annotations in the benchmark (following \cite{prabhakaran2021releasing})  for further analysis on subjectivity.

\begin{comment}

\begin{table}[t]
\centering
\small
% \caption{Annotation Guidelines Summary}

\begin{tabular}{p{2.9cm}|p{7.1cm}|p{2.6cm}}
\toprule
Measure & Question (Scale/Response) & Additional Notes \\
\midrule
Cultural Relevance & Assess if the image depicts an item found in the annotator's country (Yes/No/Maybe) & Based solely on the image. \\
\midrule
Image-Text faithfulness & If culturally relevant, rate how closely the image matches the item described in the text. (1: Not at all - 5: Exactly) & Considers both image and text. \\
\midrule
Realism & Rate the perceived realism of the image, independent of cultural relevance or faithfulness with the description. (1: Not at all - 5: Extremely) & Optional explanation for unrealistic aspects for scores $\leq$ 3. \\
\bottomrule
\end{tabular}
\caption{Summary of annotation guidelines to assess cultural awareness of T2I models through cultural relevance, faithfulness to input text description, and realism.\label{tab:annotation_guidelines}}
\end{table}

\end{comment}

\begin{figure*}[h]
\centering
\includegraphics[width=0.9\textwidth]{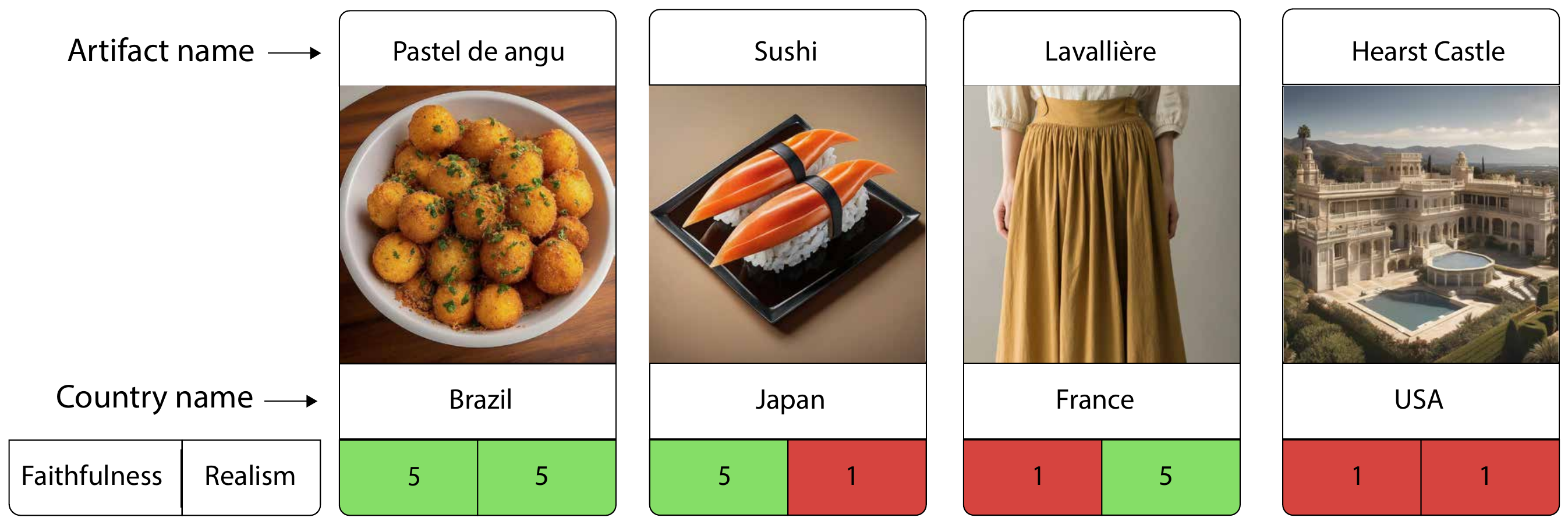}
\caption{
Examples of human evaluation results on cultural awareness for T2I models with high and low scores on faithfulness and realism. More qualitative examples are in Figures \ref{fig:qualitative_examples-1} and \ref{fig:qualitative_examples-2}.}

\label{fig:imagegrid}
\end{figure*}

\subsection{Results}

% 
% We generated images for CUBE-1K sampled prompts for 3 concepts (cuisine, landmarks, and art) and 8 locations using two T2I systems: Imagen 2 and Stable diffusion XL (SDXL). Three human raters are selected from the rater pool of a given location and  are request to answer the 3 questions in Table~\ref{tab:annotation_guidelines}.
%
Figure~\ref{fig:imagegrid} presents examples of faithfulness and realism scores for images that were deemed culturally relevant. In ~\ref{fig:imagegrid}(a), the model was prompted to generate \textit{Pastel de angu} from Brazilian cuisine and raters gave perfect score for both faithfulness and realism. In contrast, raters gave the lowest score of {1} for both aspects in ~\ref{fig:imagegrid}(d), clearly identifying that it is neither faithful nor realistic. Similarly, the image of \textit{Sushi} (\ref{fig:imagegrid}(b)) from Japanese cuisine got an faithfulness score of {5}, but realism score of {1} with an observation: "\textit{The fish looks hard and made of glossy plastic.}". Whereas, the image of \textit{lavallière} from France is realistic but not faithful, according to the raters.

\begin{table}[h]
\centering
\resizebox{\textwidth}{!}{
\begin{tabular}{llcccccccc}
\toprule
\textbf{Concept} & \textbf{Model} & \textbf{India} & \textbf{Japan} & \textbf{Italy} & \textbf{USA} & \textbf{Brazil} & \textbf{France} & \textbf{Turkey} & \textbf{Nigeria} \\
\midrule
\multicolumn{10}{c}{\textbf{Faithfulness}} \\
\midrule
\multirow{2}{*}{Cuisine} 
 & Imagen & \cellcolor[HTML]{EFEFEF}2.8 $\pm$ 1.9 & \cellcolor[HTML]{EFEFEF}2.4 $\pm$ 1.3 & \cellcolor[HTML]{EFEFEF}2.6 $\pm$ 1.5 & 3.4 $\pm$ 1.4 & \cellcolor[HTML]{D9D9D9}1.9 $\pm$ 1.5 & 3.1 $\pm$ 1.5 & \cellcolor[HTML]{EFEFEF}2.2 $\pm$ 1.4 & \cellcolor[HTML]{EFEFEF}2.7 $\pm$ 1.5 \\
 & SDXL & \cellcolor[HTML]{EFEFEF}2.1 $\pm$ 1.7 & \cellcolor[HTML]{D9D9D9}1.8 $\pm$ 0.6 & \cellcolor[HTML]{EFEFEF}2.2 $\pm$ 1.1 & 3.7 $\pm$ 1.3 & \cellcolor[HTML]{D9D9D9}1.5 $\pm$ 1.0 & \cellcolor[HTML]{EFEFEF}2.8 $\pm$ 1.4 & \cellcolor[HTML]{D9D9D9}1.8 $\pm$ 1.1 & \cellcolor[HTML]{EFEFEF}2.1 $\pm$ 1.3 \\
\cmidrule(lr){2-10}
\multirow{2}{*}{Landmarks} 
 & Imagen & 3.6 $\pm$ 1.8 & \cellcolor[HTML]{EFEFEF}2.2 $\pm$ 0.9 & \cellcolor[HTML]{EFEFEF}2.6 $\pm$ 1.2 & 3.8 $\pm$ 0.6 & \cellcolor[HTML]{EFEFEF}2.5 $\pm$ 1.7 & 4.0 $\pm$ 0.9 & 3.6 $\pm$ 0.9 & \cellcolor[HTML]{EFEFEF}2.4 $\pm$ 0.8 \\
 & SDXL & \cellcolor[HTML]{EFEFEF}2.7 $\pm$ 1.7 & \cellcolor[HTML]{EFEFEF}2.0 $\pm$ 0.7 & \cellcolor[HTML]{EFEFEF}2.2 $\pm$ 0.9 & 3.3 $\pm$ 1.3 & \cellcolor[HTML]{EFEFEF}2.5 $\pm$ 1.6 & 4.0 $\pm$ 0.6 & 3.0 $\pm$ 0.8 & \cellcolor[HTML]{D9D9D9}1.9 $\pm$ 0.7 \\
\cmidrule(lr){2-10}
\multirow{2}{*}{Art} 
 & Imagen & 3.5 $\pm$ 1.8 & \cellcolor[HTML]{EFEFEF}2.8 $\pm$ 0.9 & 4.2 $\pm$ 1.2 & 3.3 $\pm$ 1.2 & \cellcolor[HTML]{EFEFEF}2.9 $\pm$ 1.8 & 3.7 $\pm$ 1.0 & \cellcolor[HTML]{EFEFEF}2.5 $\pm$ 1.3 & \cellcolor[HTML]{EFEFEF}2.1 $\pm$ 1.4 \\
 & SDXL & 3.2 $\pm$ 1.8 & \cellcolor[HTML]{EFEFEF}2.0 $\pm$ 0.8 & 3.0 $\pm$ 1.2 & 3.9 $\pm$ 1.7 & \cellcolor[HTML]{EFEFEF}2.2 $\pm$ 1.6 & 3.2 $\pm$ 1.0 & \cellcolor[HTML]{EFEFEF}2.1 $\pm$ 1.4 & \cellcolor[HTML]{EFEFEF}2.0 $\pm$ 1.2 \\
\midrule
\multicolumn{10}{c}{\textbf{Realism}} \\
\midrule
\multirow{2}{*}{Cuisine} 
 & Imagen & 4.2 $\pm$ 0.5 & \cellcolor[HTML]{EFEFEF}2.0 $\pm$ 0.8 & 3.2 $\pm$ 0.9 & \cellcolor[HTML]{EFEFEF}2.2 $\pm$ 0.9 & 4.4 $\pm$ 0.7 & 3.4 $\pm$ 0.9 & \cellcolor[HTML]{EFEFEF}2.4 $\pm$ 0.8 & 3.3 $\pm$ 0.9 \\
 & SDXL & 3.6 $\pm$ 1.1 & \cellcolor[HTML]{D9D9D9}1.4 $\pm$ 0.6 & \cellcolor[HTML]{EFEFEF}2.2 $\pm$ 1.3 & \cellcolor[HTML]{D9D9D9}1.9 $\pm$ 0.9 & \cellcolor[HTML]{EFEFEF}2.1 $\pm$ 1.4 & \cellcolor[HTML]{EFEFEF}2.8 $\pm$ 1.4 & \cellcolor[HTML]{EFEFEF}2.2 $\pm$ 0.8 & 3.3 $\pm$ 0.9 \\
\cmidrule(lr){2-10}
\multirow{2}{*}{Landmarks} 
 & Imagen & 3.8 $\pm$ 1.0 & \cellcolor[HTML]{D9D9D9}1.7 $\pm$ 0.6 & \cellcolor[HTML]{EFEFEF}2.1 $\pm$ 1.4 & \cellcolor[HTML]{EFEFEF}2.4 $\pm$ 1.2 & \cellcolor[HTML]{EFEFEF}2.5 $\pm$ 1.6 & 3.2 $\pm$ 1.2 & \cellcolor[HTML]{EFEFEF}2.7 $\pm$ 1.0 & \cellcolor[HTML]{EFEFEF}2.9 $\pm$ 0.9 \\
 & SDXL & 3.7 $\pm$ 1.0 & \cellcolor[HTML]{D9D9D9}1.5 $\pm$ 0.6 & \cellcolor[HTML]{EFEFEF}2.7 $\pm$ 1.3 & \cellcolor[HTML]{EFEFEF}2.1 $\pm$ 0.8 & 3.5 $\pm$ 0.9 & 3.9 $\pm$ 0.6 & \cellcolor[HTML]{EFEFEF}2.5 $\pm$ 0.8 & 3.6 $\pm$ 0.7 \\
\cmidrule(lr){2-10}
\multirow{2}{*}{Art} 
 & Imagen & 3.4 $\pm$ 1.4 & \cellcolor[HTML]{EFEFEF}2.3 $\pm$ 0.8 & \cellcolor[HTML]{EFEFEF}2.6 $\pm$ 1.4 & \cellcolor[HTML]{D9D9D9}1.3 $\pm$ 0.6 & \cellcolor[HTML]{EFEFEF}2.4 $\pm$ 1.5 & \cellcolor[HTML]{D9D9D9}1.9 $\pm$ 1.3 & \cellcolor[HTML]{D9D9D9}1.6 $\pm$ 0.7 & \cellcolor[HTML]{EFEFEF}2.2 $\pm$ 1.2 \\
 & SDXL & \cellcolor[HTML]{EFEFEF}2.8 $\pm$ 1.4 & \cellcolor[HTML]{D9D9D9}1.4 $\pm$ 0.9 & \cellcolor[HTML]{D9D9D9}1.6 $\pm$ 1.1 & \cellcolor[HTML]{D9D9D9}1.4 $\pm$ 0.7 & \cellcolor[HTML]{D9D9D9}1.3 $\pm$ 0.6 & \cellcolor[HTML]{EFEFEF}2.1 $\pm$ 1.4 & \cellcolor[HTML]{D9D9D9}1.6 $\pm$ 0.8 & 3.0 $\pm$ 1.0 \\
\bottomrule
\end{tabular}
}
\caption{Comparison between Imagen 2 and Stable Diffusion XL (SDXL) for Faithfulness and Realism. The reported score is the average consensus score on the 1 to 5 scale and the standard deviation among 3 annotators for each country. Cells are highlighted to indicate scores below 3 (light gray) and below 2 (dark gray).}
\label{tab:model_comparison_restructured_final}
\end{table}
Table~\ref{tab:model_comparison_restructured_final} presents the average consensus scores (and standard deviations) for both faithfulness and realism, as rated for each model across regions and concepts. Both Imagen 2 and SDXL exhibit substantial room for improvement in both faithfulness and realism. Both models achieve relatively lower scores for countries regarded as the Global South (such as Brazil, Turkey, and Nigeria), with this disparity particularly pronounced for faithfulness. On average, in comparison to faithfulness, realism scores are lower across geo-cultures. While Imagen generally outperforms SDXL, exceptions exist, such as art faithfulness in the USA where SDXL scores higher. Table~\ref{tab:yes_no_q1} (in Appendix) shows the percentage of times our raters from each region deemed the images generated by each model to be culturally relevant (i.e., a yes answer to the first question in Annotation guidelines~\ref{human_annot_details}) showing non-uniform disparities across models and cultures. This suggests that the cultures marginalized by any particular model may depend on factors such as training data, reiterating the need for such cross-cultural benchmarks.

\section{ Evaluating Cultural Diversity}
\label{diversity_5}
\vspace{-2mm}
We seek to assess the cultural diversity of T2I outputs across different seeds as a way to measure the model's intrinsic latent space cultural diversity \citep{xu2024good}. For instance, a model capable of generating a diverse array of cultural artifacts across a range of seeds demonstrates the cultural richness of its learned representations. A more detailed note on our motivation for seed variation is outlined in Appendix \ref{seed_story}. For this, we focus on \textit {under-specified} prompts \citep{hutchinson2022underspecification} - prompts that elicit the generation of diverse cultural artifacts (e.g. "Image of tourist landmarks") rather than specific objects (e.g. "Image of Eiffel Tower"). 
We then seek to answer: \emph{What is the geo-cultural diversity of the generated cultural artifacts for prompts that mention just a concept?}. We further study the within-culture diversity in Appendix (\ref{within-culture}) and perform a correlation analysis of cultural diversity with existing metrics (in Appendix\ref{correlation_analysis}).
% can serve as a grounding to effectively assess the cultural diversity exhibited by the T2I models. 
% We start by describing the metric followed by the experimental setup and results. 
% Motivated by the importance of T2I models capturing the richness of global cultures, we propose cultural diversity as a new evaluation component in T2I models. We firstly discuss the metric, which is an adaptation of quality-weighted Vendi score, and it's suitability to measure cultural diversity. Next, we describe the experimental setup and research questions we wish to answer.

\vspace{-2mm}
\subsection{Cultural Diversity (CD)}
\label{metric_theory}

Existing works that focus on visual diversity use LPIPS~\citep{zameshina2023diverse} and Coverage~\citep{hall2024dig}, based on image embeddings. However, these metrics are not directly applicable in our case, as the similarity here may be attributed to color, texture, spatial orientation, and other visual aspects of the images. Measuring the cultural diversity of text-to-image (T2I) models requires a approach that accounts for both the variety of generated cultural artifacts and the quality of images generated from text prompts. To address this, we introduce \textbf{Cultural Diversity (CD)}, a new T2I evaluation component leveraging the quality-weighted Vendi Score~\citep{nguyen2024quality}. 
\vspace{-2mm}
\subsubsection{Foundation: Vendi Scores}
Vendi scores are a family of interpretable diversity metrics that satisfy the axioms of ecological diversity \citep{dan2023vendi, pasarkar2023cousins}. Vendi score  captures the "effective number" of distinct items within a collection, considering both richness (number of unique elements) and evenness (distribution of those elements), and is defined as follows

% To measure diversity, we start with Vendi scores, a family of interpretable diversity metrics that satisfy the axioms of ecological diversity~\citep{friedman2023vendi, pasarkar2023cousins} and are defined as follows:
% , defined as follows:
% We seek a metric for cultural diversity that can also account for the quality of the generated images. Recently, \citep{nguyen2024qualityweighted} have proposed \emph{quality-weighted Vendi scores} (qVS), to extend the Vendi scores to account for quality of the items in a given collection. The Vendi scores are a family of interpretable diversity metrics that satisfy the axioms of ecological diversity~\citep{friedman2023vendi, pasarkar2023cousins}. They are defined as follows:
\begin{definition}[Vendi Scores]\label{def:vendi-scores}
     Let $X = (x_1, \dots, x_N)$ be a collection of $N$ items. Let $k: \mathcal{X} \times \mathcal{X} \rightarrow \mathbb{R}$ be a positive semi-definite similarity function, such that $k(x, x) = 1$ for all $x \in \mathcal{X}$. Denote by $K \in \mathbb{R}^{N \times N}$ the kernel matrix whose $i,j$ entry $K_{i, j} = k(x_i, x_j)$. Further denote by $\lambda_1, \lambda_2, \ldots, \lambda_N$ the eigenvalues of $K$ and their normalized counterparts by $\overline{\lambda}_1, \dots, \overline{\lambda}_N$ where $\overline{\lambda}_i = \lambda_i / \sum_{i = 1}^N \lambda_i$. The Vendi score of order $q \geq 0$ is defined as the exponential of the Renyi entropy of the normalized eigenvalues of $K$,
     
    %  $\mathrm{VS}q (X; k) = \exp \left( \frac{1}{1 - q} , \log \biggl( \sum{i = 1}^N (\overline{\lambda}_i)^q \biggr) \right)$ where  $0 *\, \log 0 = 0$.

     \begin{equation}\label{eq:vs_q}
        \mathrm{VS}_q (X; k) = \exp \left( \frac{1}{1 - q} \, \log \biggl( \sum_{i = 1}^N (\overline{\lambda}_i)^q \biggr) \right),
    \end{equation}

    where we use the convention $0 *\, \log 0 = 0$.
\end{definition}

 The order $q$ determines the sensitivity allocated to feature prevalence, with values of $q < 1$ being more sensitive to rarer features and $q > 1$ putting more emphasis on more common features. When $q = 1$, we recover the original Vendi score~\citep{dan2023vendi}, the exponential of the Shannon entropy of the normalized eigenvalues of $K$.

\setlength{\abovedisplayskip}{3pt}
\setlength{\belowdisplayskip}{3pt}

\subsubsection{Incorporating Quality: Quality-Weighted Vendi Scores}

While Vendi scores measure diversity, they treat all items equally without considering individual quality. In the context of T2I, however, it is crucial to account for the quality of the generated images conditional on text prompts. 
% However, the original version of Vendi score (VS) \citep{friedman2023vendi} treat items in the collection the same regardless of their quality. To measure cultural diversity for T2I models, it is crucial to account for the quality of the generated images conditional on text prompts.
We therefore rely on \emph{quality-weighted Vendi scores} (qVS) \citep{nguyen2024quality} that extends VS to account for the quality of items in a given collection.
% to measure cultural diversity in a T2I setting. 
qVS is defined as the product of the average quality of the items in the collection and their diversity,
\begin{equation}\label{eq:qvs}
    \mathrm{qVS}_q(X; k, s) = \left( \frac{1}{N}\sum_{i = 1}^N s(x_i) \right) \, \mathrm{VS}_q(X; k),
\end{equation}
where $s(\cdot)$ is a function that scores the quality of the items. 
% In our setting, the quality of generated images is a crucial component along with diversity of cultural artifacts. 

In order to be able to compare different collections of images with different sizes, we normalize qVS by the size of the collection to measure cultural diversity:

\begin{equation}\label{eq:CD}
     q\overline{\mathrm{VS}}_q(X; k, s) = \left( \frac{1}{N}\sum_{i = 1}^N s(x_i) \right) \, \left(\frac{\mathrm{VS}_q(X; k)}{N}\right).
\end{equation}

We employ the HPS-v2 metric \citep{wu2023human} as the  $s(\cdot)$ function to score the quality of T2I outputs. HPS-v2, trained on 790k human preferences, provides a quality score $s \in [0, 1]$ for an image conditioned on text prompt, making it suitable proxy to measure image quality in our case. We leave exploration of other quality measures of salience of generated artifacts, for future work.

$q\overline{\mathrm{VS}}$ is minimized to 0 when every element has a quality score of 0, and is maximized to 1, when all elements have a perfect quality (s = 1) and are all distinct from each other ($\overline{\mathrm{VS}}$ = 1) 

% To quantify the quality of T2I outputs, we employ the HPS-v2 metric \citep{wu2023human} that provides a quality score $s \in [0, 1]$ for of an image conditioned on text prompt. HPS-v2 is trained on 790k human preferences across key aspects such as faithfulness, aesthetics, and realism of generated images, making it suitable proxy to measure image quality in our case. We leave exploration of other quality metrics that can also accomodate saliency of generates artifacts for future works
% 

% In order to be able to compare different collections of images with different sizes, we normalize qVS by the size of the collection, that leads us to the Cultural Diversity (CD) evaluation component for T2I models:

% \begin{equation}\label{eq:qvs}
% \mathrm{CD}_q(X; k, s) = q\overline{\mathrm{VS}}q(X; k, s) = \left( \frac{1}{N}\sum{i = 1}^N s(x_i) \right) , \left(\frac{\mathrm{VS}_q(X; k)}{N}\right),
% \end{equation}

\subsubsection{Desirable Properties of $q\overline{\mathrm{VS}}$}

$q\overline{\mathrm{VS}}$ has many desiderata in the context of T2I models:  it accounts for similarity and inherits several desirable features of the Vendi scores such as sensitivity to richness and evenness. It also exhibits quality-awareness, duplication scaling and offers flexibility to define kernels that capture different facets of geo-cultural diversity.
\paragraph{Properties.}
    Consider the same setup as in Definition \ref{def:vendi-scores}.
    \begin{itemize}[label=$\star$]
        \item \textbf{Quality-awareness.} Denote by $\mathcal{C}_1 = (x_1, \dots, x_M)$ and $\mathcal{C}_2 = (y_1, \dots, y_L)$ two collections such that $\mathrm{VS}_q(\mathcal{C}_1; k) = \mathrm{VS}_q(\mathcal{C}_2; k)$. Denote by $s(\cdot)$ a function that scores the quality of an item such that $\frac{1}{M}\sum_{i = 1}^{M} s(x_i) \geq \frac{1}{L}\sum_{j = 1}^{L} s(y_j)$. Then
        \begin{itemize}
      	\item[] \centering 
      	    $q\overline{\mathrm{VS}}_q(\mathcal{C}_1; k, s) \geq q\overline{\mathrm{VS}}_q(\mathcal{C}_2; k, s)$ .
      	\end{itemize}
        \item \textbf{Duplication scaling.} Denote by $\mathcal{C} = (x_1, \dots, x_N)$ a collection of $N$ items. Define $\mathcal{C}'$ as the collection containing all elements of $\mathcal{C}$ each duplicated M times. Then
        \begin{itemize}
      	\item[] \centering 
      	  $q\overline{\mathrm{VS}}_q(\mathcal{C}; k, s) = M\cdot q\overline{\mathrm{VS}}_q(\mathcal{C}'; k, s)$ .
      	\end{itemize}
      	\item \textbf{Kernel generalizability.}  Let $k_1(\cdot,\cdot)$ and $k_2(\cdot,\cdot)$ represent two different positive semi-definite similarity functions. Then, given a collection $\mathcal{C}=(x_1,\dots,x_N)$, the quantities $q\overline{\mathrm{VS}}_q(\mathcal{C};k_1,s)$ and $q\overline{\mathrm{VS}}_q(\mathcal{C};k_2,s)$ may capture different aspects of diversity based on the properties of $k_1$ and $k_2$.
      	
    \end{itemize}
%\end{theorem}
% The proof of the quality-awareness property is immediate following the definition of the normalized qVS. See Appendix \ref{proof_theorem} for a proof of the duplication scaling property. While the kernel generazalizable property is intuitive, we demonstrate its usefulness in measuring different aspects of cultural diversity in Section \ref{global_diversity}.  
We state above 3 core properties of $q\overline{\mathrm{VS}}$ that makes it suitable for measuring cultural diversity in T2I models:  1) prioritizes collections with higher-quality items when other factors are equal, 2) penalizes the duplication of elements, and 3) exhibits flexibility in capturing various aspects of diversity through the selection of an appropriate similarity kernel. The proof of the quality-awareness property is immediate following the definition of $q\overline{\mathrm{VS}}$. See Appendix \ref{proof_theorem} for a proof of duplication scaling.

\subsection{Experimental Setup}

We discuss the experimental pipeline: 1) \textbf{Prompting and Seeding:} We calculate $q\overline{\mathrm{VS}}$ for 8 images per prompt, matching the typical number of output images of image-generation APIs. To account for variances in prompt wording as well as seed selections, we report scores averaged over 50 repetitions.  2) \textbf{Mapping Generated Images to cultural artifacts:} We map each image to to its most closely resembling artifact from the concept space of the domain.\footnote{Not all text-to-image generated images perfectly represent real-world cultural entities.} Note that since the prompts focus on global concepts, we obtain the {continent, country, and artifact name} annotation for each generated image. 3) \textbf{Computing Vendi Scores:} With each generated image linked to its closest artifact, we compute the cultural diversity of the generated outputs using the metric defined in Section \ref{metric_theory}. We expand on the details of each of these steps in Appendix \ref{cd_pipeline}. Different kernels can capture different aspects of geo-cultural diversity.

\paragraph{Kernel definition.} With each generated image linked to its closest cultural artifact, we now compute the \textit{cultural diversity} (CD) of the model's output using the definition in Section \ref{metric_theory}. We define a general similarity kernel that allows us to analyze different aspects of geo-cultural diversity:

\begin{equation}
k(x_i, x_j) = w_1 \cdot k_1(x_i, x_j) + w_2 \cdot k_2(x_i, x_j) + w_3 \cdot k_3(x_i, x_j)
\end{equation}

where  $k_1(\cdot, \cdot)$, $k_2(\cdot, \cdot)$, and $k_3(\cdot, \cdot)$ are three distinct kernels measuring different aspects of similarity, and $w_1$, $w_2$, $w_3$ assign weights to each. We define $k_1(x_i, x_j) = 1$ if $x_i$ and $x_j$ have the same continent, and 0 otherwise. Similarly,  $k_2(x_i, x_j) = 1$ if the two items share the same country, and 0 if not. Lastly, $k_3(x_i, x_j) = 1$ if the two items represent the same artifact, regardless of geographical origin, and 0 otherwise.
To illustrate this flexibility, we present results under different kernel configurations:

\begin{itemize}
    \item \textbf{\colorbox{ContinentColor}{Continent-level diversity}}: $w_1 = 1$, $w_2 = 0$, $w_3 = 0$. Considers continent-level similarity.
    \item \textbf{\colorbox{CountryColor}{Country-level diversity}}: $w_1 = 0$, $w_2 = 1$, $w_3 = 0$. Considers country-level similarity.
    \item \textbf{\colorbox{ArtifactColor}{Artifact-level diversity}}: $w_1 = 0$, $w_2 = 0$, $w_3 = 1$. Only considers distinct artifacts.
    \item \textbf{\colorbox{HierarchicalColor}{Hierarchical geographical diversity}}: $w_1 = 1/2$, $w_2 = 1/2$, $w_3 = 0$. This captures a hierarchical notion of diversity where both continent and country similarities are penalized equally, without explicitly considering individual artifacts.
    \item \textbf{\colorbox{UniformColor}{Uniformly weighted diversity}}: $w_1 = 1/3$, $w_2 = 1/3$, $w_3 = 1/3$. 
\end{itemize}

\textbf{Models.} We evaluate 4 models across closed-source and open-source model types: 1) Imagen 2, 2) Stable-Diffusion-XL, 3) Playground, and 4) Realistic Vision. More details about the model usage and hyperparameters are provided in Appendix \ref{models_eval}. 
%We further analyze within cultural diversity for Imagen in Appendix \ref{fig:result_within_imagen}.
 %Further details regarding the computation of cultural diversity scores as well as justification for the usage of HPS-v2 as the quality score can be found in Appendix \ref{cult_div_scores_appendix}.

\subsection{Results}
\label{global_diversity}

Results in Figure \ref{fig:wold_map_models_combined} reveals that when prompted with under-specified prompts mentioning general concepts (Fig \ref{tab:task_prompts}), current T2I models tend to generate artifacts that lack comprehensive geographical representation. This finding aligns with previous observations \citep{basu2023inspecting}, suggesting a bias towards well-represented and popular countries.

\begin{table}[t]
\centering
\captionsetup{font=small}
\caption{Breakdown of the mean quality component (q) and mean diversity component ($q\overline{\mathrm{VS}}$) averaged over 50 repetitions. While all models show relatively low quality scores (as per HPS-v2), Playground (PG) has best quality for \textit{cuisine} and \textit{art} concepts and Imagen-2 (IM) for \textit{landmarks}. Different kernels $(w_1, w_2, w_3)$ capture different aspects of diversity.}
\renewcommand{\arraystretch}{1.2} % Increase row spacing
\scalebox{0.83}{
\begin{tabular}{l cccc|cccc|cccc}
\toprule
& \multicolumn{4}{c|}{\cellcolor{gray!20}\textbf{Cuisine}} & \multicolumn{4}{c|}{\cellcolor{gray!20}\textbf{Landmarks}} & \multicolumn{4}{c}{\cellcolor{gray!20}\textbf{Art}} \\
\cmidrule(lr){2-5} \cmidrule(lr){6-9} \cmidrule(lr){10-13}
& \textbf{IM} & \textbf{SDXL} & \textbf{PG} & \textbf{RV} & \textbf{IM} & \textbf{SDXL} & \textbf{PG} & \textbf{RV} & \textbf{IM} & \textbf{SDXL} & \textbf{PG} & \textbf{RV} \\
\midrule
\cellcolor{gray!30}q ($\rightarrow$) & {0.27} & 0.21 & \textbf{0.29} & 0.27 & \textbf{0.25} & 0.22 & 0.21 & 0.23 & {0.31} & 0.30 & \textbf{0.34} & 0.33 \\
\midrule
\cellcolor{gray!30}$\overline{\mathrm{VS}}(w_1, w_2, w_3)$ & \multicolumn{12}{c}{} \\
\cmidrule(lr){2-13}
\cellcolor{ContinentColor}$\overline{\mathrm{VS}}$ (1, 0, 0) & \textbf{0.32} & 0.23 & 0.24 & \underline{0.27} & 0.17 & \textbf{0.27} & 0.23 & \underline{0.25} & \textbf{0.23} & 0.14 & \underline{0.18} & 0.16 \\
\cellcolor{CountryColor}$\overline{\mathrm{VS}}$ (0, 1, 0) & \textbf{0.59} & \underline{0.53} & 0.51 & 0.51 & 0.50 & \textbf{0.65} & 0.34 & \underline{0.52} & \textbf{0.42} & 0.29 & \underline{0.37} & 0.23 \\
\cellcolor{ArtifactColor}$\overline{\mathrm{VS}}$ (0, 0, 1) & \textbf{0.91} & 0.71 & \underline{0.82} & 0.74 & 0.73 & \textbf{0.84} & 0.58 & \underline{0.81} & \textbf{0.72} & \underline{0.60} & 0.51 & 0.44 \\
\cellcolor{HierarchicalColor}$\overline{\mathrm{VS}}$ ($\frac{1}{2}$, $\frac{1}{2}$, 0) & \textbf{0.51} & \underline{0.44} & 0.41 & 0.38 & 0.42 & \textbf{0.53} & 0.31 & \underline{0.45} & \textbf{0.36} & 0.24 & \underline{0.31} & 0.22 \\
\cellcolor{UniformColor}$\overline{\mathrm{VS}}$ ($\frac{1}{3}$, $\frac{1}{3}$, $\frac{1}{3}$) & \textbf{0.72} & 0.58 & \underline{0.66} & 0.59 & \underline{0.55} & \textbf{0.66} & 0.45 & 0.52 & \textbf{0.52} & 0.39 & \underline{0.41} & 0.30 \\
\bottomrule
\end{tabular}
}
\label{fig:breakdown_div}
\end{table}

Table \ref{fig:breakdown_div} presents the results for both the average quality score (q) and the diversity component ($q\overline{\mathrm{VS}}$) across different kernels. Playground and Imagen generally achieve the highest quality scores based on the HPS-v2 metric. As anticipated, models exhibit the lowest diversity for $w_1 = 1$, $w_2 = 0$, $w_3 = 0$, which considers only continent-level similarity, due to the limited number of continents. Conversely, $w_1 = 1$, $w_2 = 0$, $w_3 = 0$, focusing solely on artifact diversity, yields the highest scores, reflecting the wide array of potential cultural artifacts. In terms of overall performance, Imagen 2 consistently demonstrates the best $q\overline{\mathrm{VS}}$ scores across different kernels for the \textit{Cuisine} and \textit{Art} concepts, whereas SDXL obtains the highest scores for \textit{Landmarks} concept. Table \ref{tab:CD_results} shows the cultural diversity ($q\overline{\mathrm{VS}}$).  Note that the scores across the board are still low, remaining far from the maximum score of 1. Current T2I models fall short of representing the true breadth and richness of global cultural diversity.

\begin{table}[t]
\centering
\captionsetup{font=small}
\caption{CD scores across models and concepts using various similarity kernels averaged over 50 repetitions. Imagen 2 (IM) performs best for \textit{Cuisine} and \textit{Art}, while SDXL performs best for \textit{Landmarks}. Importantly, even the best scores are low, indicating significant room for improvement in the cultural diversity of T2I outputs.}
\renewcommand{\arraystretch}{1.2} % Increase row spacing
\scalebox{0.83}{
\begin{tabular}{lcccc|cccc|cccc}
\toprule
& \multicolumn{4}{c|}{\cellcolor{gray!20}\textbf{Cuisine}} & \multicolumn{4}{c|}{\cellcolor{gray!20}\textbf{Landmarks}} & \multicolumn{4}{c}{\cellcolor{gray!20}\textbf{Art}} \\
\cmidrule(lr){2-5} \cmidrule(lr){6-9} \cmidrule(lr){10-13}
\textbf{CD} & \textbf{IM} & \textbf{SDXL} & \textbf{PG} & \textbf{RV} & \textbf{IM} & \textbf{SDXL} & \textbf{PG} & \textbf{RV} & \textbf{IM} & \textbf{SDXL} & \textbf{PG} & \textbf{RV} \\
\midrule
\cellcolor{ContinentColor}$q\overline{\mathrm{VS}}$(1, 0, 0) & \textbf{0.08} & 0.04 & 0.07 & \underline{0.07} & 0.04 & \textbf{0.06} & 0.04 & \underline{0.05} & \textbf{0.07} & 0.042 & \underline{0.06} & 0.05 \\
\cellcolor{CountryColor}$q\overline{\mathrm{VS}}$(0, 1, 0) & \textbf{0.15} & 0.11 & \underline{0.14} & 0.13 & \underline{0.12} & \textbf{0.14} & 0.07 & \underline{0.12} & \textbf{0.13} & 0.08 & \underline{0.12} & 0.07 \\
\cellcolor{ArtifactColor}$q\overline{\mathrm{VS}}$(0, 0, 1) & \textbf{0.24} & 0.14 & 0.23 & 0.20 & \textbf{0.18} & \textbf{0.18} & \underline{0.12} & \textbf{0.18} & \textbf{0.22} & \underline{0.18} & 0.17 & 0.14 \\
\cellcolor{HierarchicalColor}$q\overline{\mathrm{VS}}$($\frac{1}{2}$, $\frac{1}{2}$, 0) & \textbf{0.13} & 0.09 & \underline{0.12} & 0.10 & \underline{0.10} & \textbf{0.11} & 0.06 & \underline{0.10} & \textbf{0.11} & 0.07 & \underline{0.10} & 0.072 \\
\cellcolor{UniformColor}$q\overline{\mathrm{VS}}$($\frac{1}{3}$, $\frac{1}{3}$, $\frac{1}{3}$) & \textbf{0.19} & 0.12 & \textbf{0.19} & \underline{0.16} & \underline{0.14} & \textbf{0.15} & 0.09 & 0.12 & \textbf{0.16} & 0.12 & \underline{0.14} & 0.10 \\
\bottomrule
\end{tabular}
}
\label{tab:CD_results}
\end{table}

\vspace{-2mm}
\section{Discussion and Limitations} \label{sec:discussion}
\vspace{-2mm}

To the best of our knowledge, CUBE is the first large-scale cultural competence benchmark for text-to-image models. From our investigations so far, one clear finding stands out: there is yet significant headroom for improvement of global cultural competence in the current generation of text-to-image models --- both in terms of awareness and diversity. This seems especially true for the Global South, highlighting the urgency of the need for comprehensive and informative cultural competence testing frameworks. Towards that goal, we have made CUBE dataset and code public\footnote{https://github.com/google-deepmind/cube}, and encourage its adoption and expansion by the multimodal generative AI community.

Design of this benchmark required many challenging decisions that needed careful thought. The increased coverage of our benchmark as a result of largely automated, scalable approaches - built on curated data sources and existing model capabilities still comes with its own limitations. For instance, existing structured knowledge bases such as WikiData are known to have inherent cultural biases reflecting disparities in global distribution of knowledge production \citep{callahan2011cultural,bjork2021new}. Therefore, it is important to note that our approach of expanding coverage using existing knowledge bases should be complemented with community based and participatory approaches for richer socio-cultural representation \citep{alonso-alemany-etal-2023-bias,dev2024building}. Notably, the World Wide Dishes effort \citep{magomere2024eatfeedingfoundationmodels} builds a dataset of images representing dishes from around the world through a community-led effort that complements our dataset that relies on existing knowledge bases.\footnote{www.worldwidedishes.com}

Furthermore, even with significant filtering and completion, we expect our data to be noisier than other methods. We have not yet explored the potential utility of our curation method, nor of the dataset itself, to empower other research on evaluating or improving cultural competence in generative models in general. Many aspects of the curation process for CUBE is automated, due to the large scale of the problem --- e.g., we rely on image similarity scores and LLM-based selection flows to ground generated images to a specific cultural artifact. This approach risks ingraining biases that may already exist within these tools into the benchmark construction itself. The process of mapping artifacts to country/continent may also introduce biases since the annotator VLM itself may not be aware of several cultural artifacts around the world. On the other hand, human annotation to measure faithfulness and realism is also quite challenging and subjective, as annotators may not be aware of the multitude of representations of their own culture. Even beyond cultural knowledge, different cultures may also have different standards for realism --- which could result in mis-calibrated results obtained from human annotations, making it hard to compare across different cultures (e.g., Table~\ref{fig:interesting_examples}). 

We acknowledge that our results are susceptible to such errors stemming from both the subjective nature of human annotations (for faithfulness/realism), as well as issues in VLM annotations (for diversity). Nevertheless, the evaluation methods and frameworks we introduce in this work hold significant relevance, as we move towards training multicultural models with globally diverse datasets \citep{pouget2024filter} in our path to equitable representation in generative AI.
Finally, we use a narrow definition of culture, defined in terms of geo-political boundaries such as countries and continents in our kernel definition for diversity. However, culture is a more complex concept --- countries are rarely monolithic in terms of cultures, and cultural scopes may often transcend geo-political boundaries. Future work could investigate applying our metric to other finer-grained definitions of cultural groups.

\vspace{-2mm}
\section{Conclusion}
\vspace{-2mm}

We introduced CUBE, a T2I benchmark to assess the cultural competence of T2I models along two crucial dimensions: cultural awareness and cultural diversity. We presented a scalable methodology with a potential to be scaled beyond eigth countries and three concepts considered in this work. Furthermore, we proposed a novel T2I evaluation component: cultural diversity (CD) and measured it using the quality-aware Vendi score. Our comprehensive human evaluation reveals substantial gaps in cultural awareness across cultures and concepts, as well as the gaps in the geo-cultural diversity of model generations. Our correlation analysis reveals a noteworthy trend: while faithfulness and realism exhibit a moderate positive correlation, suggesting they can be improved in tandem, cultural diversity remains weakly correlated to these metrics. By highlighting existing limitations in cultural competence of T2I models, we believe our work contributes to a critical dialogue surrounding the development of truly inclusive generative AI systems. 

\section{Ethical Considerations}
\vspace{-2mm}
We built a large repository of cultural artifacts with the intended use of evaluation of T2I models. Our approach to build this resource relies partly on automated tools, including LLMs, that have been shown to exhibit various societal biases. Hence, care must be taken in interpreting the results of evaluation using this benchmark. While the CUBE benchmark enables a broad-coverage and flexible evaluation of cultural competence in T2I models, their coverage is still limited by the underlying resources it is built on --- namely WikiData (the KB) and GPT-4 Turbo (the LLM). Future work should explore bridging the gaps in coverage through participatory efforts in partnership with communities of people within respective cultures. 
Furthermore, both CUBE-CSpace and CUBE-1K are intended to be used in evaluation pipelines, rather than training or mitigation efforts. 

\section{Acknowledgements}
\vspace{-2mm}
We thank Partha Talukdar, Kathy Meier-Hellstern, Caroline Pantofaru, Remi Denton, Susanna Ricco, David Madras, Nitish Gupta and Sunipa Dev for their feedback and advice; Lucas Beyer and Xiaohua Zhai for  insights and support on use of mSigLIP for auto-evals; Sagar Gubbi and Kartikeya Badola for helpful discussions on the human rating template; Shikhar Vashishth for discussions on the use of WikiData; Preetika Verma for assistance with the SLING framework %Simran Khanuja for providing crucial inputs on LAION dataset
and Dinesh Tewari and the annotation team for facilitating our human evaluation work. 

\newpage
\bibliographystyle{neurips_2024}
\bibliography{neurips_2024}

\newpage

\appendix

\vspace{-2mm}

\section{Contributions}
\vspace{-2mm}
This paper was the result of close collaboration and teamwork. Nithish worked on the ideation of the dataset extraction and metrics, and implemented the end-to-end pipelines for dataset extraction and evaluation. Arif was part of the explorations and contributed to data cleaning, image generation, and the quality evaluation pipeline under the guidance of Pushpak. Marco participated in the design discussions for datasets and metrics and owned the data analysis of human annotation results. Utsav and Vinod kept us honest on the cultural dimension of this work. Adji contributed to defining how to measure cultural diversity, including how to use qVS to measure it and how to design the similarity kernel. Shachi oversaw the entire project and provided guidance and mentorship to Nithish and Arif. Everyone contributed to paper writing.

\section{Correlation Analysis of CD}
\label{correlation_analysis}

In this section, we investigate the correlations between our three key metrics: faithfulness, realism (Table \ref{tab:model_comparison_restructured_final}), and diversity (Figure \ref{fig:result_within_imagen}) across different geo-cultures, focusing on the Imagen model. Our analysis reveals a positive correlation between faithfulness and realism ($\rho$ = 0.400), as shown in Table \ref{tab:correlation_only} for cultural prompts. This suggests that images judged as more faithful to cultural prompts tend to also be perceived as more realistic on average.

%  and relatively lower correlation between diversity-faithfulness (0.016) and diversity-realism (0.156) metrics. This shows that cultural diversity is not necessarily higher for geo-cultures that have higher faithfulness and realism of generated images.

\begin{table}[htbp]
\centering
\footnotesize
\begin{tabular}{lccc}
\toprule
\textbf{Concept} & \multicolumn{3}{c}{\textbf{Correlation ($\rho$)}} \\
\cmidrule(lr){2-4}
 & \textbf{Faithfulness-Realism} & \textbf{Faithfulness-Diversity} & \textbf{Realism-Diversity} \\
\midrule
Cuisine & 0.306 & -0.138 & 0.117 \\
Landmarks & 0.548 & 0.435 & 0.183 \\
Art & 0.347 & -0.248 & 0.167 \\
\midrule
\textbf{Mean} & \textbf{0.400} & \textbf{0.016} & \textbf{0.156} \\
\bottomrule
\end{tabular}
\caption{Pearson correlation coefficients ($\rho$) for different metric pairs for Imagen-2. We use the mean ratings (for faithfulness and realism) and within-culture diversity scores for each culture.}
\label{tab:correlation_only}
\end{table}

Conversely, we observe much weaker correlations between diversity and both faithfulness ($\rho$ = 0.016) and realism ($\rho$ = 0.156). This key finding suggests that higher faithfulness and realism in generations for certain cultures do not necessarily translate to higher diversity of the generated cultural artifacts.

Our findings resonate with a a recent work \citep{astolfi2024consistencydiversityrealismparetofrontsconditional} that discusses the faithfulness-diversity-realism Pareto fronts on a geodiverse dataset, where the prompts concern everyday real-world objects. We show that even in the cultural context, faithfulness and realism are improved concurrently, whereas there is little correlation between diversity and the other metrics.  This raises a critical question: \textit{does the current trajectory of text-to-image (T2I) model development, optimized for human preferences of aesthetics, faithfulness, and realism, suffices to improve the intrinsic cultural diversity of T2I outputs for under-specified prompts?} Our findings suggest a need to explicitly incorporate diversity as a core pillar in the multi-objective development of T2I models, as models become increasingly accessible to diverse cultures globally.

\section{Additional details of CUBE}

Below we provide details on the choice of countrues,  CUBE-1K dataset breakdown, WikiData root nodes and some technical details for CUBE construction.

\subsection{Justification for the choice of countries}

We selected eight countries from different geo-cultural regions across continents and the Global South-North divide: Brazil (LatAm), France (Europe), India (SEA), Italy (Europe), Japan (East Asia), Nigeria (SSA), Turkey (Middle East), and USA (North America). Our goal was to choose countries with the largest population in each of these regions, while also taking into account (a) their representation in training data (e.g. Nigeria is “low-resource” whereas USA is “high-resource”), and (b) availability of raters from that region through our vendor. We limited our study to 8 countries because of time and monetary constraints. While we acknowledge that this list of countries is necessarily incomplete, and may result in a biased global sampling, future iterations of this work could include a wider range of countries for a more comprehensive evaluation.

.

\subsection{CUBE-1K statistics}

\begin{table}[H]
\centering
\small
\begin{tabular}{lcccccccc}
\toprule
\textbf{} & \textbf{Brazil} & \textbf{India} & \textbf{Japan} & \textbf{Nigeria} & \textbf{Turkey} & \textbf{Italy} & \textbf{USA} & \textbf{France} \\
\midrule
Cuisine              & 58 & 73 & 62 & 61 & 63 & 77 & 56 & 67 \\
Landmarks   & 33 & 40 & 41 & 25 & 39 & 36 & 44 & 37 \\
Art                  & 27 & 27 & 26 & 22 & 26 & 22 & 22 & 21 \\
\midrule
\textbf{Total} & 118 & 140 & 129 & 108 & 128 & 135 & 122 & 125 \\
\bottomrule
\end{tabular}
\caption{Dataset Statistics of CUBE-1K used for evaluating Cultural Awareness \label{tab:CUBE-1K-stats}}
\end{table}

\subsection{Wikidata}

Table \ref{tab:wiki_nodes} shows the seed set of manually selected root nodes from WikiData, that each represent different each concepts, used to extract CUBE-CSpace.

\begin{table*}[t]
\centering
\caption{Wikidata IDs of the Root Nodes for Art, Cuisine and Tourism cultural artifacts}
\label{tab:wiki_nodes}
\begin{minipage}[t]{0.45\textwidth}  % Adjust widths as needed
\centering
\begin{tabular}{ll} 
\toprule
\rowcolor{blue!15}
\textbf{Wikidata ID} & \textbf{Art} \\
\midrule
Q11460      & Clothing \\
Q9053464    & Costume  \\
Q3172759    & Traditional costume \\
\midrule
Q17399019   & Style of Painting \\
\midrule
Q107357104  & Type of dance      \\
Q1153484   & Folk art \\
Q45971958  & Performing arts genre \\
\midrule
\rowcolor{green!15}
\textbf{Wikidata ID} & \textbf{Cuisine} \\
\midrule
Q746549     & Dish \\
Q2095       & Food \\
Q19861951   & Type of food or dish \\
\midrule % Separator line
\rowcolor{gray!15} 
\textbf{Wikidata ID} & \textbf{Landmarks} \\
\midrule
Q210272 & Cultural Heritage \\
Q41176 & Building \\
Q33506 & Museum \\
Q16560 & Palace \\
Q23413 & Castle \\
Q22698 & Park \\
Q1107656 & Garden \\
Q24398318 & Religious building \\
Q4989906 & Monument \\
Q2416723 & Theme park \\
% Q11635 & Theatre \\
% Q39614 & Cemetery \\
Q16999091 & Landmarks \\
Q1785071 & Fort \\

\bottomrule
\end{tabular}
\end{minipage}
\hfill % Add some space between tables
\begin{minipage}[t]{0.45\textwidth}  % Adjust width as needed
\centering
\begin{tabular}{ll}  
\toprule
\rowcolor{gray!15}
\textbf{Wikidata ID} & \textbf{Landmarks} \\
\midrule
Q9259 & World Heritage Site \\
Q3395377 & {Ancient monument} \\
Q109607 & Ruins \\
Q207694 & Art museum \\
Q7075 & Library \\
Q811979 & {Architectural  structure} \\
Q842858 & National museum \\
Q3152824 & {Cultural  institution} \\
Q1060829 & Concert hall \\
Q153562 & Opera house \\
Q1007870 & Art gallery \\
Q15243209 & {Historic district} \\
Q143912 & Triumphal arch \\
Q1329623 & Cultural center \\
Q28737012 & {Museum of culture} \\
Q622425 & Nightclub \\
Q11635 & Theatre \\
Q839954 & Archaeological site \\
Q39614 & Cemetery \\
Q12271 & Architecture \\
Q11303 & Skyscraper \\
Q12280 & Bridge \\
Q39715 & Lighthouse \\
Q483110 & Stadium \\
Q1200957 & {Tourist  destination} \\
Q167346 & Botanical garden \\
Q2281788 & Public aquarium \\ 
\bottomrule
\end{tabular}
\end{minipage}

\end{table*}

\subsection{Technical Details}
\label{tech_details}

We have provided additional technical details on CUBE.

\paragraph{KB Extraction.} We have reported the root nodes for each domain in \ref{tab:wiki_nodes} We iterated for a total of 4 hops beginning from these root nodes as the majority of the artifacts were found in the second and third hop for all the 3 domains we considered. The new artifacts extracted began to plateau after the 4th hop. 

\paragraph{Self-Refinement.} We divide the self refinement of the concept space into two steps: 1) Removing noise: An incorrect artifact that does not belong to either that country or cultural concept. We leverage LLMs for this filtering step by asking “Can you classify if the <item> belongs to <country> <concept>? Answer yes or no.”,  2) Adding missing artifacts: We leverage the self-critiquing technique introduced in \citep{lahoti-etal-2023-improving} by following “critique the response” and “address the critiques and rewrite” steps for each of the artifact lists.

\paragraph{Manual Filtering.} CUBE-1K is intended to serve as a high-quality curated prompt set to represent cultural artifacts selected for relevance and popularity. As noted earlier, we use the local Google Search results as a proxy for popularity within that local context. While this provides us with a broad set of artifacts in each cultural context, some of them may show inflated results because of the commonality of certain words in their names (e.g., “Puri” is the name of a famous temple in India and also a popular dish.). To mitigate this, we used a manual filtering process conducted by annotators from their respective cultures. This process removes any artifacts that may have been artificially boosted by inflated search results. The criteria for manual filtering include visual distinctiveness, alignment with the corresponding geo-cultural category, and the artifact's popularity within the culture.

\paragraph{Noise in Google Search.} Google Search can potentially inflate the search results due to the presence of common words in the names of cultural artifacts. Here are some examples:

\begin{table}[h]
\centering
\begin{tabular}{lllc}
\toprule
\textbf{Culture} & \textbf{Artifact} & \textbf{Search Results} & \textbf{Reason} \\ 
\midrule
Turkey          & Van Museum        & 35.5M            & Contains "Van" \\ 
Japan           & Japan Monkey Park & 75.9M            & Contains "monkey" \\ 
Nigeria         & Freedom Park      & 17.8M            & Contains "freedom" \\ 
\bottomrule
\end{tabular}
\caption{Examples of inflated search results due to the presence of common words in artifact names.}
\label{tab:search_inflation}
\end{table}

These artifacts are examples of noisy outcomes where the search results were inflated due to the presence of common words.

\paragraph{Annotation Details.} Our annotators were recruited based on the following criteria:
 \begin{itemize}
     \item Be fluent in English
     \item Be diverse in Gender
     \item Be from specific countries with familiarity of local culture
 \end{itemize}

% \begin{table}[H]
% \centering
% \footnotesize
% \resizebox{0.9\textwidth}{!}{
% \begin{tabular}{lcccccccc}
% \toprule
% \textbf{Model} & \textbf{India} & \textbf{Japan} & \textbf{Italy} & \textbf{USA} & \textbf{Brazil} & \textbf{France} & \textbf{Turkey} & \textbf{Nigeria} \\
% \midrule
% Imagen & 98.4 & 90.8 & 86.8 & 95.0 & 94.3 & 95.2 & 68.1 & 73.6 \\
% SDXL   & 95.2 & 84.4 & 82.2 & 99.2 & 52.8 & 96.0 & 66.4 & 86.1 \\
% \bottomrule
% \end{tabular}
% }
% \caption{Percentage of generated images by Imagen and SDXL for each cultural region that our raters from that region deemed to be culturally relevant to their region. All concepts are combined here; see Table \ref{tab:model_comparison_restructured} for the breakdown by different responses.
% \label{tab:yes_no_q1}}
% \end{table}

\begin{table}[h]
\centering

\resizebox{\textwidth}{!}{
\begin{tabular}{llcccccccc}
\toprule
\textbf{Model} & \textbf{Response} & \textbf{India} & \textbf{Japan} & \textbf{Italy} & \textbf{USA} & \textbf{Brazil} & \textbf{France} & \textbf{Turkey} & \textbf{Nigeria} \\
\midrule
\multirow{4}{*}{Imagen} 
 & Yes          & 98.4 & 90.8 & 86.8 & 95.0 & 94.3 & 95.2 & 68.1 & 73.6 \\
 & Maybe        &  1.6 &  5.5 & 10.9 &  4.1 &  5.7 &  2.4 & 28.3 & 14.2 \\
 & No           &  0.0 &  0.9 &  0.8 &  0.0 &  0.0 &  0.0 &  0.0 &  2.8 \\
 & No consensus &  0.0 &  2.8 &  1.6 &  0.8 &  0.0 &  2.4 &  3.5 &  9.4 \\
\cmidrule(lr){2-10}
\multirow{4}{*}{SDXL} 
 & Yes          & 95.2 & 84.4 & 82.2 & 99.2 & 52.8 & 96.0 & 66.4 & 86.1 \\
 & Maybe        &  4.0 &  7.3 & 14.7 &  0.8 & 45.3 &  3.2 & 31.0 & 13.9 \\
 & No           &  0.8 &  3.7 &  0.8 &  0.0 &  0.9 &  0.0 &  0.9 &  0.0 \\
 & No consensus &  0.0 &  4.6 &  2.3 &  0.0 &  0.9 &  0.8 &  1.8 &  0.0 \\
\bottomrule
\end{tabular}
}
\caption{Comparison between Imagen 2 and Stable Diffusion (SDXL) for Cultural relevance question when all concepts are combined.
}
\label{tab:model_comparison_restructured}
\end{table}

\section{Detailed Human Annotation Questions}
\label{human_annot_details}

\paragraph{Q1: Based on your country's culture, is this image something one might see in your country?}
\begin{itemize}
\item \textbf{Yes:} This image is definitely something someone in my country could come across. It aligns with what I know about our culture.
\item \textbf{Maybe:} This image looks somewhat familiar for someone from my country, but I'm not entirely sure.
\item \textbf{No:} This image does not look like it could be from my country at all. (Provide a mandatory justification.)
\end{itemize}
\textit{Note: Only consider the image for this question.}

\paragraph{If the answer to Q1 is "Yes":}

\paragraph{Q2: How well does the image match the item in the text description? (Rate from 1 to 5)}
\begin{itemize}
\item \textbf{1: Not at all:} The item in the image doesn't look anything like the item described in the text.
\item \textbf{2: A little:} The image has some resemblance to the item, but there are major differences.
\item \textbf{3: Somewhat:} The image is somewhat similar to the item, but there are noticeable differences.
\item \textbf{4: Mostly:} The image closely matches the item, but with some small differences.
\item \textbf{5: Exactly:} The image perfectly matches the description.
\end{itemize}
\textit{Note: Consider both the image and the textual description for this question.}

\paragraph{Q3: How realistic does the image look? (Rate from 1 to 5)}
\begin{itemize}
\item \textbf{1: Not at all:} The image looks completely artificial or fake.
\item \textbf{2: A little:} The image has some realistic elements, but overall looks unrealistic.
\item \textbf{3: Somewhat:} The image is somewhat realistic, but has noticeable flaws.
\item \textbf{4: Mostly:} The image is mostly realistic, but there are some small details that look artificial.
\item \textbf{5: Extremely:} The image looks extremely real, like a photograph.
\end{itemize}
\textit{Optional: If the rating is $\leq$ 3, please add a short comment explaining the unrealistic parts.}
\textit{Note: Only consider the image for this question.}

% \begin{table*}[h]
% \centering
% \caption{Example task prompts to probe the model for cultural awareness and cultural diversity}
% \label{tab:task_prompts}
% \begin{tabular}{p{0.57\textwidth} p{0.39\textwidth}}
% \toprule
% \textbf{Cultural Awareness} & \textbf{Cultural Diversity } \\
% \midrule
% \textbf{Cuisine} & \textbf{Cuisine} \\
% - A high resolution image of \textcolor{blue}{Sushi} from Japanese cuisine & - Produce a high quality image of a dish \\
% - A high resolution image of \textcolor{blue}{Eba} from Nigerian cuisine & - Photo of a food dish \\
% \midrule
% \textbf{Landmarks} & \textbf{Landmarks} \\
% - A panoramic view of \textcolor{red}{Christ the Redeemer} in Brazil & - High definition image of a landmark \\
% - A panoramic view of \textcolor{red}{Qutub Minar} in India & - A realistic image of a monument\\
% \midrule
% %\textbf{Clothing \& Art} & \textbf{Clothing \& Art} \\
% \textbf{Art} & \textbf{Art} \\
% - An image of \textcolor{teal}{Bhangra} performance from India & - Image of traditional  clothing \\
% - An image of \textcolor{teal}{Kanzashi} from Japanese clothing & - A realistic image of a ethnic costume \\
% \bottomrule
% \end{tabular}
% \end{table*}

\begin{table}[ht]
\centering

\begin{tabular}{lccc}
\toprule
         &  \textbf{Cultural}  &  & \\
\textbf{Location} &  \textbf{Relevance} & \textbf{Faithfulness} & \textbf{Realism} \\
         &  \textbf{(majority agreement)} & \textbf{(Krippendorff's $\alpha$)} & \textbf{(Krippendorff's $\alpha$)} \\
\midrule
India   & 100\% & 0.58 & 0.29 \\
\midrule
Japan   & 96\% & 0.31 & 0.21 \\
\midrule
Italy   & 98\% & 0.16 & 0.21 \\
\midrule
USA & 99\% & 0.42 & 0.43 \\
\midrule
Brazil  & 99\% & 0.30 & 0.29 \\
\midrule
France  & 98\% & 0.09 & 0.08 \\
\midrule
Turkey  & 97\% & 0.21 & 0.08 \\
\midrule
Nigeria & 95\% & 0.21 & 0.12 \\
\bottomrule
\end{tabular}
\caption{Inter rater reliability for the 3 annotation tasks described in Section \ref{human_annot_details} and for all the rater pools across the different locations of our study. For the Cultural Relevance question, we report the observed majority agreement. For both the Faithfulness and the Realism questions, we report the Krippendorff's $\alpha$. }
\label{tab:irr_analysis}
\end{table}

\begin{table}[H]
\centering
\footnotesize
\resizebox{0.9\textwidth}{!}{
\begin{tabular}{lcccccccc}
\toprule
\textbf{Model} & \textbf{India} & \textbf{Japan} & \textbf{Italy} & \textbf{USA} & \textbf{Brazil} & \textbf{France} & \textbf{Turkey} & \textbf{Nigeria} \\
\midrule
Imagen & 98.4 & 90.8 & 86.8 & 95.0 & 94.3 & 95.2 & 68.1 & 73.6 \\
SDXL   & 95.2 & 84.4 & 82.2 & 99.2 & 52.8 & 96.0 & 66.4 & 86.1 \\
\bottomrule
\end{tabular}
}
\caption{Percentage of generated images by Imagen and SDXL, for each cultural region, that raters from that region deemed culturally relevant. All concepts are combined here; see Table \ref{tab:model_comparison_restructured} for the breakdown by different responses.
\label{tab:yes_no_q1}}
\end{table}

\begin{figure*}[h]  
    \centering
    \includegraphics[width=0.97\textwidth]{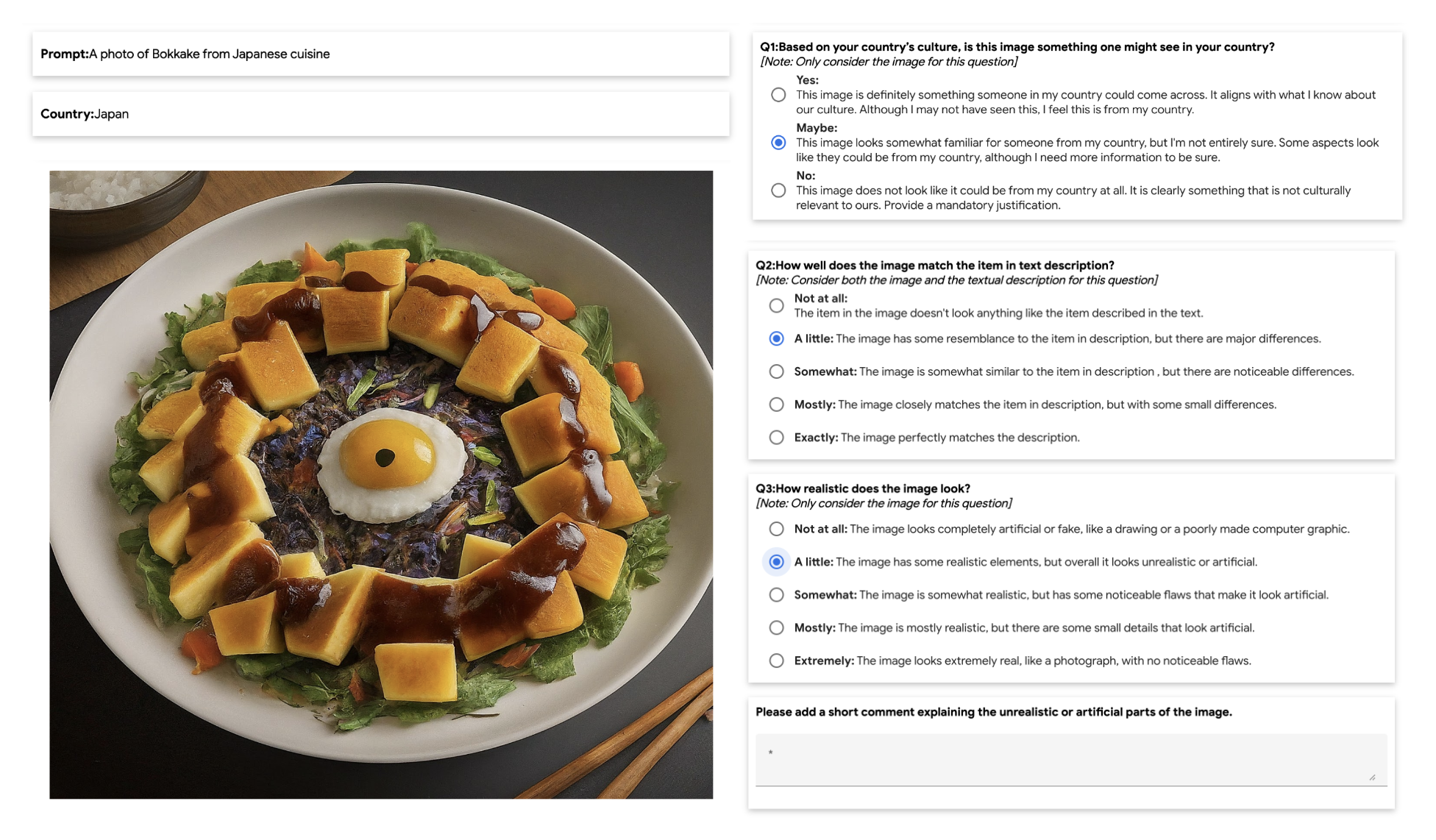}  % Set width to textwidth
    \caption{\textbf{Human annotation interface}. Each question was annotated by 3 raters. The first question tested cultural relevance and the second and third question were only shown if the raters agreed the images had relevance to their cultures (yes/maybe). An additional text box was provided for raters to comment on unrealistic elements in the image.}
    \label{fig:human_rater_ui}
\end{figure*}

\section{Inter-Annotator Agreement}
\label{irr_section}
We obtain high inter-rater agreement for the question on cultural relevance across all countries (all above 95\%; see Table~\ref{tab:irr_analysis}), suggesting that the question of whether an image is relevant to a particular culture is a relatively objective task. 
However, the question of faithfulness and realism yielded moderate to low agreement (especially for France and Turkey) among annotators (measured as Krippendorff's $\alpha$, which is better suited for Likert scale ordinal values) in line with the relatively more complex and subjective nature of the task (see Table \ref{fig:interesting_examples} for examples of edge cases). 

\section{On Realism}

Our focus on realism in our evaluation stems from the fact that generative language models are being deployed in products that increasingly shape the discovery of socio-cultural knowledge such as search, online education, and travel planning. In such contexts, cultural awareness is especially important, and realism of generated images is a crucial aspect in this regard. We agree that there may be usage contexts of the T2I models where realism of generated images may not be relevant — for instance, in creative contexts where people use these models to generate photo-realistic images which may be non-realistic in practice (e.g., “photo of Taj Mahal in a desert”). Such generations are not inherently bad, but in contexts where cultural awareness is relevant, our methodology enables the study of cultural awareness of any given model.

\section{Background on T2I Evaluation}

\paragraph{T2I Evaluation Metrics: }Early T2I evaluation approaches such as Inception Score \citep{salimans2016improved} and Frechet Inception Distance \citep{heusel2018gans} focused on the similarity of generated images to real ones, also called the realism. While there is active research on improving the realism metrics (e.g., \citep{jayasumana2024rethinking}), more recent work also assess faithfulness, through
embedding-based metrics such as CLIPScore \citep{hessel2022clipscore} and ALIGNScore \cite{zha2023alignscore}, VQA-based metrics such as TIFA \citep{hu2023tifa}, DSG \citep{cho2024davidsonian}, and VQAScore \citep{lin2024evaluating}, captioning-based metrics like LLMScore \citep{lu2023llmscore} and VIEScore \citep{ku2023viescore}, or approaches like VPEval \citep{cho2023visual} and ViperGPT \citep{suris2023vipergpt} that use visual programming.
% embedding-based metrics such as CLIPScore \citep{hessel2022clipscore} and ALIGNScore \cite{zha2023alignscore} that examine the semantic faithfulness between input text and output images. In contrast, VQA-based metrics such as TIFA \citep{hu2023tifa}, DSG \citep{cho2024davidsonian}, and VQAScore \citep{lin2024evaluating} leverage visual question answering frameworks to assess  faithfulness. Finally, captioning-based metrics like LLMScore \citep{lu2023llmscore} and VIEScore \citep{ku2023viescore} focus on how well captions generated from generated images correspond to the input text. Recent approaches like VPEval \citep{cho2023visual} and ViperGPT \citep{suris2023vipergpt} use visual programming for evaluating complex prompts that require reasoning. 
Other metrics such as ImageReward \citep{xu2023imagereward}, PickScore \citep{kirstain2023pickapic}, and HPSv2 \citep{wu2023human} fine-tune vision-language models on human ratings to better align with human preferences.
% 
% Evals of Social Nature
% While most of the research on T2I evaluation focuses on technical capabilities discussed above, t
There have also been some recent work on social aspects such as bias and fairness reflected in T2I models \citep{feng2022towards,naik2023social,zhang2023iti}, as well as several bias mitigation strategies \citep{wan2024survey}. 
Notably, there is work demonstrating biases around geo-cultural differences in model performance; e.g., ViSAGe \citep{jha2024visage} presents a global-scale analysis of stereotypes using a structured repository of stereotypes.
While these efforts demonstrate the importance of geo-cultural considerations in model evaluations, they are focused social stereotypes which is only one of the ways in which cultural differences show up in model predictions.

% Benchmarks
\paragraph{T2I Benchmarks: }There have also been efforts to build comprehensive evaluation benchmarks aimed at tracking the progress of model capabilities over time, focusing on tasks such as
realism, text faithfulness, and compositional abilities. These benchmarks, such as DrawBench \citep{saharia2022photorealistic}, CC500 \citep{feng2023trainingfree},
T2I-CompBench \citep{huang2023t2icompbench}, 
TIFA v1.0 \citep{hu2023tifa}, 
DSG-1k \citep{cho2024davidsonian}, 
GenEval \citep{ghosh2023geneval}, and 
GenAIBench \citep{lin2024evaluating} employ diverse prompts and metrics to assess factors such as image-text coherence, perceptual quality, attribute binding, faithfulness, semantic competence, and compositionality, to list a few. While more recent work such as HEIM \citep{lee2023holistic} do include more socially situated aspects such as toxicity, bias, and aesthetics, they do not probe for the cultural awareness of T2I models.
Table \ref{tab:overview} contrasts existing T2I evaluation benchmarks with ours, which we believe is a timely contribution to track and foster culturally inclusive T2I technology. 

\section{Background on seed variation in T2I models }
\label{seed_story}

Text-to-image models are predominately latent diffusion models \citep{rombach2022highresolution} that generate images conditioned on text prompts. The stochastic nature of the Gaussian noise in forward diffusion and the reparameterization step in reverse diffusion, influenced by random seeds \citep{xu2024good}, allows these models to produce different images for the same text prompt \citep{samuel2023generating, poyuan2023synthetic} - by simply varying the seeds. While there have been studies exploring the effect of seeds on neural network architectures \citep{picard2023torchmanualseed3407}, there has been little exploration on the impact of seeds in the diffusion process. A recent work studies the influence of seeds on interpretable visual dimensions such as style and quality of images \citep{xu2024good}. However, the diversity of concepts produced for different seeds is largely under-explored. 

\section{Cultural Diversity Pipeline}
\label{cd_pipeline}

% \begin{table*}[b]
% \centering
% \footnotesize % Reduce font size
% \caption{Example task prompts to probe the model for cultural awareness and cultural diversity}
% \label{tab:task_prompts}
% \begin{tabular}{p{0.57\textwidth} p{0.39\textwidth}}
% \toprule
% \textbf{Cultural Awareness} & \textbf{Cultural Diversity } \\
% \midrule
% \textbf{Cuisine} & \textbf{Cuisine} \\
% - A high resolution image of \textcolor{blue}{Sushi} from Japanese cuisine & - Produce a high quality image of a dish \\
% - A high resolution image of \textcolor{blue}{Eba} from Nigerian cuisine & - Photo of a food dish \\
% \midrule
% \textbf{Landmarks} & \textbf{Landmarks} \\
% - A panoramic view of \textcolor{red}{Christ the Redeemer} in Brazil & - High definition image of a landmark \\
% - A panoramic view of \textcolor{red}{Qutub Minar} in India & - A realistic image of a monument\\
% \midrule
% %\textbf{Clothing \& Art} & \textbf{Clothing \& Art} \\
% \textbf{Art} & \textbf{Art} \\
% - An image of \textcolor{teal}{Bhangra} performance from India & - Image of traditional  clothing \\
% - An image of \textcolor{teal}{Kanzashi} from Japanese clothing & - A realistic image of a ethnic costume \\
% \bottomrule
% \end{tabular}
% \end{table*}

We seek to analyze the global geo-cultural diversity of generated artifacts for under-specified prompts \citep{hutchinson2022underspecification} that simply mention the concept, such as "Image of traditional clothing.". These prompts serve as great test-beds to analyse model's intrinsic cultural diversity.  We further analyze within-culture diversity (\textit{What is the diversity of cultural artifacts produced by model for different cultures?}) for Imagen 2 in Section \ref{within-culture}. 

\subsection{Prompting and Seeding}
We employ a straightforward prompting strategy tailored to our two research questions on geo-cultural diversity. Figure \ref{tab:task_prompts} (bottom half) shows some example cultural diversity prompts. For evaluating \textit{global geo-cultural diversity}, prompts consist only of the target concept (e.g., “Image of monuments”), enabling measurement of both the model’s global cultural inclination and the geo-cultural diversity in its generated artifacts. To assess \textit{within-culture diversity}, prompts specify both the concept and the culture (e.g., “Image of a Nigerian dish”), allowing for analysis of the model’s cultural richness within specific cultural contexts.

To account for prompt wording and seed dependency, we use five distinct prompt templates and generate a batch of eight images per template with consecutive seed values, beginning with seed 0. This batch size aligns with typical outputs from image-generation APIs, which usually produce four or eight images per prompt. This process is repeated across ten seed batches per prompt, yielding a total of 80 unique seed values (0 to 79) for each prompt. The diversity metric, capable of processing larger batches, is computed over the 400 images generated per prompt (5 prompts x 10 seed batches x 8 images/batch). The resulting diversity scores represent the mean over 50 repetitions to ensure robustness. Figure \ref{fig:result_within_imagen} illustrates the sensitivity of diversity to prompt variation.

\subsection{Computing Cultural Diversity} 
\label{vendi_compute}

In this section we describe the steps involved in computing the cultural diversity (CD) of T2I outputs for under-specified prompts. It first details the approach to mapping each generated image to a cultural artifact from CUBE-CSpace, followed by the Vendi score kernel definition to measure different aspects of geo-cultural diversity.

\subsubsection{Mapping generated images to cultural artifacts}
\label{mapping_to_artifacts}
To compute the cultural diversity of the generated images, we first map each generated image to its closest resembling artifact within CUBE-CSpace. This mapping is essential to anchor our analysis in real-world cultural artifacts, even though we acknowledge that not all text-to-image generated images may perfectly represent actual cultural entities\footnote{Generated images may deviate from real-world cultural representations.}. Given that our prompts are designed to focus on broad global concepts, this approach allows us to associate each generated image \( I \) with its corresponding continent \( c \), country \( r \), and artifact name \( a \), such that \( I \in \{c, r, a\} \).

For the mapping process, we employ an automated method that combines GPT-4-Turbo for verification with mSigLIP \citep{zhai2023sigmoid}-based retrieval techniques. In cases where generated images might contain multiple artifacts, we use negative prompting to encourage the depiction of a single, dominant artifact. GPT-4-Turbo is then used to validate that each image contains a clearly identifiable primary artifact. For \textit{global geo-cultural diversity}, where prompts describe global concepts, we use GPT-4-Turbo to confirm that the generated image aligns with the target concept. Following this, GPT-4-Turbo identifies the country most closely associated with the artifact, focusing on the country of prevalence or association rather than the origin. In the \textit{within-culture diversity} case, where prompts specify both concept and culture, GPT-4-Turbo verifies that the generated image aligns with the specified culture in the prompt. This multi-step verification leverages GPT-4-Turbo’s proficiency in recognizing cultural concepts \citep{cao2024exploring}. For both \textit{global} and \textit{within-culture diversity} analyses, we retrieve the top five most similar images from a reference set of cultural artifact images associated with the identified country. This retrieval process leverages the mSigLIP S400m model \citep{zhai2023sigmoid} for image-image similarity, which has been trained on a comprehensive global image dataset and is thus well-suited for this type of cultural analysis. The reference set comprises images sourced from Google Images\footnote{\url{https://developers.google.com/custom-search/v1/overview}} for artifacts within the prompt’s concept space. We recognize that Google Images, though extensive, may not represent the full range of global cultural artifacts. Finally, GPT-4-Turbo classifies each generated image by comparing it to the five retrieved reference artifacts, refining our results by reducing reliance on purely similarity-based retrieval. The entire mapping process is further validated by human reviewers on a small subset of images across diverse cultures, yielding an approximate accuracy of \( \sim 70\% \). Exploration of improved mapping strategies, potentially using multicultural visual language models (VLMs) or diverse annotator models, is left as an avenue for future work.

\subsubsection{Kernel Definition} 
\label{mapping_to_artifacts}
With each generated image mapped to its closest cultural artifact, we compute the \textit{cultural diversity} (CD) of the model's output using the definition from Section \ref{metric_theory}. We define a general similarity kernel to analyze various aspects of geo-cultural diversity:

\begin{equation}
k(x_i, x_j) = w_1 \cdot k_1(x_i, x_j) + w_2 \cdot k_2(x_i, x_j) + w_3 \cdot k_3(x_i, x_j)
\end{equation}

where \( k_1(\cdot, \cdot) \), \( k_2(\cdot, \cdot) \), and \( k_3(\cdot, \cdot) \) are three distinct kernels representing different aspects of similarity, and \( w_1 \), \( w_2 \), \( w_3 \) assign weights to each. Specifically, \( k_1(x_i, x_j) = 1 \) if \( x_i \) and \( x_j \) share the same continent, and 0 otherwise. Similarly, \( k_2(x_i, x_j) = 1 \) if the items share the same country, and 0 otherwise. Lastly, \( k_3(x_i, x_j) = 1 \) if the items represent the same artifact, irrespective of geographical origin, and 0 otherwise.

To demonstrate the flexibility of this kernel, we analyze cultural diversity using the following configurations:

\begin{itemize}
    \item \textbf{\colorbox{ContinentColor}{Continent-level diversity}}: \( w_1 = 1 \), \( w_2 = 0 \), \( w_3 = 0 \). This configuration considers only continent-level similarity.
    \item \textbf{\colorbox{CountryColor}{Country-level diversity}}: \( w_1 = 0 \), \( w_2 = 1 \), \( w_3 = 0 \). This focuses solely on country-level similarity.
    \item \textbf{\colorbox{ArtifactColor}{Artifact-level diversity}}: \( w_1 = 0 \), \( w_2 = 0 \), \( w_3 = 1 \). This disregards geographical associations and measures diversity based solely on distinct artifacts.
    \item \textbf{\colorbox{HierarchicalColor}{Hierarchical geographical diversity}}: \( w_1 = \frac{1}{2} \), \( w_2 = \frac{1}{2} \), \( w_3 = 0 \). This captures a hierarchical notion of diversity, balancing both continent and country similarities equally without accounting for individual artifacts.
    \item \textbf{\colorbox{UniformColor}{Uniformly weighted diversity}}: \( w_1 = \frac{1}{3} \), \( w_2 = \frac{1}{3} \), \( w_3 = \frac{1}{3} \). This provides equal weight to all three forms of similarity.
\end{itemize}

\subsubsection{HPS-v2 to measure quality}
\label{hpsv2}
To quantify the cultural diversity in a set of generated images, we employ the normalized qVS metric described in Section \ref{metric_theory}. This metric combines both the diversity of represented cultural artifacts and the quality of generated images. For the latter, we leverage the HPS-v2 metric \citep{wu2023human}, a state-of-the-art metric for evaluating text-to-image generation based on human preferences. HPS-v2 captures key aspects of image quality and faithfulness, effectively reflecting both the accuracy and aesthetic appeal of generated images.  While HPS-v2’s training data may not fully encompass the long-tail cultural artifacts considered in this work, it remains the most comprehensive and robust metric available for assessing human preferences in image generation, having been trained on a dataset of 790,000 human preference ratings. In the absence of datasets and evaluation models specifically designed for cultural contexts, we adopt HPS-v2 as a proxy for overall generation quality. For the diversity component of the metric, we apply the different kernel functions defined above to capture distinct aspects of geo-cultural diversity. We provide additional details on the computation and application of these kernels in Section \ref{vendi_compute}.

\subsubsection{Models Evaluated}
\label{models_eval}
We consider 4 models across closed-source and open-source model types. For closed-source, we evaluate  Imagen 2 via the Vertex AI \footnote{https://cloud.google.com/vertex-ai/generative-ai/docs/image/generate-images} and for open-sourced models we evaluate 1) Stable-Diffusion-XL-base-1.0, which is the most downloaded model on Huggingface, 2) Playground - highest rated open model on \texttt{T2I arena}\footnote{https://artificialanalysis.ai/text-to-image/arena} and 3) Realistic Vision - highest rated model on \texttt{imgsys.org}\footnote{https://imgsys.org/rankings}. The open models are downloaded from Hugingface \citep{wolf2020huggingfaces}. We use the default recommended hyperparameter settings for generation with each model.

\section{Within-Culture  Artifact Diversity }
\label{within-culture}

\begin{figure}[h]
    \centering
    \includegraphics[width=0.85\textwidth]{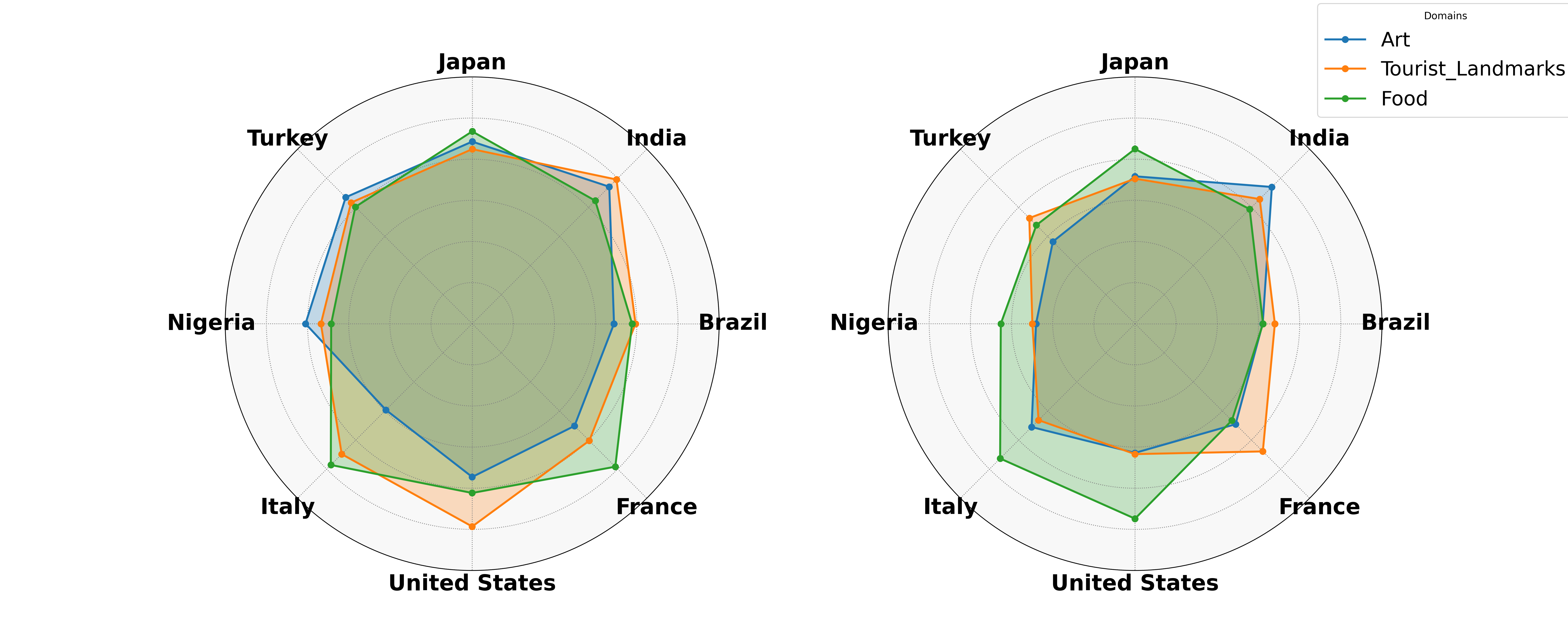}
    \begin{subfigure}{0.5\textwidth}
        \centering
        \caption*{(a) Imagen}
    \end{subfigure}%
    \begin{subfigure}{0.5\textwidth}
        \centering
        \caption*{(b) SDXL}
    \end{subfigure}
    
    \vspace{-0.5\baselineskip}
    \caption{Using within culture prompts, the above plot shows HPSv2 scores across all the three concepts to show quality of images produced for each geo-culture. Each subfigure compares the HPSv2 score for the models: (a) Imagen, and (b) SDXL}
    \label{fig:spider_hps}
\end{figure}

\begin{figure*}[h]  
    \centering
    \includegraphics[width=0.75\textwidth]{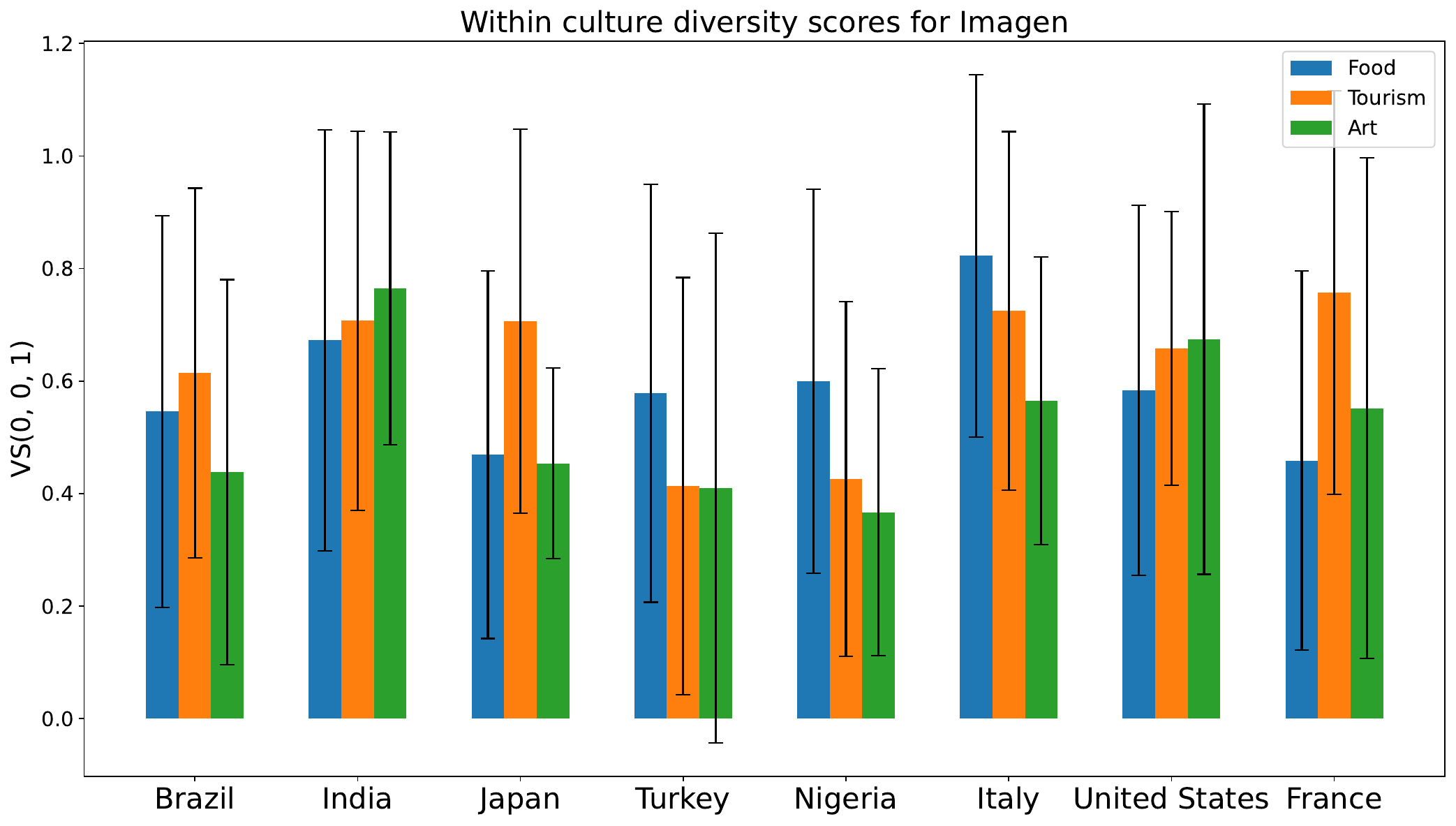}  % Set width to textwidth
    \caption{Within culture $\overline{\mathrm{VS}}$(0, 0, 1) scores for Imagen}
    \label{fig:result_within_imagen}
\end{figure*}

For evaluating within-culture diversity, prompts specify both the concept and the culture (e.g., "Image of a Nigerian dish"), enabling us to assess the richness of representations within a specific cultural context.
We analyze within-concept cultural diversity for country-specific, under-specified prompts, such as ``Image of a dish from Brazilian cuisine.'' Note that we explicitly mention both the concept and the geo-culture for which diversity is to be computed. We conduct this analysis for the aforementioned set of 8 countries and report the $\overline{\mathrm{VS}}$(0, 0, 1) for Imagen across countries in Figure \ref{fig:result_within_imagen}.

In order to make sure the artifacts are faithful to the input prompt, we use a VQA filtering step to make sure the image adheres to the mentioned cultural concept. For example, if T2I images are generated for the prompt, ``Image of Japanese cuisine'', we verify faithfulness by passing the image and the question, ``Does dish in the image belong to Japanese cuisine?''. The unfaithful images are simply removed from the $\overline{\mathrm{VS}}$ score calculations, this affecting the score. We assume uniform quality of artifacts for this experiment.

\section{Proof of the scaling property.} 
Denote by $s(\cdot)$ the importance scoring function. We have 
\label{proof_theorem}
\begin{align*}
     q\overline{\mathrm{VS}}_q(\mathcal{C}; k, s) 
     &= \left(\frac{1}{N} \sum_{i=1}^{N} s(x_i)\right) \left(\frac{VS_q(\mathcal{C}; k)}{N}\right)\\
     &= \left(\frac{1}{NM} \sum_{i=1}^{N} M\cdot s(x_i)\right)\left(\frac{M\cdot VS_q(\mathcal{C}; k)}{NM}\right)\\
     &= \left(\frac{1}{NM} \sum_{i=1}^{NM} s(x_i)\right)\left(\frac{M\cdot VS_q(\mathcal{C}'; k)}{NM}\right)\\
     &= M\cdot q\overline{\mathrm{VS}}_q(\mathcal{C}'; k, s)
     .
\end{align*}

\section{Sensitivity of VS to prompt rephrasing and random seeds selection}

\begin{figure*}[h]  
    \centering
    \includegraphics[width=0.9\textwidth]{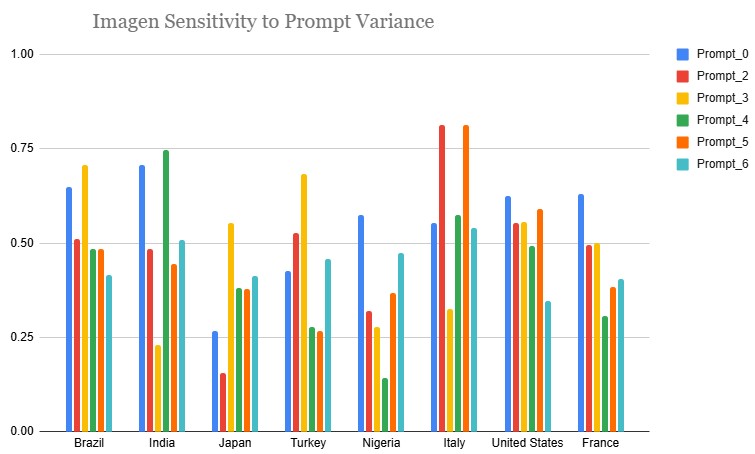}  % Set width to textwidth
    \caption{Sensitivity of VS for different prompt templates reported on Imagen 2}
    \label{fig:results_imagen_sensitivity}
\end{figure*}

Like seeds play an important role, we wish to see the effect of prompt rephrasing on the diversity score. We rephrase the country-specific under-specified prompt into 5 variants using GPT-4-Turbo and report the scores across 8 countries for the Imagen-2 model. Results are presented in Figure \ref{fig:results_imagen_sensitivity}.

\section{Input Prompt Templates}

Prompts in the CUBE-1K Benchmark are used for evaluation of Cultural Awareness of the Text-to-Image models. These include 1K+ prompts spanning across 8 countries and 3 cultural concepts. The prompts are constructed by sampling artifacts from CUBE-CSpace, and using them to fill prompt templates for each cultural concept. These prompt templates are given in Table \ref{tab:prompt_templates}.

\begin{table*}[h]
\centering
\footnotesize
\begin{tabular}{p{0.2\textwidth} p{0.65\textwidth}}
\toprule
\textbf{Cultural Concept} & \textbf{Prompt Template} \\
\midrule
\textbf{Cuisine} & A high resolution image of <food> from <country\_name> cuisine. \\
\midrule
\textbf{Landmarks} & A panoramic view of <place\_name> in <country\_name>. \\
\midrule
%\textbf{Art \& Clothing} & \\
\textbf{Art} & \\
- Clothing & Image of a person in <clothes> from <country\_name>. \\
- Painting & A <style\_of\_painting> painting from <country\_name>. \\
% - Folk Art & A traditional performance of <folk\_art> in <country\_name>. \\
% - Dance Style & A live performance of <dance\_style> dance from <country\_name>.\\
- Performance Art & An image of performance of <performing\_art> from <country\_name>.\\
\midrule
\midrule
\multicolumn{2}{p{0.93\textwidth}}{\textbf{Negative Prompt}: "multiple items, blurry, painting, cartoon, people, human, man, woman, artificial, multiple images, nsfw, bad quality, bad anatomy, worst quality, low quality, low resolutions, extra fingers, blur, blurry, ugly, wrong proportions, watermark, image artifacts, lowres, jpeg artifacts, deformed, noisy"} \\
\bottomrule
\end{tabular}
\caption{Prompt templates used to probe the model for cultural awareness for a given country and cultural concept. Here <country\_name> is replaced by the appropriate country, and <food>, <place\_name> and so on are artifacts sampled from CUBE-1K that are replaced appropriately for each cultural concept.}
\label{tab:prompt_templates}
\end{table*}

\begin{figure}[htbp] % 'htbp' to specify figure placement preference
    \centering
    
    % Heading for Cuisine Maps
    \textbf{Cuisine}
    \vspace{0.5em} % Add some vertical space

    % Cuisine Maps
    \begin{subfigure}{0.46\textwidth}  % Allocate width of half the text width
        \centering
        \fbox{\includegraphics[width=0.98\linewidth]{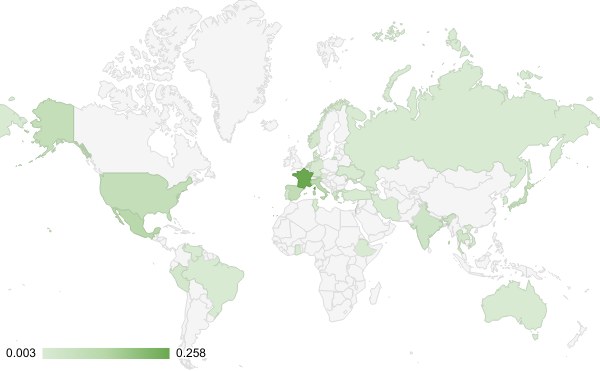}}
        \caption{Imagen }
        \label{fig:sub1}
    \end{subfigure}
    \hfill % Horizontal fill to push next subfigure to the edge
    \begin{subfigure}{0.46\textwidth}  
        \centering
         \fbox{\includegraphics[width=0.98\linewidth]{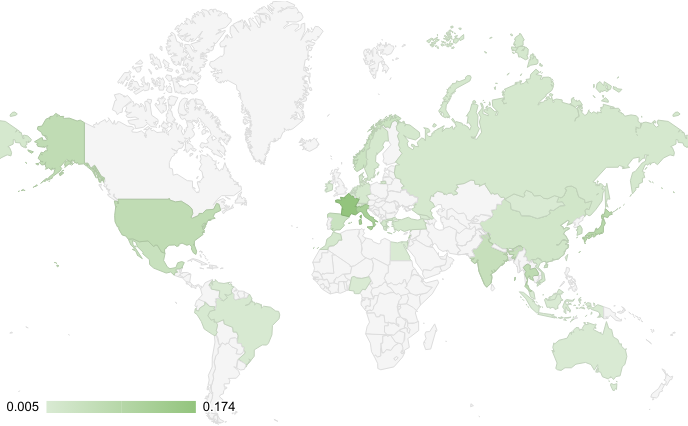}}
        \caption{SDXL }
        \label{fig:sub2}
    \end{subfigure}

    \begin{subfigure}{0.46\textwidth}  
        \centering
        \fbox{\includegraphics[width=0.98\linewidth]{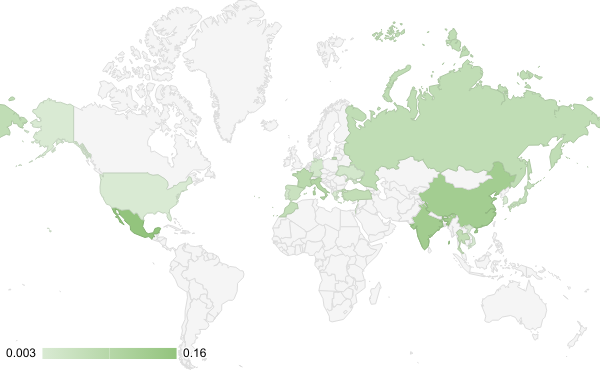}}
        \caption{Playground }
        \label{fig:sub3}
    \end{subfigure}
    \hfill % Horizontal fill
    \begin{subfigure}{0.46\textwidth}  
        \centering
        \fbox{\includegraphics[width=0.98\linewidth]{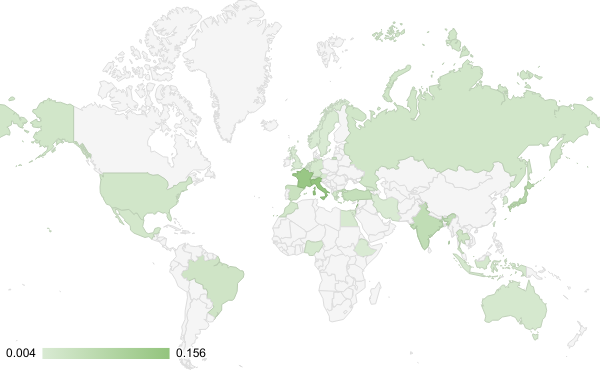}}
        \caption{RealVis }
        \label{fig:sub4}
    \end{subfigure}
    
    % Add vertical space for separation
    \vspace{2em} % Add some vertical space

    % Heading for Landmark Maps
    \textbf{Landmarks}
    \vspace{0.5em} % Add some vertical space

    % Landmark Maps
    \begin{subfigure}{0.46\textwidth}  % Allocate width of half the text width
        \centering
        \fbox{\includegraphics[width=0.98\linewidth]{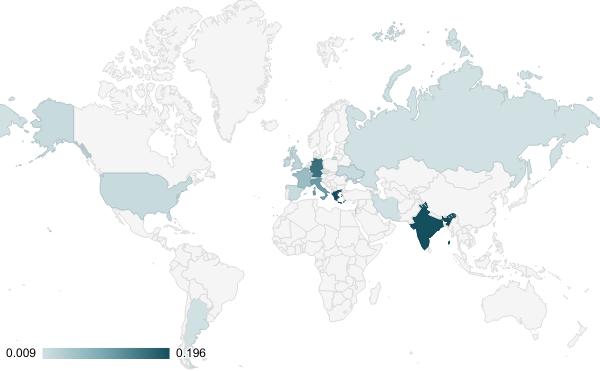}}
        \caption{Imagen }
        \label{fig:sub5}
    \end{subfigure}
    \hfill % Horizontal fill to push next subfigure to the edge
    \begin{subfigure}{0.46\textwidth}  
        \centering
         \fbox{\includegraphics[width=0.98\linewidth]{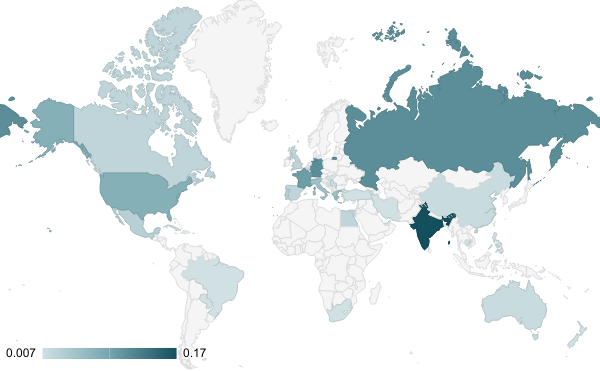}}
        \caption{SDXL }
        \label{fig:sub6}
    \end{subfigure}

    \begin{subfigure}{0.46\textwidth}  
        \centering
        \fbox{\includegraphics[width=0.98\linewidth]{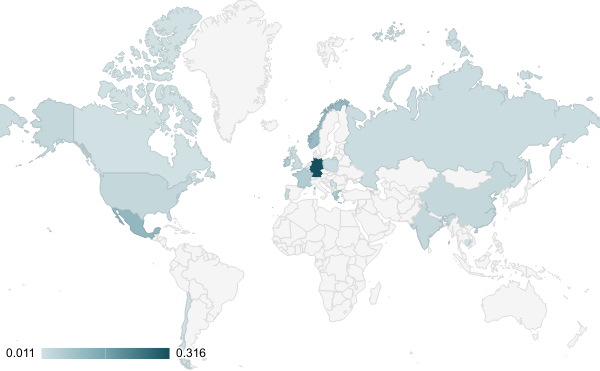}}
        \caption{Playground}
        \label{fig:sub7}
    \end{subfigure}
    \hfill % Horizontal fill
    \begin{subfigure}{0.46\textwidth}  
        \centering
        \fbox{\includegraphics[width=0.98\linewidth]{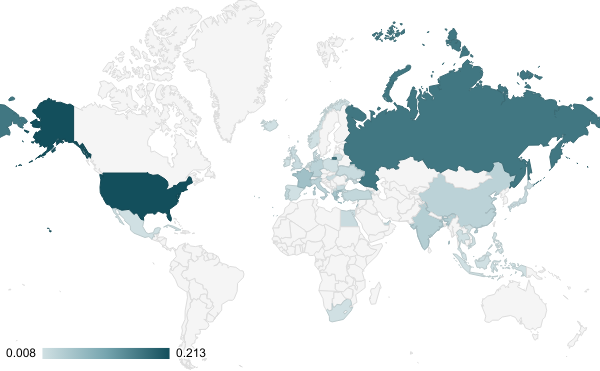}}
        \caption{RealVis }
        \label{fig:sub8}
    \end{subfigure}
    
    \caption{Geo-cultural inclination for \textbf{Cuisine} (top) and \textbf{Landmark} (bottom) concepts when the models are prompted to measure \textit{global geo-cultural diversity}. "Produce a high quality image of a dish." and "High definition photo of a monument." are used as the prompts for cuisine and landmark respectively. 5 prompt variants and 80 different seeds are used to generate 400 images per concept. The figures represent the normalized frequency of the country associated with each generated image.}
    \label{fig:wold_map_models_combined}
\end{figure}

\begin{comment}

\begin{table}[h]
\centering
\caption{Comparison between the generated images from Imagen 2 and Stable diffusion for Question 2 and Question 3. \marco{Removing this since I think it was based on the consensus annotation only, not the majority one we use.}} 
\label{tab:human_annotation_results}
\begin{tabular}{l|c|c}
\toprule
Model &	Imagen 2 & Stable Diffusion	\\
\midrule
\midrule
Question 2 & consensus rating & consensus rating \\ 
\midrule
Exactly & 73.08\% & 44.83\% \\
Mostly	& 3.85\% & 0.00\% \\
Somewhat & 3.85\% & 6.90\% \\
Not at all & 19.23\% & 37.93\% \\
No agreement & 0.00\% & 10.34\% \\
Average Score & \bf{4.12} & 3.12 \\
\midrule
Question 3 & consensus rating & consensus rating \\ 
\midrule
Extremely & 61.54\% & 41.38\% \\
Mostly & 34.62\% & 20.69\% \\
Somewhat & 3.85\% & 34.48\% \\
Not at all & 0.00\% & 3.45\% \\
Average Score & \bf{4.58} & 3.96 \\
\bottomrule
\end{tabular}
\end{table}

\end{comment}

\begin{figure*}[!ht]
    \centering
    \includegraphics[width=0.98\textwidth]{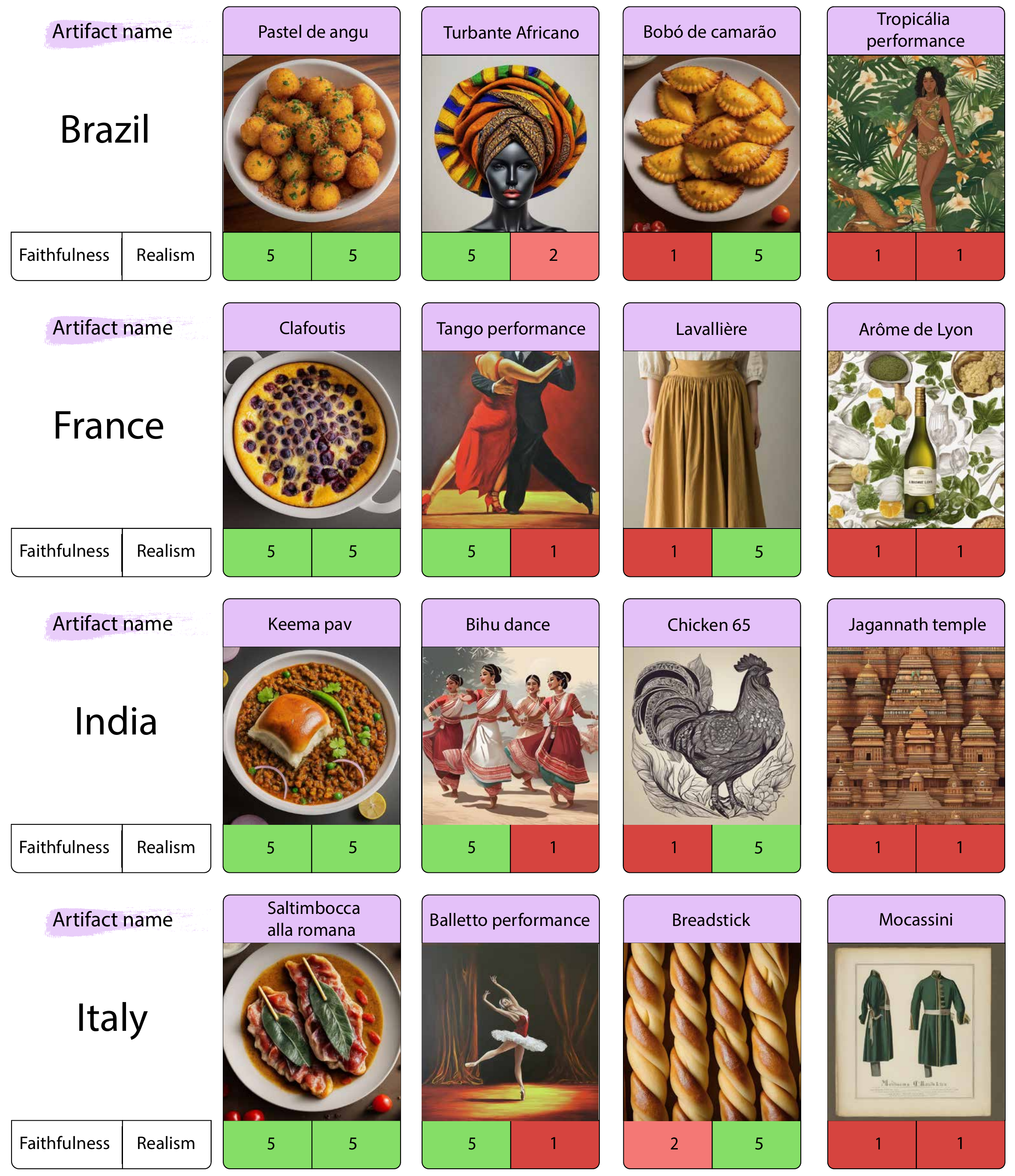}
    
    \vspace{-0.2\baselineskip}
    \caption{Qualitative examples of artifacts generated from T2I models, along with Faithfulness and Realism scores as described in Section \ref{sec:evaluating_cultural_awareness}: Evaluating Cultural Awareness.}
    \label{fig:qualitative_examples-1}
\end{figure*}

\begin{figure*}[!ht]
    \centering
    \includegraphics[width=0.98\textwidth]{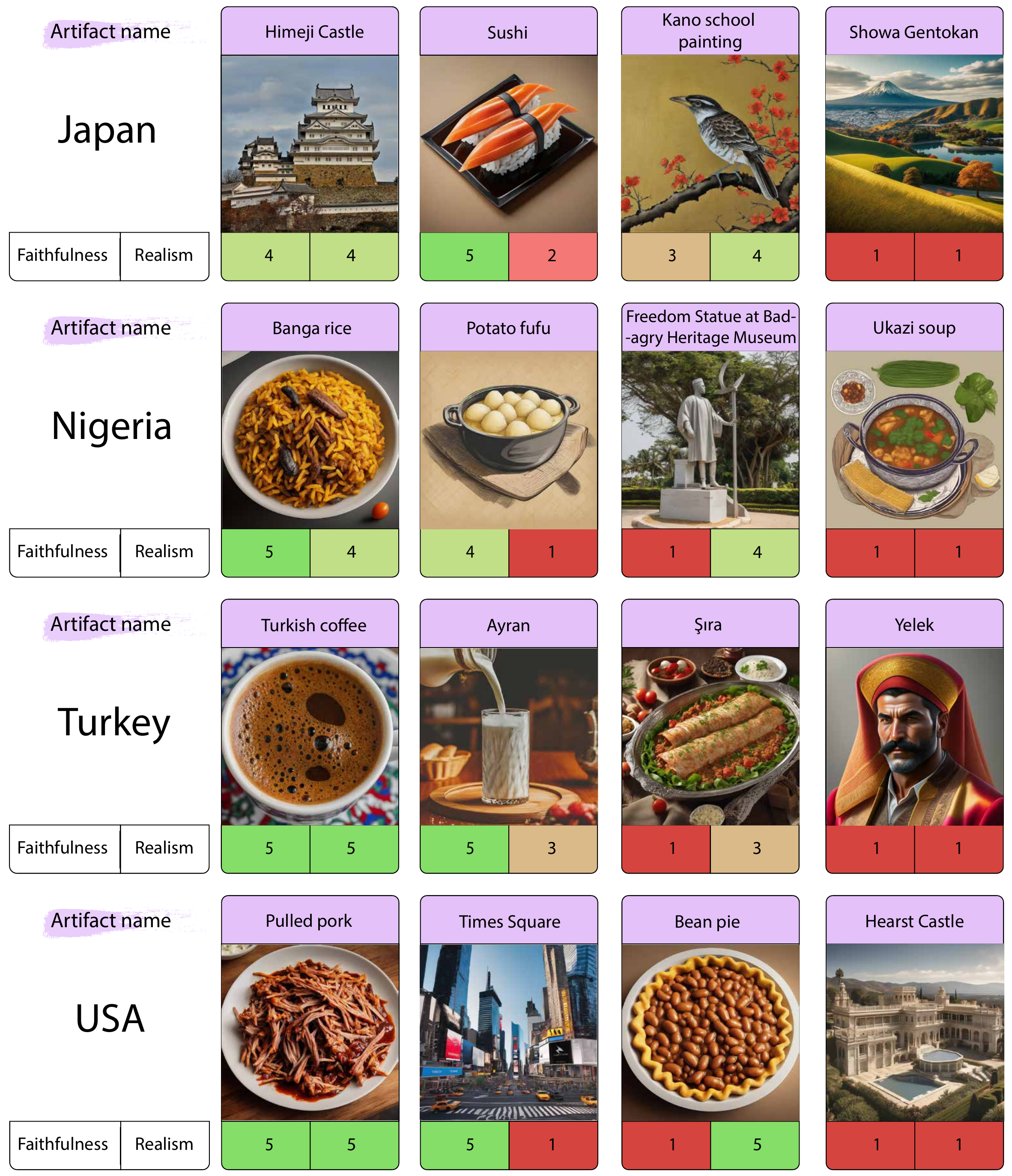}
    
    \vspace{-0.2\baselineskip}
    \caption{Qualitative examples of artifacts generated from T2I models, along with Faithfulness and Realism scores as described in Section \ref{sec:evaluating_cultural_awareness}: Evaluating Cultural Awareness.}
    \label{fig:qualitative_examples-2}
\end{figure*}

\renewcommand\arraystretch{1.5}  % Adds padding inside cells
\begin{table}[ht]
\centering
\captionsetup{justification=centering}
\resizebox{\textwidth}{!}{%
\begin{tabular}{|>{\centering\arraybackslash}m{2.5cm}|>{\centering\arraybackslash}m{2cm}|>{\centering\arraybackslash}m{1.5cm}|>{\centering\arraybackslash}m{3cm}|>{\arraybackslash}m{7cm}|}
\hline
\textbf{Image} & \textbf{Artifact} & \textbf{Country} & \textbf{Edge case type} & \textbf{Rater comment} \\ \hline
\includegraphics[width=2.5cm]{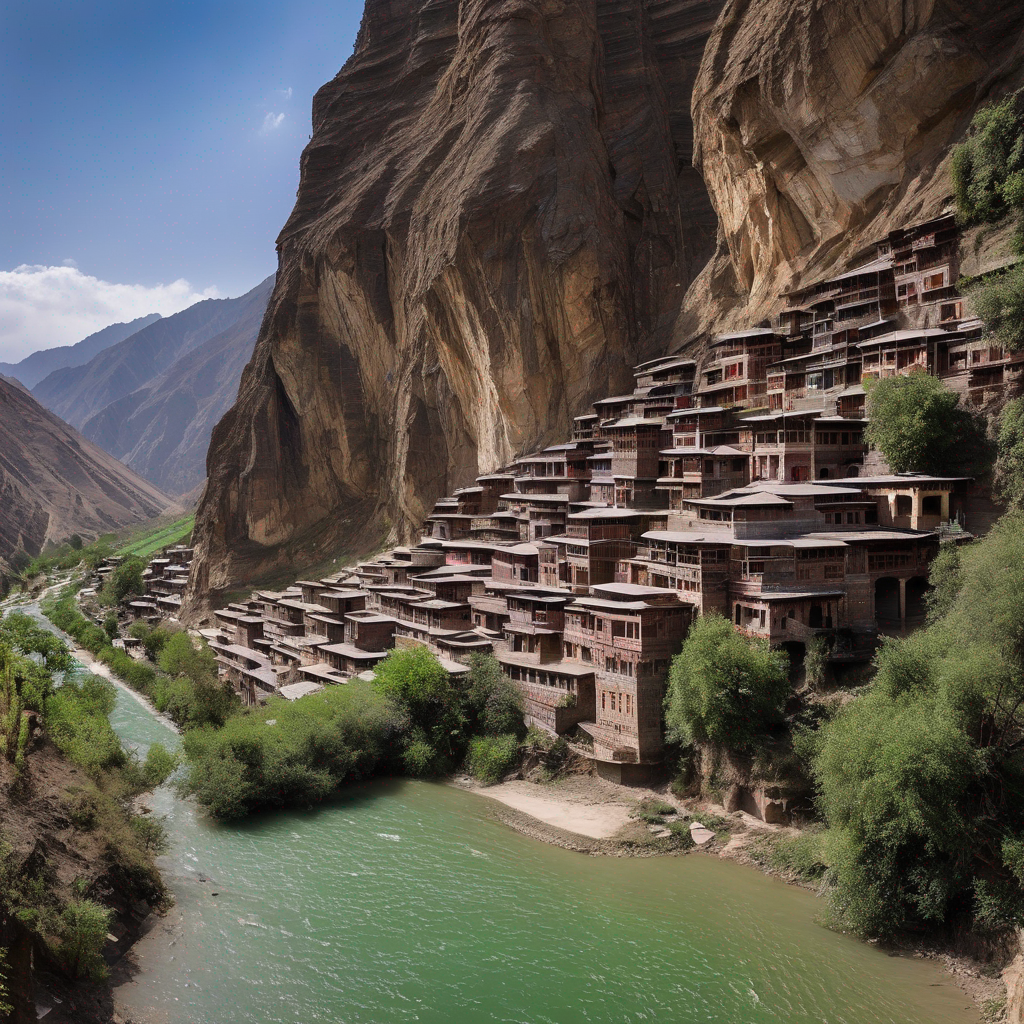} & Phuktal Monastery & India & Faithfulness  & The Phuktal Monastery is actually at a certain on the mountain, however in the image they are on the ground. \\ \hline
\includegraphics[width=2.5cm]{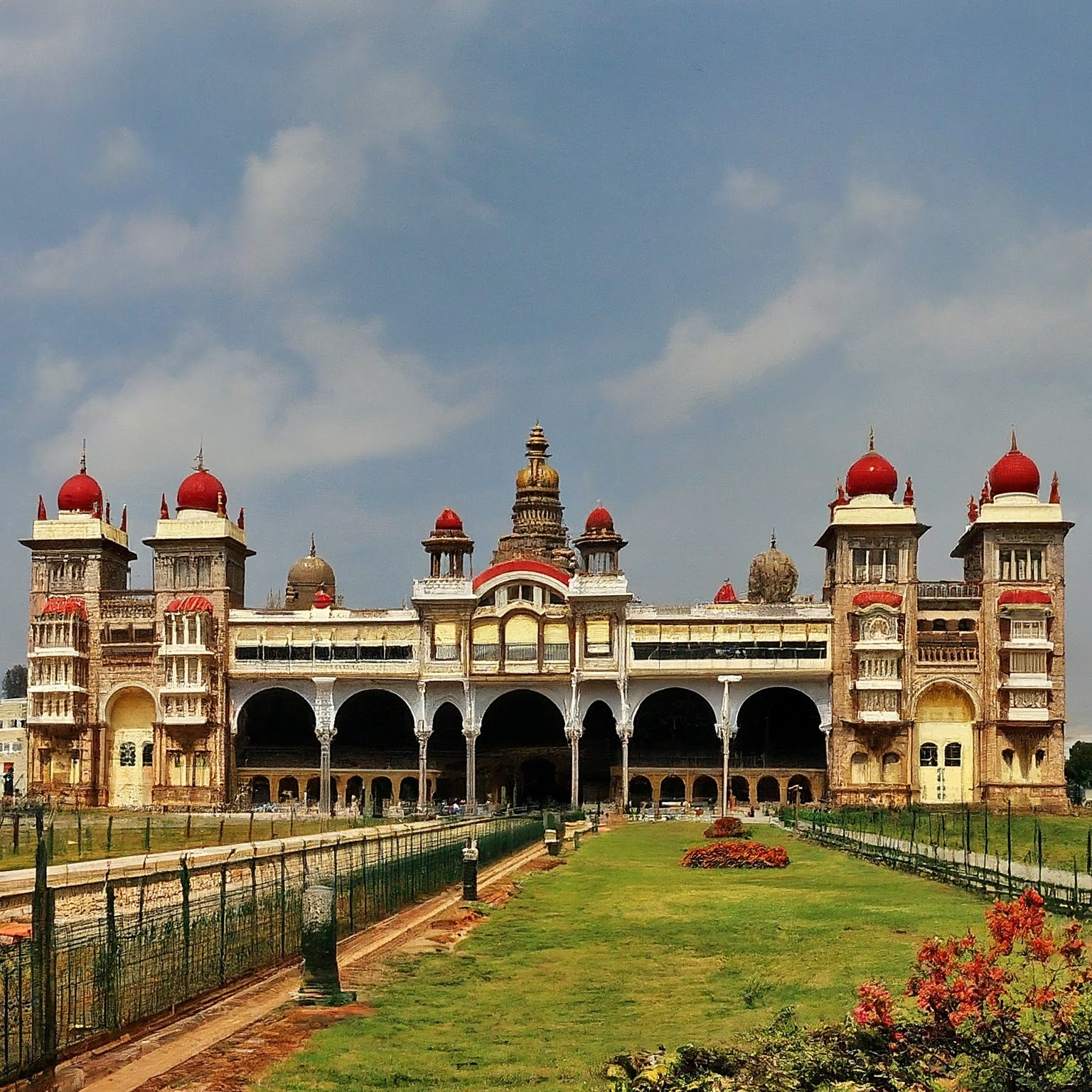} & Mysore Palace & India & Realism & Though the image looks perfect the minor distortions which looks unrealistic \\ \hline
\includegraphics[width=2.5cm]{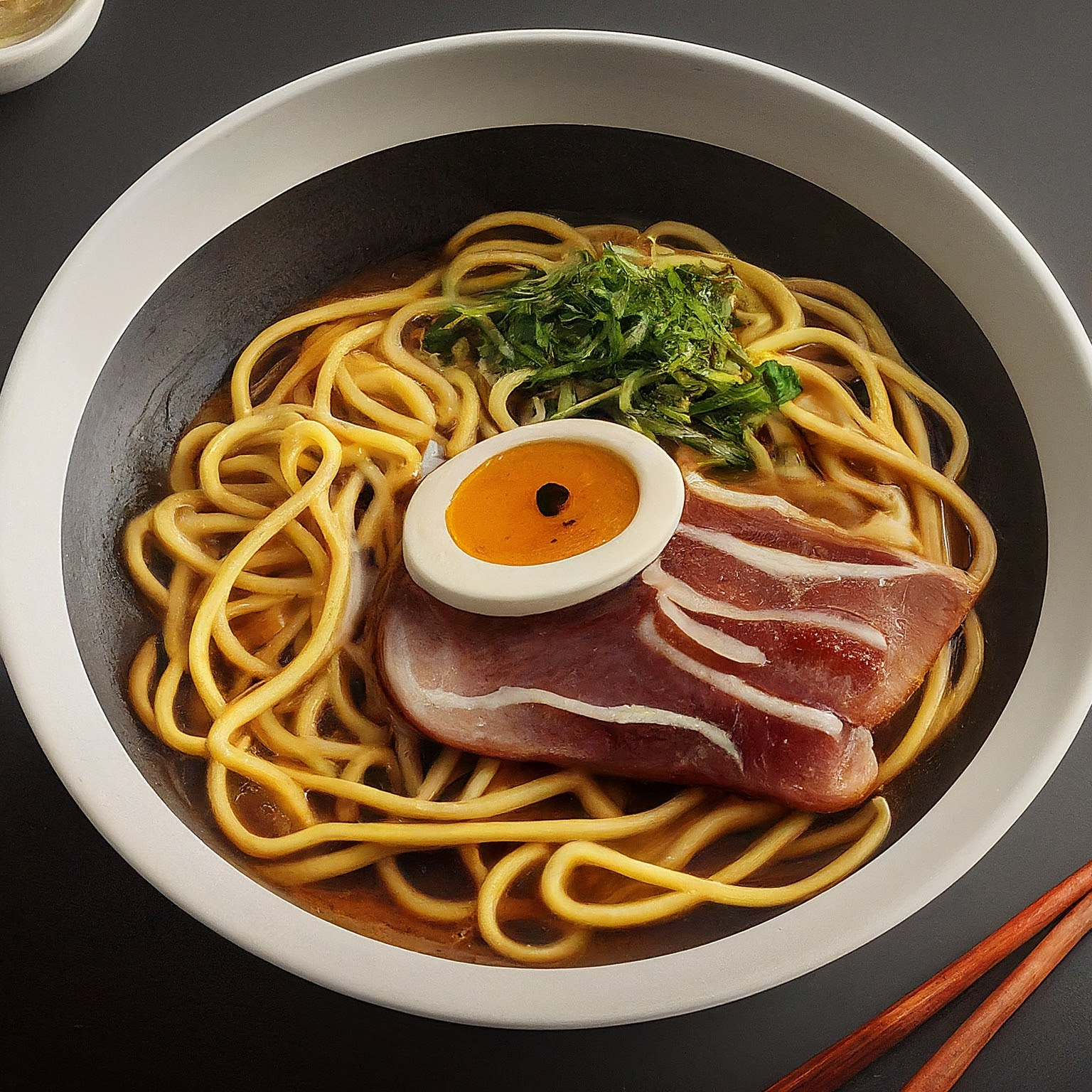} & Ramen & Japan & Variations in dishes & Ramen is a traditional dish with many regional varieties and a wide range of toppings. It's difficult to assess whether this image looks like ramen, and any answer ranging from "Somewhat" to "Exactly" is reasonable. \\ \hline
\includegraphics[width=2.5cm]{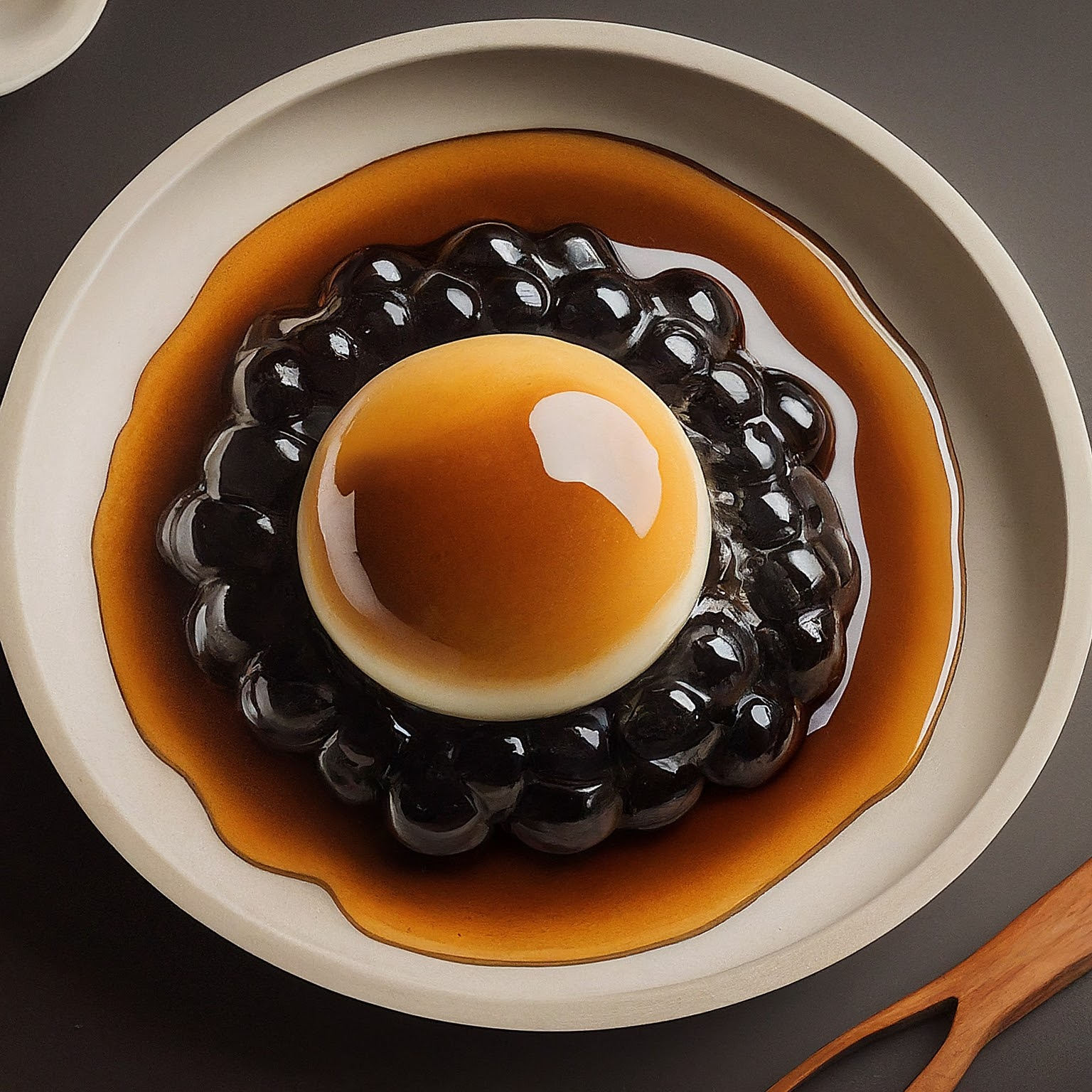} & Raindrop Cake & Japan & Disregarding the prompt for Q1 & The food image is unrealistic, and it's unclear what it is. Judging from the image alone, the correct answer is "Maybe" or even "No" (one rater argued it looks more like a Taiwanese dish). But if you consider the prompt for a "raindrop cake," it's easy to see how this is an unrealistic/inaccurate version of a raindrop cake, so the answer would be "Yes."  \\ \hline
\includegraphics[width=2.5cm]{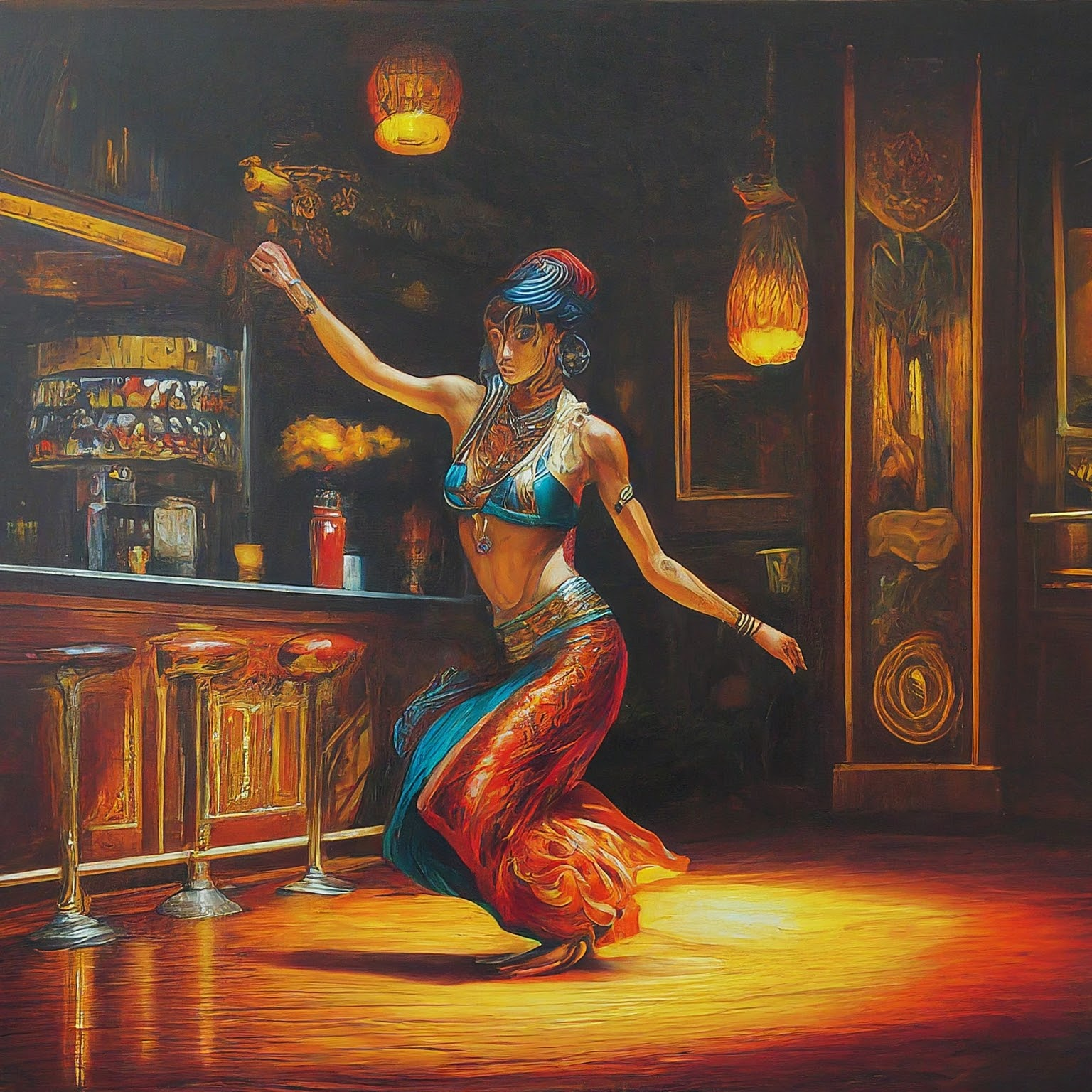} & Bar performance & Turkey & Unrealism for Q1 & For this image, two raters selected "Maybe" while one picked "No" because this person and their clothing are so unrealistic, it's difficult to assess whether they could belong to Turkish culture. When images are cartoonish, they may also be interpreted as stereotypical. \\ \hline
\includegraphics[width=2.5cm]{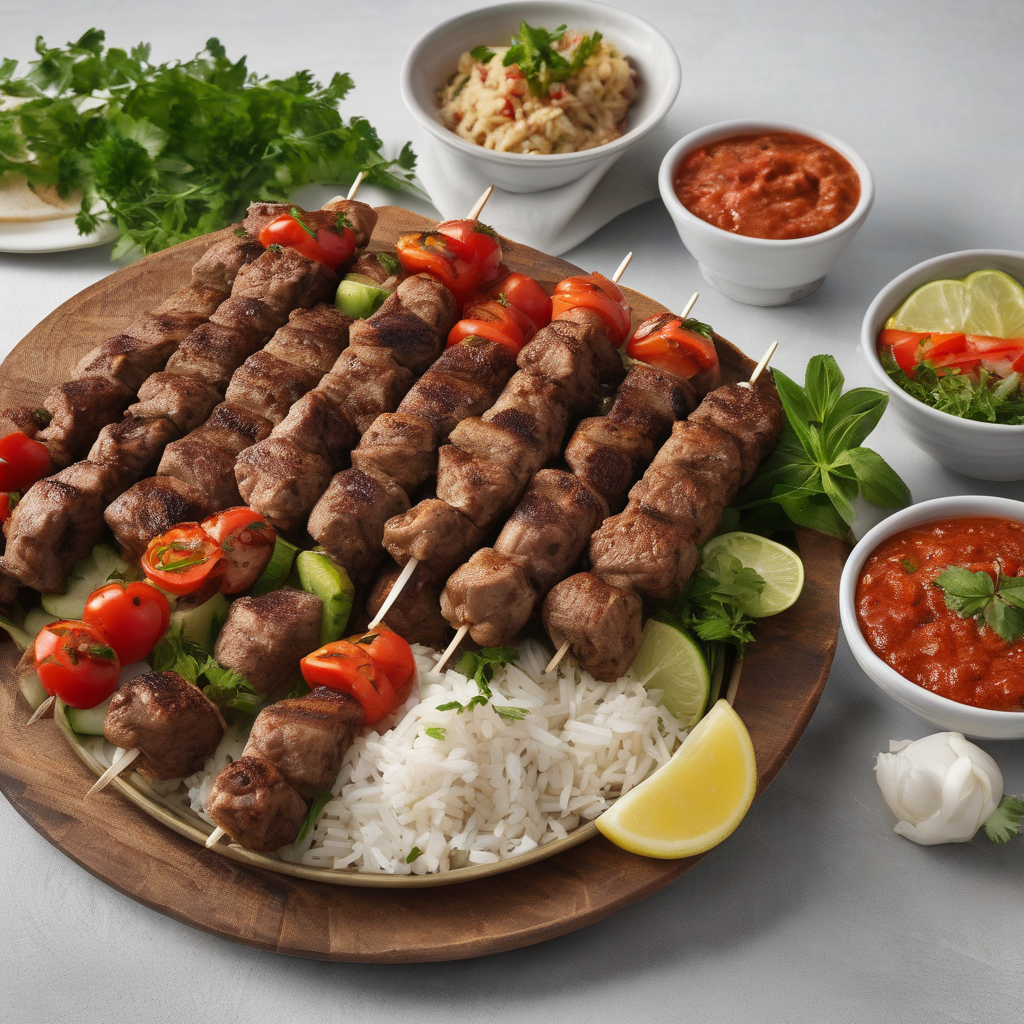} & Kebab & Turkey & Unrealism for Q2 & It's unusual to see a lemon next to this particular dish. Two raters interpreted this as a realism issue, while the third marked "A little" for Question 2 because it doesn't match the usual appearance of the dish. \\ \hline
\includegraphics[width=2.5cm]{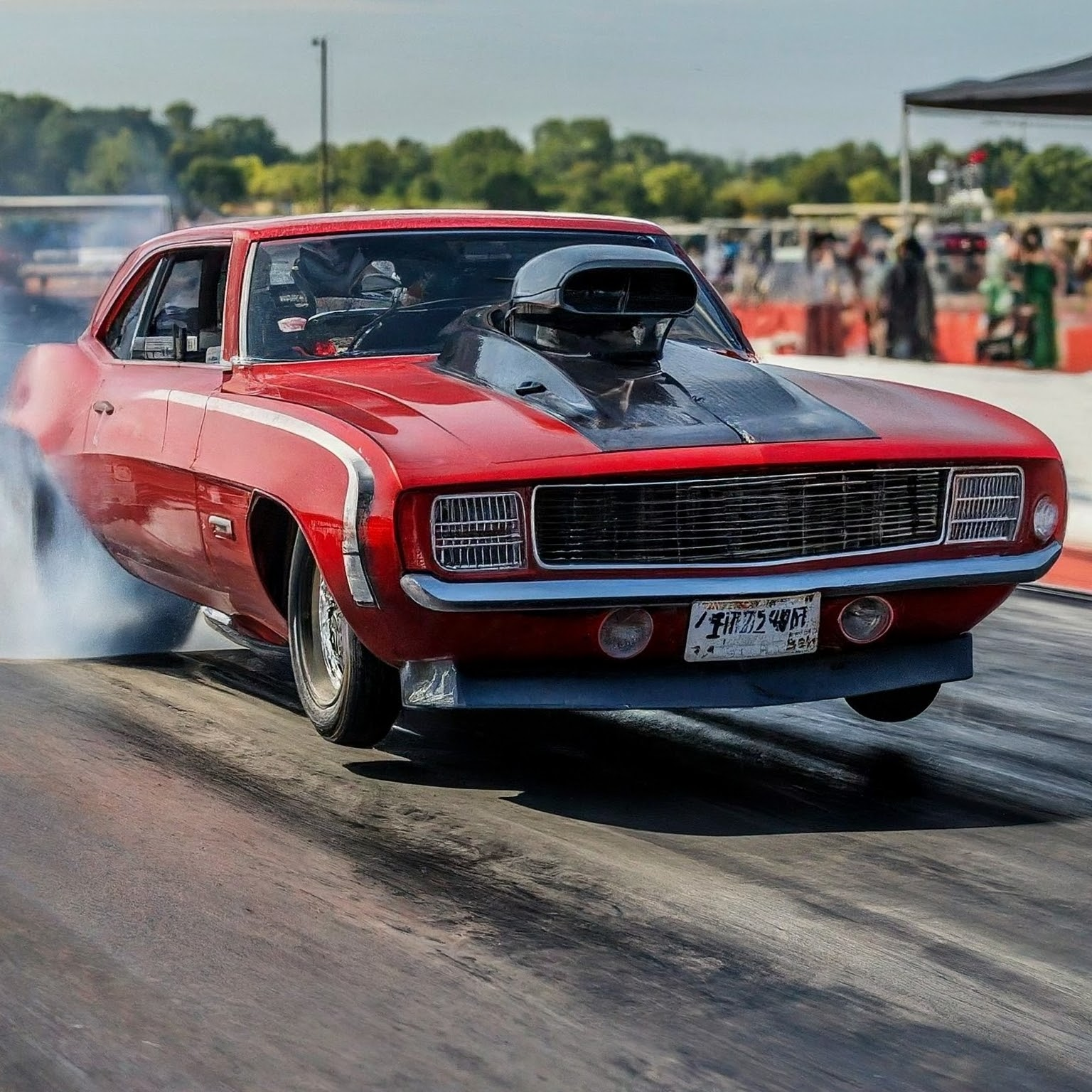} & Drag Performance & USA & Literal interpretation & This prompt asks for a "drag performance," and the photo is of a car race. A drag race could be a kind of "performance," but it's obviously not what the prompt meant. We chose 2: A Little, because the picture was not related to drag shows but still had some logic to it. \\ \hline
\includegraphics[width=2.5cm]{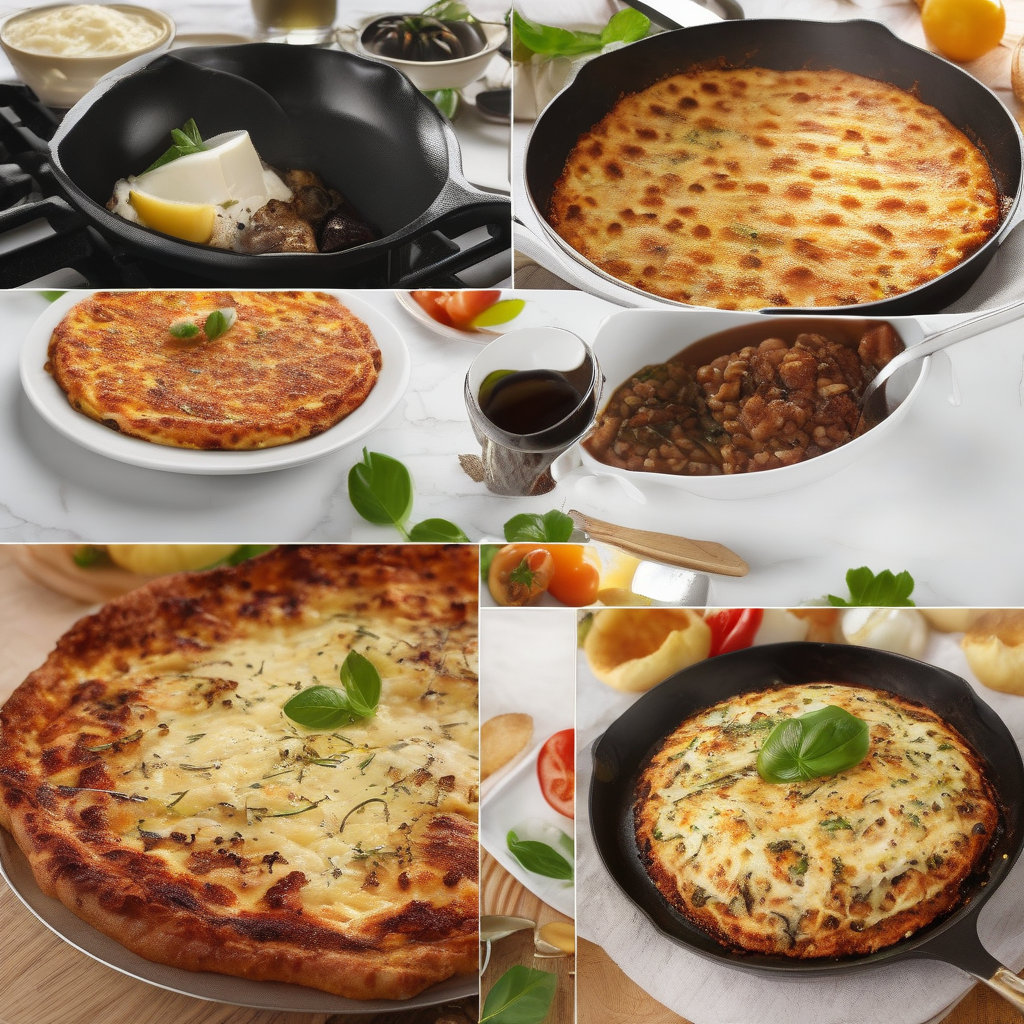} & Frico & Italy & Composite photo & In this case, we advised the rater to evaluate the realism of the individual photos. They landed on a 3 because some of the pots were deformed, the basil wasn't right, etc. \\ \hline
\end{tabular}
}
\caption{Interesting edge cases in cultural awareness evaluation across geo-cultures}
\label{fig:interesting_examples}
\end{table}

\end{document}